\documentclass[10pt,twocolumn]{article}
\usepackage[utf8]{inputenc}
\usepackage[T1]{fontenc}
\usepackage{mathptmx}
\usepackage{xcolor}
\usepackage{amsmath,amssymb}
\usepackage{tikz}
\usepackage{pgfplots}
\pgfplotsset{compat=1.18}
\usetikzlibrary{shapes,arrows,arrows.meta,backgrounds,calc,fit,positioning}
\usepackage{booktabs}
\usepackage{multirow}
\usepackage{geometry}
\usepackage{hyperref}
\hypersetup{colorlinks=true,linkcolor=blue,citecolor=blue,urlcolor=blue,pdftitle={Not All Attention Is Equal: A Quantitative Survey of the EEI Trade-off},pdfauthor={Aditya Singh}}
\usepackage{natbib}
\setcitestyle{aysep={}}
\usepackage{caption}
\usepackage{array}
\usepackage{titlesec}
\titlespacing*{\section}{0pt}{6pt}{4pt}
\titlespacing*{\subsection}{0pt}{4pt}{2pt}
\titleformat{\section}{\normalfont\Large\bfseries}{\thesection}{0.4em}{}
\titleformat{\subsection}{\normalfont\large\bfseries}{\thesubsection}{0.4em}{}
\usepackage{graphicx}
\usepackage{microtype}
\usepackage{placeins}
\makeatletter
\AtBeginDocument{%
  \newif\ifshrinkfloats
  \ifdim\f@size pt=12pt\relax\shrinkfloatstrue\else\shrinkfloatsfalse\fi
  \ifshrinkfloats\hfuzz=8pt\relax\fi
}
\newcommand{\sflt}[2]{\ifshrinkfloats#1\else#2\fi}
\makeatother

\title{Not All Attention Is Equal: A Quantitative Survey of the EEI Trade-off}
\author{Aditya Singh}
\date{}

\begin{document}
\maketitle

\section*{Abstract}

Attention mechanisms have driven machine learning for a decade, from neural machine translation to language models that do general-purpose reasoning. This survey covers four connected threads: how attention was first formulated for sequence-to-sequence tasks, how it was adapted for computer vision, what efficiency innovations broke the quadratic bottleneck, and what we've learned about interpretability along the way.

We define three criteria (efficiency, expressiveness, interpretability) and use them to compare attention mechanisms across the methods we survey. A scoring rubric, a comparison table, and a map of twenty-one methods on the efficiency--expressiveness plane accompany the analysis. Scores come from a single rater with an assumed $\pm 1$-point perturbation range; a deterministic, seed-fixed Monte Carlo analysis (200\,000 samples, NumPy PCG64 generator, seed 42) shows that, under the assumed perturbation model, the mean fraction of perturbation samples in which a method's rank changes by more than one position is 67--70\% (in essentially every sample, at least one method changes rank). Because a rank-matched null model with randomly drawn scores reproduces the same overall stability profile, the perturbation analysis supports coarse tier-level comparisons rather than fine-grained rank claims. Scores are single-rater estimates; the perturbation analysis assumes an independent one-point range and tests ranking sensitivity, not empirical measurement uncertainty.

The survey traces attention from Bahdanau--Luong alignment through the scaled dot-product formulation in the Transformer (Vaswani et al., 2017), then into vision via patch embeddings, object queries, and hierarchical windows. The main body reviews efficiency innovations (fixed and learnable sparse patterns, linear-time kernel approximations, IO-aware exact algorithms such as FlashAttention (Dao et al., 2022; Dao, 2024; Shah et al., 2024), and state-space alternatives including Mamba (Gu \& Dao, 2024)). On interpretability, we cover induction heads (Olsson et al., 2022), superposition (Elhage et al., 2022), and the attention-SSM duality (Dao \& Gu, 2024).

Other contributions include a structured narrative review methodology, a benchmark synthesis with cross-study caveats, a gap analysis of five open problems, and an evolution timeline from 2015 to 2026. The paper closes by framing attention's history as an expansion of the efficiency--expressiveness--interpretability (EEI) frontier and identifying directions for future work: unified efficiency benchmarks, learned routing for hybrid architectures, length generalization, and scalable mechanistic interpretability.

\begin{center}\rule{0.5\linewidth}{0.5pt}\end{center}

\section{Introduction}\label{introduction}

The Transformer architecture (Vaswani et al., 2017) replaced recurrent and convolutional sequence mixing with self-attention as the primary mechanism within a feedforward network stack. This let sequence models trade linear-time recurrence for quadratic-time pairwise interactions, and this trade yielded substantial gains in translation quality and training parallelism in the settings evaluated by Vaswani et al. (2017). What followed was a wave of architectural innovations that pushed attention well beyond machine translation.

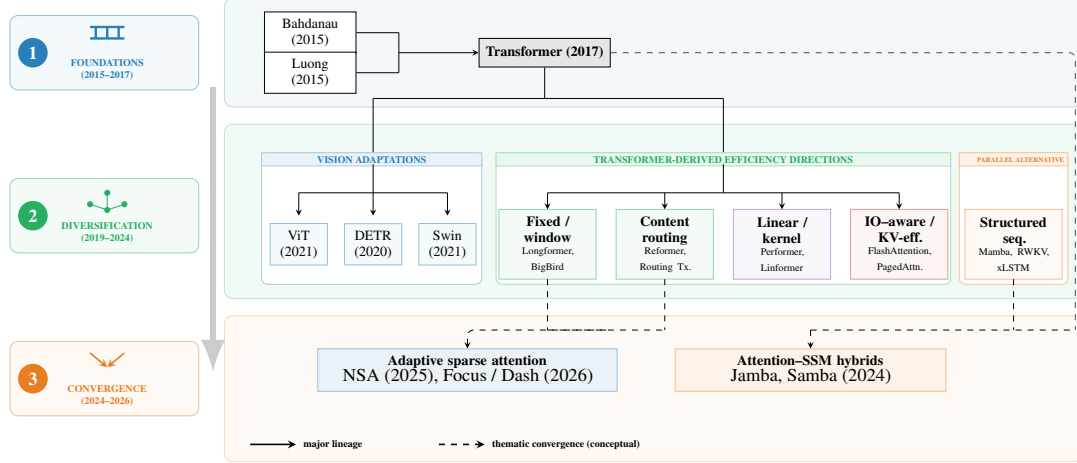
\begin{figure*}[!t]
\centering
\resizebox{\ifshrinkfloats 0.87\textwidth\else 0.9\textwidth\fi}{!}{\definecolor{colorblue}{HTML}{2980B9}
\definecolor{colorgreen}{HTML}{27AE60}
\definecolor{colorpurple}{HTML}{8E44AD}
\definecolor{colorred}{HTML}{C0392B}
\definecolor{colororange}{HTML}{E67E22}
\definecolor{colorcyan}{HTML}{1ABC9C}

\definecolor{bgblue}{HTML}{F2F7FA}
\definecolor{bggreen}{HTML}{F1F9F4}
\definecolor{bgorange}{HTML}{FEF9F5}

\begin{tikzpicture}[
  x=1cm,
  y=1cm,
  familybox/.style={
    font=\scriptsize,
    align=center,
    text width=1.50cm,
    minimum width=1.70cm,
    minimum height=1.22cm,
    inner sep=1pt,
    outer sep=0pt
  },
  branchheadertext/.style={
    font=\bfseries\tiny,
    inner sep=0pt,
    align=center
  }
]

\draw[draw=colorblue!30, fill=bgblue, rounded corners=4pt]
  (2.25,5.0) rectangle (17.25,6.9);
\draw[draw=colorgreen!30, fill=bggreen, rounded corners=4pt]
  (2.25,1.65) rectangle (17.25,4.7);
\draw[draw=colororange!30, fill=bgorange, rounded corners=4pt]
  (2.25,-1.2) rectangle (17.25,1.35);

\draw[draw=colorblue!50, fill=colorblue!5, rounded corners=4pt]
  (-1.5,5.3) rectangle (1.8,6.6);
\node[circle, fill=colorblue, text=white, font=\bfseries\small,
  inner sep=2.5pt] at (-1.1,5.95) {1};
\draw[colorblue, line width=1.2pt] (-0.1,6.4) -- (0.5,6.4);
\draw[colorblue, line width=1.2pt] (0.0,6.4) -- (0.0,6.2);
\draw[colorblue, line width=1.2pt] (0.2,6.4) -- (0.2,6.2);
\draw[colorblue, line width=1.2pt] (0.4,6.4) -- (0.4,6.2);
\draw[colorblue, line width=1.2pt] (-0.1,6.2) -- (0.5,6.2);
\node[color=colorblue, font=\bfseries\tiny, align=center]
  at (0.2,5.65) {FOUNDATIONS\\(2015--2017)};

\draw[draw=colorgreen!50, fill=colorgreen!5, rounded corners=4pt]
  (-1.5,2.45) rectangle (1.8,3.75);
\node[circle, fill=colorgreen, text=white, font=\bfseries\small,
  inner sep=2.5pt] at (-1.1,3.1) {2};
\fill[colorgreen] (0.2,3.5) circle (1.5pt);
\fill[colorgreen] (-0.1,3.3) circle (1.5pt);
\fill[colorgreen] (0.5,3.3) circle (1.5pt);
\fill[colorgreen] (0.2,3.2) circle (1.5pt);
\draw[colorgreen, thin] (0.2,3.2) -- (0.2,3.5);
\draw[colorgreen, thin] (0.2,3.2) -- (-0.1,3.3);
\draw[colorgreen, thin] (0.2,3.2) -- (0.5,3.3);
\node[color=colorgreen, font=\bfseries\tiny, align=center]
  at (0.2,2.8) {DIVERSIFICATION\\(2019--2024)};

\draw[draw=colororange!50, fill=colororange!5, rounded corners=4pt]
  (-1.5,-0.4) rectangle (1.8,0.9);
\node[circle, fill=colororange, text=white, font=\bfseries\small,
  inner sep=2.5pt] at (-1.1,0.25) {3};
\draw[colororange, thick, ->, >=stealth] (-0.1,0.7) -- (0.2,0.45);
\draw[colororange, thick, ->, >=stealth] (0.5,0.7) -- (0.2,0.45);
\node[color=colororange, font=\bfseries\tiny, align=center]
  at (0.2,-0.05) {CONVERGENCE\\(2024--2026)};

\draw[-{Latex[scale=1.0]}, line width=3pt, gray!40]
  (2.05,5.35) -- (2.05,0.35);

\node[draw, fill=white, font=\scriptsize, align=center,
  minimum width=1.6cm, minimum height=0.5cm]
  (Bahdanau) at (3.75,6.3) {Bahdanau\\(2015)};
\node[draw, fill=white, font=\scriptsize, align=center,
  minimum width=1.6cm, minimum height=0.5cm]
  (Luong) at (3.75,5.6) {Luong\\(2015)};
\node[draw, fill=gray!20, font=\bfseries\scriptsize, align=center,
  minimum width=2.0cm, minimum height=0.5cm]
  (Transformer) at (7.85,5.95) {Transformer (2017)};

\draw (Bahdanau.east) -- (5.3,6.3);
\draw (Luong.east) -- (5.3,5.6);
\draw (5.3,6.3) -- (5.3,5.6);
\draw[->, >=stealth] (5.3,5.95) -- (Transformer.west);
\draw[dashed, black!70, rounded corners=3pt]
  (Transformer.east) -- (17.13,5.95) -- (17.13,1.1);

\draw (Transformer.south) -- (7.85,5.15) -- (4.85,5.15);
\draw (4.85,5.15) -- (4.85,4.2);
\draw (7.85,5.15) -- (10.975,5.15);
\draw (10.975,5.15) -- (10.975,4.2);

\draw[draw=colorblue!40, rounded corners=3pt, fill=white]
  (2.9,1.9) rectangle (6.75,4.2);
\draw[draw=colorblue!40, fill=colorblue!10]
  (2.9,3.97) rectangle (6.75,4.2);
\node[branchheadertext, text=colorblue]
  at (4.825,4.085) {VISION ADAPTATIONS};

\node[draw=colorblue!50, fill=colorblue!5, font=\scriptsize,
  align=center, minimum width=1.0cm, minimum height=0.7cm]
  (ViT) at (3.55,2.6) {ViT\\(2021)};
\node[draw=colorblue!50, fill=colorblue!5, font=\scriptsize,
  align=center, minimum width=1.0cm, minimum height=0.7cm]
  (DETR) at (4.85,2.6) {DETR\\(2020)};
\node[draw=colorblue!50, fill=colorblue!5, font=\scriptsize,
  align=center, minimum width=1.0cm, minimum height=0.7cm]
  (Swin) at (6.15,2.6) {Swin\\(2021)};

\draw (4.85,4.0) -- (4.85,3.5);
\draw (3.55,3.5) -- (6.15,3.5);
\draw[->, >=stealth] (3.55,3.5) -- (3.55,3.1);
\draw[->, >=stealth] (4.85,3.5) -- (4.85,3.1);
\draw[->, >=stealth] (6.15,3.5) -- (6.15,3.1);

\draw[draw=colorgreen!40, rounded corners=3pt, fill=white]
  (7.0,1.9) rectangle (14.95,4.2);
\draw[draw=colorgreen!40, fill=colorgreen!10]
  (7.0,3.97) rectangle (14.95,4.2);
\node[branchheadertext, text=colorgreen]
  at (10.975,4.085) {TRANSFORMER-DERIVED EFFICIENCY DIRECTIONS};

\draw[draw=colororange!45, rounded corners=3pt, fill=white]
  (15.1,1.9) rectangle (17.0,4.2);
\draw[draw=colororange!45, fill=colororange!10]
  (15.1,3.97) rectangle (17.0,4.2);
\node[branchheadertext, text=colororange]
  at (16.175,4.085)
  {\resizebox{1.52cm}{!}{\textbf{PARALLEL ALTERNATIVE}}};

\node[familybox, draw=colorgreen!50, fill=colorgreen!5]
  (Fixed) at (7.9,2.6)
  {\textbf{Fixed /\\window}\\{\tiny Longformer,\\BigBird}};
\node[familybox, draw=colorgreen!50, fill=colorgreen!5]
  (Routing) at (9.95,2.6)
  {\textbf{Content\\routing}\\{\tiny Reformer,\\Routing Tx.}};
\node[familybox, draw=colorpurple!50, fill=colorpurple!5]
  (Linear) at (12.0,2.6)
  {\textbf{Linear /\\kernel}\\{\tiny Performer,\\Linformer}};
\node[familybox, draw=colorred!50, fill=colorred!5]
  (IO) at (14.05,2.6)
  {\textbf{IO--aware /\\KV-eff.}\\{\tiny FlashAttention,\\PagedAttn.}};
\node[familybox, draw=colororange!50, fill=colororange!5]
  (Structured) at (16.05,2.6)
  {\textbf{Structured\\seq.}\\{\tiny Mamba, RWKV,\\xLSTM}};

\draw (10.975,4.0) -- (10.975,3.5);
\draw (7.9,3.5) -- (14.05,3.5);
\draw[->, >=stealth] (7.9,3.5) -- (Fixed.north);
\draw[->, >=stealth] (9.95,3.5) -- (Routing.north);
\draw[->, >=stealth] (12.0,3.5) -- (Linear.north);
\draw[->, >=stealth] (14.05,3.5) -- (IO.north);

\node[draw=colorblue!50, fill=colorblue!10, font=\scriptsize,
  align=center, text width=5.0cm, minimum height=0.7cm]
  (Adaptive) at (6.5,0.4)
  {\textbf{Adaptive sparse attention}\\ \small NSA (2025), Focus / Dash (2026)};
\node[draw=colororange!50, fill=colororange!10, font=\scriptsize,
  align=center, text width=4.5cm, minimum height=0.7cm]
  (Hybrids) at (12.5,0.4)
  {\textbf{Attention--SSM hybrids}\\ \small Jamba, Samba (2024)};

\draw[dashed] (Fixed.south) -- (7.9,1.1);
\draw[dashed] (Routing.south) -- (9.95,1.1);
\draw[dashed] (7.9,1.1) -- (9.95,1.1);
\draw[dashed, ->, >=stealth, rounded corners=3pt]
  (8.925,1.1) -| (Adaptive.north);

\draw[dashed] (Structured.south) -- (16.05,1.1);
\draw[dashed] (12.5,1.1) -- (17.13,1.1);
\draw[dashed, ->, >=stealth] (12.5,1.1) -- (Hybrids.north);

\draw[->, >=stealth, thick] (2.7,-0.9) -- (3.5,-0.9)
  node[right, font=\tiny\bfseries] {major lineage};
\draw[dashed, ->, >=stealth, thick] (6.0,-0.9) -- (6.8,-0.9)
  node[right, font=\tiny\bfseries] {thematic convergence (conceptual)};

\end{tikzpicture}}
\caption{Roadmap of attention evolution. The field progressed from foundational sequence-to-sequence mechanisms through vision adaptations, an efficiency-focused era producing diverse sub-quadratic strategies, and into the current hybrid era where adaptive routing and attention-SSM convergence dominate. Dashed arrows indicate convergent lines of research. Arrows trace thematic lineage rather than strict chronology or direct code inheritance.}
\label{fig:roadmap}
\end{figure*}

Three moments stand out. BERT showed that deep bidirectional Transformers pretrained on unlabeled text produce transferable representations that improved scores across eleven NLP benchmarks (Devlin et al., 2019). Scaling decoder-only autoregressive models then produced emergent capabilities: GPT-3 could do few-shot reasoning without gradient updates (Brown et al., 2020). And the Vision Transformer (ViT) demonstrated that, with sufficient data, patch-based attention can match or exceed convolutional nets on image classification (Dosovitskiy et al., 2021). By 2021, Transformer-based attention had become a dominant sequence-modeling paradigm. A prior survey charted the resulting landscape of efficient attention variants (Tay et al., 2022). Table~\ref{tab:survey-comparison} positions this survey relative to prior work. Figure~\ref{fig:roadmap} previews this lineage, from the Bahdanau--Luong foundations through the efficiency era to the current hybrid architectures.

\begin{table*}[!t]
\centering
\footnotesize
\renewcommand{\arraystretch}{1.15}
\caption{Comparison with prior attention surveys. ``EEI'' denotes the Efficiency--Expressiveness--Interpretability framework introduced here. Checkmarks indicate substantive coverage: for Vision, a dedicated section comparing ViT, DETR, and/or Swin; for the other columns, dedicated coverage of the listed topic. Tay et al.\ (2022) discusses vision applications in passing but lacks a dedicated vision-transformer section.}
\label{tab:survey-comparison}
\begin{tabular}{lccccc}
\toprule
& \rotatebox{55}{\textit{Vision}} & \rotatebox{55}{\textit{FlashAttn}} & \rotatebox{55}{\textit{Mamba/SSM}} & \rotatebox{55}{\textit{Quant.\,EEI}} & \rotatebox{55}{\textit{2025--26}} \\
\midrule
Tay et al.\ (2022) & \texttimes & \texttimes & \texttimes & \texttimes & \texttimes \\
This survey         & \checkmark & \checkmark & \checkmark & \checkmark & \checkmark \\
\bottomrule
\end{tabular}
\end{table*}

But attention has a catch. The standard scaled dot-product mechanism computes pairwise similarities between every pair of tokens, giving $O(L^2)$ time and memory complexity in sequence length $L$. For 128K-token contexts, now supported by several production systems (\mbox{DeepSeek}-AI, 2024; Dubey et al., 2024), this quadratic cost can dominate inference budgets. At sufficiently long contexts, attention can become a substantial contributor to inference FLOPs with standard multi-head attention, although the fraction depends strongly on model architecture, including the number of layers, the feedforward expansion factor, and the Mixture-of-Experts configuration. The problem gets worse as context windows approach the million-token regime.

Efficiency is not the only problem. The same expressiveness that makes attention powerful also makes it opaque. A 70B-parameter Transformer has thousands of attention heads across dozens of layers (Dubey et al., 2024), and standard full-attention heads produce dense attention patterns over the entire context. Understanding what these heads learn, how they compose, and whether they implement interpretable algorithms is now a central question for both safety and science (Olsson et al., 2022; Elhage et al., 2022).

This survey addresses both problems together. Our argument is that the evolution of attention mechanisms cannot be understood separately from the efficiency innovations and interpretability insights that shaped the field. The threads interconnect: what attention heads learn tells us which tokens can be pruned, and efficiency constraints determine which attention patterns are feasible at scale.

We organize the survey around a single idea: the \textbf{Efficiency-Expressiveness-Interpretability (EEI) Framework}. It defines three axes. \emph{Efficiency} tracks compute and memory costs as sequence length and hardware change. \emph{Expressiveness} checks whether an attention variant can handle complex token interactions, including retrieval, copying, and long-range dependencies. \emph{Interpretability} rates how far attention patterns and head functions can be understood mechanistically. The central claim is simple: no surveyed method combines the strongest observed levels across all three EEI axes, and progress in attention research is driven by the tension between them. The EEI framework runs through the whole survey as a classification device.

Beyond the EEI framework itself, we contribute a taxonomy of efficient attention methods (seven efficiency/architecture families, 21 subcategories; Table~\ref{tab:deep-taxonomy}), a descriptive benchmark synthesis with explicit cross-study caveats (Table~\ref{tab:benchmarks}), an explicit research gap analysis (five open problems at the attention efficiency frontier; Section~\ref{explicit-research-gap-analysis}), and an evolution timeline covering 2015--2026 (Figure~\ref{fig:timeline}).

Section 2 defines the EEI framework: the scoring rubric, composite score, exploratory quantitative analysis, frontier pattern, and limitations. Section 3 covers attention foundations, from Bahdanau--Luong through scaled dot-product, multi-head design, positional encoding, and causal versus bidirectional architectures. Section 4 examines attention in computer vision: ViT, DETR, Swin, and vision-language cross-attention. Section 5 treats efficiency innovations, including fixed sparse patterns, learnable sparsity, linear attention, IO-aware exact attention, state-space alternatives, and the post-2025 adaptive frontier. Section 6 covers theoretical and interpretability results: mechanistic interpretability, expressiveness limits, superposition, and the attention-SSM duality. Section 7 covers open challenges, and the Conclusion summarizes the framework's contributions and its limitations.

\subsection*{Survey Methodology}

This survey follows a structured narrative review with PRISMA-inspired flow reporting (Moher et al., 2009), adapted for the dynamic and rapidly evolving machine learning literature. It was not preregistered, and the raw database exports, exact database-specific query strings, deduplication decisions, and paper-level screening log were not retained. The counts below are therefore an author-maintained reconstruction of the search process, not an independently reproducible systematic-review record.

We searched arXiv, IEEE Xplore, the ACL Anthology, and the ACM Digital Library for papers published between January 2015 and 31 May 2026, with foundational works predating this window cited for context. The 31 May 2026 cutoff governs survey scope, score construction, and synthesis; code- and checkpoint-availability statements are reported separately, as of 7 Aug 2026 (the reproducibility audit, Table~\ref{tab:reproducibility}). The two references must not be conflated: the literature cutoff is not a statement about code availability, and vice versa. The title-and-abstract keyword families covered attention, Transformers, efficient and sparse attention, state-space models, and mechanistic interpretability. Table~\ref{tab:methodology} summarizes the reconstructed search counts by source.

\begin{table}[!htbp]
\centering
\footnotesize
\renewcommand{\arraystretch}{1.2}
\caption{Author-maintained literature-search summary by source. Database counts are author-maintained approximate retrieval counts; records may overlap across databases (each row lists the number of records retrieved from that source). After deduplication, approximately 1,123 unique records remained for screening; counts are not independently auditable because the underlying export and deduplication logs were not retained. ``Source-level retained'' counts refer to approximately 94 papers that survived title/abstract screening; these per-source figures are not additive across rows because the same paper may be retained from multiple sources. The unique count of 94 was reconstructed after deduplication; of these, 63 were included in the final synthesis after full-text review.}
\label{tab:methodology}
\setlength{\tabcolsep}{4pt}
\begin{tabular}{@{}lrr@{}}
\toprule
Source & Retrieved & Source-level retained \\
\midrule
arXiv & $\approx$820 & $\approx$48 \\
IEEE Xplore & $\approx$140 & $\approx$15 \\
ACL Anthology & $\approx$210 & $\approx$22 \\
ACM Digital Library & $\approx$95 & $\approx$9 \\
\midrule
Total retrieved & $\approx$1{,}265 & --- \\
Unique after dedup. & $\approx$1{,}123 & $\approx$94 \\
Included after full-text review & -- & 63 \\
\bottomrule
\end{tabular}
\end{table}

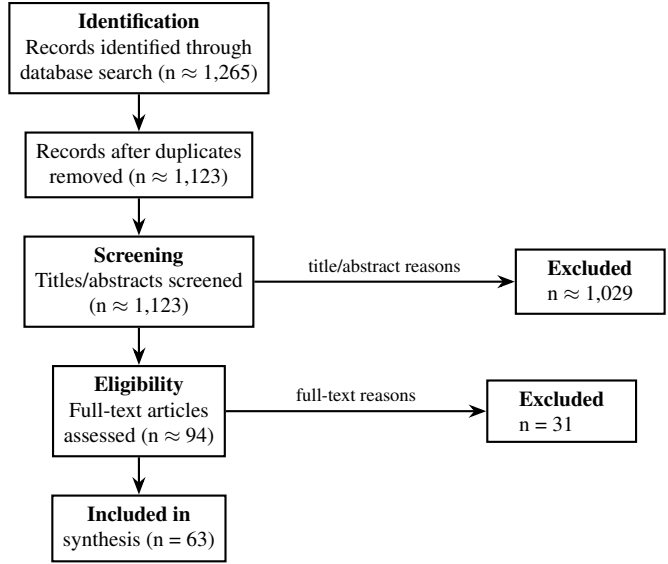
\begin{figure*}[t]
\centering
\begin{tikzpicture}[
node distance=1.4em and 1.6em,
flow/.style={draw, thick, rectangle, minimum width=5.6em, align=center, inner sep=4pt, font=\footnotesize},
excl/.style={flow, align=left, xshift=8.2em},
arr/.style={-{Stealth}, thick},
]
\node[flow] (id)  {\textbf{Identification}\\Records identified through\\database search (n $\approx$ 1,265)};
\node[flow, below=of id] (dups) {Records after duplicates\\removed (n $\approx$ 1,123)};
\node[flow, below=of dups] (scr) {\textbf{Screening}\\Titles/abstracts screened\\(n $\approx$ 1,123)};
\node[excl, right=of scr] (ex1) {\textbf{Excluded}\\n $\approx$ 1,029};
\node[flow, below=of scr] (ft) {\textbf{Eligibility}\\Full-text articles\\assessed (n $\approx$ 94)};
\node[excl, right=of ft] (ex2) {\textbf{Excluded}\\n = 31};
\node[flow, below=of ft] (inc) {\textbf{Included in}\\synthesis (n = 63)};
\draw[arr] (id.south) -- (dups.north);
\draw[arr] (dups.south) -- (scr.north);
\draw[arr] (scr.east) -- (ex1.west) node[midway, above, font=\scriptsize] {title/abstract reasons};
\draw[arr] (scr.south) -- (ft.north);
\draw[arr] (ft.east) -- (ex2.west) node[midway, above, font=\scriptsize] {full-text reasons};
\draw[arr] (ft.south) -- (inc.north);
\end{tikzpicture}
\caption{Literature flow diagram, inspired by the PRISMA 2009 reporting convention but using reconstructed author-maintained counts rather than an independently reproducible systematic-review record. Database search returned approximately 1,265 records; after duplicate removal, approximately 1,123 unique records were screened on title/abstract. Approximately 94 records retained after screening (Table~\ref{tab:methodology}) proceeded to full-text review, from which 31 were excluded, yielding 63 papers in the final synthesis. The screening log is unavailable, so the diagram documents scope rather than evidence of a fully reproducible systematic review.}
\label{fig:prisma}
\end{figure*}

The reconstructed search retrieved approximately 1,265 papers across the four databases. Figure~\ref{fig:prisma} summarizes the reconstructed screening flow from database retrieval through the final synthesis. After duplicate removal, approximately 1,123 unique papers were screened against the following inclusion criteria: (i) the paper introduces a novel attention mechanism, positional encoding, or efficiency innovation; (ii) the paper provides theoretical analysis or mechanistic interpretability results for Transformer architectures; or (iii) the paper presents a benchmark, survey, or empirical comparison of attention methods. Exclusion criteria eliminated papers focused exclusively on application domains without architectural novelty, papers superseded by later extended versions, and non-peer-reviewed technical reports except where they introduced widely adopted methods (e.g., FlashAttention, Mamba). After full-text review, 63 papers were retained for the final narrative synthesis; 21 methods were subsequently selected for the EEI scored panel. The 21-method panel was selected purposively to span six of the seven taxonomy families and to provide sufficient methodological diversity for rubric illustration; it is not a statistically representative sample of the 63-paper synthesis. The EEI panel is a design-space comparison across architectural levels (algorithms, attention variants, serving systems, alternative architectures), not a head-to-head performance ranking; methods are not assumed to be interchangeable competitors. Table~\ref{tab:eei-scores} groups the panel by architectural layer, and composite-score comparisons are accordingly most meaningful within a layer. Note that papers published in early 2026 (Focus, DashAttention) were available only as arXiv preprints at the time of the search; they are counted under the arXiv category in Table~\ref{tab:methodology} and their peer-review status is noted throughout the survey.

\section{The EEI Framework}\label{sec:eei-framework}

The EEI framework organizes attention research along three interacting axes. It is designed as a conceptual vocabulary and organizing heuristic, a language for describing trade-offs, rather than as a definitive ranking instrument. Each axis is defined through measurable properties rather than qualitative labels alone.

\textbf{Efficiency (E).} The efficiency score integrates three factors: (i) asymptotic time complexity as a function of sequence length $L$, (ii) constant-factor overhead including kernel launch costs and memory bandwidth utilization, and (iii) hardware compatibility, measured by whether the method can exploit tensor cores, SRAM tiling, or FP8 acceleration. Scores assume contemporary accelerator execution unless otherwise stated. A score of 1 corresponds to $O(L^2 d)$ without accelerator-aware optimization (e.g., a naive eager softmax implementation); a score of 10 corresponds to $O(L)$ with a demonstrated hardware-efficient implementation in a published evaluation (e.g., Mamba's $O(L)$ selective scan kernel on A100), rather than an exacting claim of full theoretical device utilization.

\textbf{Expressiveness (Ex).} Expressiveness judges whether an attention variant can represent complex token interactions. We consider four sub-dimensions: (i) retrieval accuracy on long-range dependency tasks and benchmarks such as Needle-in-a-Haystack and RULER (Hsieh et al., 2024); language-modeling quality on datasets such as PG-19 is treated as complementary evidence, (ii) ability to implement copying and induction heads, (iii) sensitivity to fine-grained positional distinctions, and (iv) capacity to represent full pairwise interactions versus constrained patterns. A score of 1 indicates a method that cannot model any token interactions; 10 indicates full-attention representational capacity with exact pairwise computation and no approximation or structural restriction. Throughout, we distinguish \emph{exact attention computation} (the softmax is computed exactly over its defined scope, with no score approximation) from the \emph{full-attention parameterization} (one key/value projection per head, i.e., the standard MHA setting): a method may satisfy the former without the latter, as in KV-sharing variants (GQA, MQA), which compute exact softmax but alter the parameterization and therefore do not receive the full-expressiveness designation.

\textbf{Interpretability (I).} Interpretability captures how amenable a mechanism is to mechanistic analysis. Sub-dimensions include: (i) whether attention weights are directly analyzable (softmax distributions) or mediated by recurrence/SSM dynamics, (ii) availability of open-source probing tools and SAEs for the architecture, (iii) existence of established theoretical frameworks (e.g., induction heads, the SSD duality) that explain model behavior, and (iv) empirical verification that individual heads or state dimensions correspond to interpretable features. A score of 10 is an aspirational anchor representing comprehensive causal characterization of head or state functions rather than a currently demonstrated empirical standard; no surveyed method currently satisfies this level. The I axis primarily reflects the availability of interpretability tooling and theoretical frameworks for each architecture class, rather than being a model-independent property of inherent interpretability; a method scoring low on I may be analyzable in principle but lack developed tools or established frameworks. Direct observability of attention weights should not be equated with faithful causal explanation: mechanistic studies have repeatedly shown that attention-weight patterns alone do not reliably identify the algorithms a model implements, so the I scores treat weight observability as one input among several rather than as evidence of causal understanding.

The EEI triplet $(E, \mathit{Ex}, I)$ for each method is assigned through the rubric in Table~\ref{tab:eei-rubric}. Figure~\ref{fig:eei-plot} visualizes these scores on a two-dimensional projection where Efficiency (x-axis) and Expressiveness (y-axis) define the position, and Interpretability (bubble size) captures the third dimension.

\begin{figure*}[!p]
\centering
\begin{minipage}{0.98\textwidth}
\centering
\sflt{\renewcommand{\baselinestretch}{0.84}\fontsize{12}{14.4}\selectfont}{\footnotesize}
\renewcommand{\arraystretch}{\sflt{0.7}{1.0}}

\begin{tabular}{p{0.8cm}p{3.2cm}p{3.2cm}p{3.2cm}}
\toprule
Score & Efficiency (E) & Expressiveness (Ex) & Interpretability (I) \\
\midrule
1--3 & $>O(L^2)$ without accelerator-aware optimization & Cannot model long-range retrieval & Opaque; no probing tools exist \\
4--5 & $O(L^2)$ with basic GPU kernel & Full-attention-level quality on short-context evaluations reported by the source, with limited evidence of long-range retrieval & Internal representations partially analyzable; limited component attribution \\
6--7 & Sub-quadratic ($O(L\sqrt{L})$ or $O(L\log L)$), $O(L^2)$ with IO-aware tiling, or substantial KV-cache/serving-memory reduction (e.g., KV sharing, paged/block-level KV management) & Near-baseline quality on evaluated long-context tasks, with some degradation on retrieval-sensitive evaluations & Components partially mappable; probing or SAE/state-analysis tools available for some models \\
8--9 & $O(L)$ or near-linear ($O(L\log L)$) with strong hardware utilization, $O(L^2)$ at near-peak IO-awareness, or demonstrated serving-level memory/throughput optimization & Quality comparable to the corresponding full-attention baseline on the source's evaluated long-context benchmarks, with only limited reported degradation & Established theoretical framework; components align with identifiable functional roles \\
10 & $O(L)$ with demonstrated hardware-efficient implementation in a published evaluation & Full-attention representational capacity with exact pairwise computation and no structural restriction & Every relevant component fully characterized with causal evidence \\
\bottomrule
\end{tabular}

\captionof{table}{EEI scoring rubric for each axis. Bands are qualitative anchors, not strict ceilings; IO-aware exact methods achieving near-peak hardware utilization (e.g., FlashAttention) receive high E despite $O(L^2)$ compute.}
\label{tab:eei-rubric}
\end{minipage}
\vspace{8pt}
\centering

\resizebox{\ifshrinkfloats 0.50\textwidth\else 0.80\textwidth\fi}{!}{
\definecolor{fullblue}{HTML}{377EB8}
\definecolor{iored}{HTML}{E41A1C}
\definecolor{sparsegreen}{HTML}{4DAF4A}
\definecolor{linearpurple}{HTML}{984EA3}
\definecolor{ssmorange}{HTML}{FF7F00}
\definecolor{adaptteal}{HTML}{009E73}
\definecolor{paretored}{HTML}{B2182B}

\newcommand{\scorebubble}[7]{%
  \pgfmathsetmacro{\bubbleradius}{0.15*sqrt(#3)}%
  \draw[gray!65, line width=0.35pt] (axis cs:#1,#2) -- (axis cs:#6,#7);
  \fill[#4, opacity=0.64] (axis cs:#1,#2) circle (\bubbleradius cm);
  \draw[#4!65!black, line width=0.7pt] (axis cs:#1,#2) circle (\bubbleradius cm);
  \begin{pgfonlayer}{axis foreground}
    \node[
      font=\scriptsize,
      align=center,
      fill=white,
      fill opacity=0.9,
      text opacity=1,
      inner sep=0.8pt,
      rounded corners=0.8pt
    ] at (axis cs:#6,#7) {#5};
  \end{pgfonlayer}
}

\begin{tikzpicture}
\begin{axis}[
  width=0.97\linewidth,
  trim axis left,
  height=0.61\linewidth,
  xlabel={Efficiency score (E) $\rightarrow$},
  ylabel={Expressiveness score (Ex) $\rightarrow$},
  xmin=3.7, xmax=10.65,
  ymin=5.35, ymax=10.65,
  xtick={4,5,6,7,8,9,10},
  ytick={6,7,8,9,10},
  grid=both,
  grid style={dashed, gray!15},
  tick label style={font=\small},
  label style={font=\normalsize},
  axis lines*=left,
  axis line style={line width=0.8pt},
  set layers=standard,
  legend style={
    at={(0.5,1.035)},
    anchor=south,
    legend columns=4,
    font=\scriptsize,
    draw=gray!45,
    fill=white,
    rounded corners=1pt,
    /tikz/every even column/.append style={column sep=5pt}
  }
]

\draw[gray!75, line width=1.5pt, -{Stealth[length=4pt]}]
  (axis cs:9.16,8.73) -- (axis cs:10.16,10.61);
\node[gray!75, font=\small, anchor=south east]
  at (axis cs:10.0,10.10) {Desired direction};

\scorebubble{4}{10}{6}{fullblue}{Transformer}{4.08}{10.36}

\scorebubble{7}{10}{6}{iored}{Ring Attention}{6.62}{10.36}
\scorebubble{8}{10}{6}{iored}{FlashAttention-2/3\\PagedAttention}{8.32}{10.42}
\scorebubble{6.82}{9.08}{6}{iored}{GQA}{6.5}{9.36}
\scorebubble{7.18}{7.92}{6}{iored}{MQA}{6.7}{7.58}

\scorebubble{7.18}{8.92}{7}{adaptteal}{Focus$^\dagger$}{7.35}{8.58}
\scorebubble{8}{9}{7}{adaptteal}{DashAttention$^\dagger$}{8.47}{9.2}

\scorebubble{6}{8}{6}{sparsegreen}{Routing Transformer}{5.43}{8.18}
\scorebubble{6.82}{8.08}{7}{sparsegreen}{BigBird}{6.48}{8.37}
\scorebubble{7}{7}{6}{sparsegreen}{Sparse Transformer}{6.48}{6.63}
\scorebubble{7.82}{7.08}{7}{sparsegreen}{Longformer}{7.48}{7.43}
\scorebubble{8}{8}{8}{sparsegreen}{Swin Transformer}{8.42}{8.27}
\scorebubble{6}{7}{5}{sparsegreen}{Reformer}{5.55}{7.18}

\scorebubble{7}{6}{5}{linearpurple}{Performer / Linformer}{7}{5.61}

\scorebubble{8.12}{6.92}{5}{ssmorange}{xLSTM / RWKV}{8.25}{6.55}
\scorebubble{10}{8}{6}{ssmorange}{Mamba-2 (SSD)}{9.42}{8.25}
\scorebubble{9}{7}{5}{ssmorange}{Hyena}{9}{6.61}
\scorebubble{10}{7}{5}{ssmorange}{Mamba}{10}{6.61}

\node[
  font=\scriptsize\itshape,
  text=gray!75!black,
  align=center,
  fill=white,
  fill opacity=0.9,
  text opacity=1,
  rounded corners=1pt,
  inner sep=1.2pt
] at (axis cs:4.55,9.18) {High Ex, low E:\\expressive but expensive};
\node[
  font=\scriptsize\itshape,
  text=gray!75!black,
  align=center,
  fill=white,
  fill opacity=0.9,
  text opacity=1,
  rounded corners=1pt,
  inner sep=1.2pt
] at (axis cs:9.55,5.72) {High E, lower Ex:\\efficient but less expressive};

\node[paretored, font=\scriptsize\bfseries, anchor=north east, fill=white, inner sep=0.5pt]
  at (axis cs:7.86,9.82) {P};
\node[paretored, font=\scriptsize\bfseries, anchor=north east, fill=white, inner sep=0.5pt]
  at (axis cs:9.82,7.78) {P};
\node[paretored, font=\scriptsize, anchor=west, fill=white, inner sep=1pt]
  at (axis cs:8.02,9.62) {P: E--Ex Pareto-optimal score pair};

\addlegendimage{only marks, mark=*, mark size=2.4pt, fullblue}
\addlegendentry{Full attention}
\addlegendimage{only marks, mark=*, mark size=2.4pt, sparsegreen}
\addlegendentry{Sparse / routed / window}
\addlegendimage{only marks, mark=*, mark size=2.4pt, linearpurple}
\addlegendentry{Linear / kernel}
\addlegendimage{only marks, mark=*, mark size=2.4pt, iored}
\addlegendentry{IO-aware / KV-efficient}
\addlegendimage{only marks, mark=*, mark size=2.4pt, ssmorange}
\addlegendentry{Structured sequence}
\addlegendimage{only marks, mark=*, mark size=2.4pt, adaptteal}
\addlegendentry{Adaptive}

\node[
  anchor=south west,
  align=left,
  font=\scriptsize,
  fill=white,
  fill opacity=0.93,
  text opacity=1,
  inner sep=2pt,
  rounded corners=1pt
] at (axis cs:3.78,5.39)
  {Bubble area $\propto I$ score; slight jitter separates identical $(E,Ex)$ scores.};

\end{axis}
\end{tikzpicture}}
\caption{The EEI (Efficiency--Expressiveness--Interpretability) score map. Each bubble plots a method by Efficiency ($x$-axis) and Expressiveness ($y$-axis), while bubble area encodes Interpretability. Colours identify the manuscript families: full attention, sparse/routed/window, linear/kernel, IO-aware/KV-efficient, structured sequence, and adaptive. Bubble centers reflect the assigned ordinal scores; small display-only offsets and leader lines separate overlapping methods. A red P marks each non-dominated Efficiency--Expressiveness score pair, $(8,10)$ and $(10,8)$, and the gray arrow indicates the desired direction for future work. Scores are rubric-derived ordinal judgments, not measured benchmark values.}
\label{fig:eei-plot}

\end{figure*}

Table~\ref{tab:eei-scores} lists the resulting E, Ex, and I assignments for the 21-method panel. The Family column uses panel-layer labels for readability; the formal seven-family taxonomy is defined separately in Table~\ref{tab:deep-taxonomy}.

\begin{table*}[!tbp]
\centering
\sflt{\fontsize{8.32}{10}\selectfont}{\footnotesize}
\renewcommand{\arraystretch}{\sflt{0.95}{1.1}}
\caption{EEI scores for representative attention methods. Scores are single-rater estimates with $\pm1$ assumed perturbation range; see Section~2 (Score Consistency and Limitations) for the perturbation sensitivity results. $^\dagger$Method is an arXiv preprint (mid-2026) without independent reproduction; see Section~7.7 and Table~\ref{tab:reproducibility}. Rows are grouped by architectural layer; the panel is a design-space comparison, so composite-score comparisons are most meaningful within a layer and illustrative across layers.}
\label{tab:eei-scores}
\begin{tabular}{p{2.5cm}p{1.2cm}p{1.5cm}p{1.5cm}c}
\toprule
Method & E & Ex & I & Family \\
\midrule
\multicolumn{5}{c}{\emph{Attention mechanisms}} \\
Transformer (Vaswani et al., 2017) & 4 & 10 & 6 & Full Attention \\
GQA (Ainslie et al., 2023) & 7 & 9 & 6 & Dense (KV-efficient) \\
MQA (Shazeer, 2019) & 7 & 8 & 6 & Dense (KV-efficient) \\
Longformer (Beltagy et al., 2020) & 8 & 7 & 7 & Sparse/Window \\
BigBird (Zaheer et al., 2020) & 7 & 8 & 7 & Sparse/Window \\
Sparse Transformer (Child et al., 2019) & 7 & 7 & 6 & Sparse/Window \\
Routing Transformer (Roy et al., 2021) & 6 & 8 & 6 & Sparse / Routing \\
Reformer (Kitaev et al., 2020) & 6 & 7 & 5 & Sparse / Hashing \\
Performer (Choromanski et al., 2021) & 7 & 6 & 5 & Linear/Kernel \\
Linformer (Wang et al., 2020) & 7 & 6 & 5 & Linear/Kernel \\
\midrule
\multicolumn{5}{c}{\emph{Attention systems and serving (kernels, KV-cache, parallelization)}} \\
FlashAttention-2/3 (Dao, 2024; Shah et al., 2024) & 8 & 10 & 6 & IO-Aware \\
PagedAttention (Kwon et al., 2023) & 8 & 10 & 6 & IO-Aware \\
Ring Attention (Liu et al., 2024a) & 7 & 10 & 6 & IO-Aware \\
\midrule
\multicolumn{5}{c}{\emph{Adaptive attention (learned token selection)}} \\
Focus (Yao et al., 2026)$^\dagger$ & 7 & 9 & 7 & Adaptive$^\dagger$ \\
DashAttention (Huang et al., 2026)$^\dagger$ & 8 & 9 & 7 & Adaptive$^\dagger$ \\
\midrule
\multicolumn{5}{c}{\emph{Alternative sequence operators (SSM, convolution, recurrent)}} \\
xLSTM (Beck et al., 2024) & 8 & 7 & 5 & Recurrent \\
Mamba (Gu \& Dao, 2024) & 10 & 7 & 5 & State-Space \\
Mamba-2 (SSD) (Dao \& Gu, 2024) & 10 & 8 & 6 & Structured SSM \\
Hyena (Poli et al., 2023) & 9 & 7 & 5 & Long Convolution \\
RWKV (Peng et al., 2023a) & 8 & 7 & 5 & Recurrent \\
\midrule
\multicolumn{5}{c}{\emph{Vision-specific architectures}} \\
Swin Transformer (Liu et al., 2021) & 8 & 8 & 8 & Sparse/Window \\
\bottomrule
\multicolumn{5}{p{10cm}}{\scriptsize Notable method not scored due to insufficient multi-benchmark data: \textbf{Multi-Head Latent Attention (MLA)} (DeepSeek-V2/V3, 2024--2025), a production KV-compression technique classified under Dense (KV-efficient). See Section~7.3 for qualitative discussion.}
\end{tabular}
\end{table*}

The EEI scores reveal a clear trade-off structure. Methods in the upper-left quadrant (high expressiveness, low efficiency), such as standard full attention, are the most expressive but scale poorly. Methods in the lower-right quadrant (high efficiency, lower expressiveness), such as Mamba and Performer, offer linear complexity; for Mamba the cited evidence for precise retrieval is limited, while for Performer approximation degrades sharp attention distributions, and both show reduced component-level interpretability. The adaptive sparsity methods (Focus, DashAttention) occupy a promising intermediate zone with competitive efficiency and near-full expressiveness, though at the cost of implementation complexity. The EEI framework makes explicit what individual benchmark numbers obscure: no surveyed method combines the strongest observed levels across all three EEI axes, and research progress consists of expanding the EEI frontier outward rather than finding a universally optimal point. (The frontier figure plots the E--Ex plane with I encoded by bubble size; the three-axis claim is established from the full score table rather than from the two-dimensional visualization alone.)

\paragraph{Score Derivation.}
We briefly justify the score assignment for each method to mitigate concerns about subjectivity.
\textit{Transformer (4,10,6):} Efficiency 4 because $O(L^2 d)$ with only basic GPU matrix multiplication and no IO-aware tiling; Expressiveness 10 as exact full attention is the reference standard; Interpretability 6 because attention weights are directly analyzable but mechanistic attribution of individual heads remains incomplete (Clark et al., 2019).
\textit{FlashAttention-2/3 (8,10,6):} Efficiency 8 because $O(L^2)$ compute is paired with memory linear in sequence length, with IO-aware SRAM tiling that reduces HBM accesses to $\Theta(L^2d^2/M)$, which is IO-optimal over the SRAM-size subrange for which the lower bound is established (Dao et al., 2022, Theorem 2) but remains quadratic in $L$; Expressiveness and Interpretability match full attention as the computation is mathematically identical.
\textit{GQA (7,9,6):} Efficiency 7 because KV cache is reduced by $h/g$ but $O(L^2)$ compute remains; Expressiveness 9 because shared KV heads incur minor representational loss (Ainslie et al., 2023).
\textit{MQA (7,8,6):} Efficiency 7 from full KV head sharing across all heads; Expressiveness 8 because the single shared KV projection causes slightly lower quality than MHA at matched model size (Shazeer, 2019); Interpretability same as GQA.
\textit{PagedAttention (8,10,6):} Efficiency 8 from block-level KV cache management that reduces memory fragmentation and enables larger effective batch sizes in vLLM; Expressiveness 10 because the attention computation is mathematically exact; Interpretability 6 matching full attention.
\textit{Ring Attention (7,10,6):} Efficiency 7 from distributed attention across multiple devices that enables processing of sequences longer than single-device memory, with KV-block communication fully overlapped with computation (Liu et al., 2024a); Expressiveness 10 as the computation is exact full attention; Interpretability 6 matching full attention.
\textit{Longformer (8,7,7):} Efficiency 8 through $O(Lw)$ window attention; Expressiveness 7 because local+global tokens preserve most quality but cross-window interactions are lost; Interpretability 7 because local windows simplify pattern analysis.
\textit{BigBird (7,8,7):} Efficiency 7 from $O(L)$ complexity with three attention components; Expressiveness 8 due to theoretical Turing-completeness guarantees (Zaheer et al., 2020); Interpretability 7 from its explicit theoretical structure.
\textit{Sparse Transformer (7,7,6):} Efficiency 7 from $O(L\sqrt{L})$ strided patterns; Expressiveness 7 because fixed patterns cannot adapt to content.
\textit{Routing Transformer (6,8,6):} Efficiency 6 because $k$-means clustering overhead per layer offsets the $O(L^{1.5} d)$ routing advantage (with balanced clusters $k=\Theta(\sqrt{L})$), making it slower than full attention at moderate lengths, though the source's wall-clock comparisons cover regimes (e.g., 8192-token PG-19) where full attention is infeasible, so a precise crossover is not established (the E=6 assignment may be generous, as clustering overhead can negate the complexity benefit at practical lengths, and practitioners should verify routing overhead against their specific sequence length and hardware before adopting Routing Transformer). Expressiveness 8 because content-based routing adapts to input; Interpretability 6 because cluster assignments provide some structure.
\textit{Reformer (6,7,5):} Efficiency 6 because $O(L \log L)$ is offset by high hashing overhead; Expressiveness 7 as LSH preserves approximate quality; Interpretability 5 because hash routing is opaque (Kitaev et al., 2020).
\textit{Performer (7,6,5):} Efficiency 7 via $O(L)$ kernel approximation; Expressiveness 6 because FAVOR+ approximation degrades on long-range retrieval; Interpretability 5 because the randomized feature map does not produce analyzable attention weights (Choromanski et al., 2021).
\textit{Linformer (7,6,5):} Efficiency 7 from low-rank projection of keys and values to $O(Lk)$ with $k \ll L$; Expressiveness 6 because the low-rank bottleneck can discard fine-grained positional information needed for retrieval (Wang et al., 2020); Interpretability 5 because the projection obscures direct token-token relationships.
\textit{xLSTM (8,7,5):} Efficiency 8 from $O(L)$ recurrent inference with modified LSTM gating, though sLSTM's sequential memory mixing limits parallelization; the authors' custom CUDA kernels bring sLSTM to within 2$\times$ of mLSTM throughput (Beck et al., 2024); Expressiveness 7 because the expanded memory (matrix memory via mLSTM) improves on classic LSTMs, though the cited work does not provide a directly comparable long-context retrieval evaluation, so the Ex score conservatively reflects the limited evidence for precise arbitrary-token retrieval (Beck et al., 2024); Interpretability 5 due to recurrent hidden state opacity.
\textit{Mamba (10,7,5):} Efficiency 10 through $O(L)$ recurrence with a custom GPU scan kernel; Expressiveness 7 because competitive perplexity is accompanied by limited direct retrieval evidence on long-range tasks; Interpretability 5 because SSM hidden states lack attention-weight-style analysis (Gu \& Dao, 2024).
\textit{Mamba-2 (SSD) (10,8,6):} Efficiency 10 from $O(L)$ structured algorithms with tensor-core-friendly matrix multiplication and reported utilization gains over Mamba-1; the rubric's E=10 band is an aspirational maximum for linear-time, highly optimized implementations, and Mamba-2's improvement over Mamba cannot exceed it. Expressiveness 8 reflects the structured-attention formulation and reported quality, not a general theorem of downstream-task expressiveness; Interpretability 6 because the duality offers a formal bridge to attention-like matrix structure without yet supplying head-level causal tools.
\textit{Hyena (9,7,5):} Efficiency 9 from implicit free-form long convolutions with data-controlled gating, achieving $O(L \log L)$ with hardware-efficient FFT; Expressiveness 7 reflects its global sequence processing and the limited direct evidence for precise long-range retrieval; the cited work does not provide a directly comparable retrieval evaluation; Interpretability 5 because the implicit convolution representation is even less analyzable than explicit SSM states (Poli et al., 2023).
\textit{RWKV (8,7,5):} Efficiency 8 from $O(L)$ recurrence through the WKV (weighted key-value) mechanism, implemented in PyTorch with DeepSpeed-inspired optimizations; Expressiveness 7 similar to Mamba on retrieval tasks; Interpretability 5 because the time-mixed recurrence is not directly analyzable as attention weights (Peng et al., 2023a).
\textit{Focus (7,9,7):} Efficiency 7 from learned grouping that achieves $2\times$ speedup; Expressiveness 9 because exact attention is preserved within learned groups; Interpretability 7 because centroids correspond to semantic groupings (Yao et al., 2026).
\textit{DashAttention (8,9,7):} Efficiency 8 from two-stage routing achieving up to $3.3\times$ kernel-level speedup over FlashAttention-3 in the paper's decoding benchmarks; Expressiveness 9 because $\alpha$-entmax adaptively allocates capacity; Interpretability 7 because the entmax distribution is differentiable and analyzable (Huang et al., 2026).
\textit{Swin Transformer (8,8,8):} Efficiency 8 from $O(H_pW_pM^2d)$ windowed attention; Expressiveness 8 as shifted windows enable cross-window connections; Interpretability 8 because spatial locality provides spatial structural interpretability for vision tasks (Liu et al., 2021).

\paragraph{Exploratory Quantitative Analysis.}
We distinguish two types of evidence below: \textit{internal consistency checks}, which confirm that the scoring rubric was applied coherently, and \textit{exploratory benchmark concordance analyses}, which compare EEI scores with reported benchmark results while acknowledging partial overlap between benchmark evidence and score construction. Table~\ref{tab:eei-validation} compares EEI dimensions against two benchmark metrics: Long Range Arena (LRA) average scores and published throughput speedups. We report one Spearman correlation, between Expressiveness and LRA; throughput speedups are compared descriptively only, because they span kernel-, model-, and serving-engine scopes with different baselines, hardware, sequence lengths, and precisions, and no reproducible rule currently exists for selecting one representative value per method (for example, Focus reports both 2.0$\times$ end-to-end and 8.6$\times$ at 1M tokens; see Table~\ref{tab:benchmarks}).

Expressiveness (Ex) and LRA average scores produce $\rho = 0.54$ ($p = 0.21$, $n = 7$) for the seven methods with LRA results under the official protocol of Tay et al.\ (2021); adding RWKV's self-reported 72.07 (a non-official-protocol evaluation) yields $\rho = 0.45$ ($p = 0.27$, $n = 8$). This correlation is positive but not statistically significant at conventional levels. It is worth reporting for two reasons: Ex scoring drew only partially on LRA averages, as four of the seven methods in this correlation sample (Longformer, BigBird, Reformer, Performer) cite LRA averages directly in their Ex justifications (Appendix~\ref{app:eei-justification}), while Sparse Transformer, Linformer, and the Transformer baseline rest on structural arguments, so the correlation is not purely an artifact of score construction, though the partial overlap means it should be read as suggestive rather than fully independent; and the small samples ($n = 7$--$8$) and the narrow LRA range (50.7--55.0 for all methods except RWKV's 72.07) leave any monotone relationship poorly identified. We therefore treat the Ex--LRA association as suggestive but unconfirmed rather than as validation.

For the Efficiency axis we do not report an inferential correlation. The published speedups in Table~\ref{tab:speedups} span kernel-level, engine-level, and model-level scopes (e.g., serving-engine throughput for PagedAttention, kernel throughput for DashAttention, end-to-end decoding for Focus), were measured on different hardware (A100, V100, H100) and sequence lengths, and several methods report multiple non-equivalent candidate values (Focus: 2.0$\times$ end-to-end versus 8.6$\times$ at 1M tokens; Mamba: up to 3$\times$ versus 4--5$\times$; Hyena: 5$\times$ versus 2$\times$; Linformer: 1.5--1.6$\times$). Without a prospective rule for selecting one observation per method, a coefficient computed from these values would reflect discretionary choices rather than measurement, so we restrict the quantitative synthesis to descriptive tabular comparison and treat the E score assignments as rubric-based judgments to be validated prospectively on a common protocol (Section~\ref{open-challenges-and-future-directions}).

We do not attempt to quantify a relationship between interpretability (I) scores and the benchmark values we examined, as interpretability is not directly measured by throughput or retrieval metrics; this gap shows the need for standardized interpretability benchmarks, which we identify as an open problem in Section~7. True independent validation of the EEI framework would require held-out benchmarks not consulted during score assignment, for example RULER (Hsieh et al., 2024) or downstream task performance on HELMET (Yen et al., 2025), applied prospectively to new methods. We identify this as a priority for future EEI evaluation. Because the benchmark values in Table~\ref{tab:eei-validation} were synthesized from multiple studies using different datasets, hardware, and evaluation protocols, and because quality-relative terms indicate source-reported comparability rather than a universal numerical tolerance, the correlation should be interpreted as indicative rather than causal. All $p$-values are uncorrected for multiplicity; borderline results should be treated as weak evidence. The per-score numeric anchors behind the Appendix~A justifications are recorded in the accompanying scoring worksheet (verification-spreadsheet.csv, supplementary material), making all $21 \times 3 = 63$ axis assignments auditable. The Interpretability score is assigned per the rubric in Table~\ref{tab:eei-rubric}; the four interpretability dimensions (structural observability, tool availability, formal characterization, causal validation) are qualitative descriptors intended to guide rubric assignment rather than independent numerical inputs, and they are collected in the rubric rather than in the scoring worksheet. A future multi-rater study could score the four dimensions independently and calibrate their aggregation empirically.

\begin{table*}[!tbp]
\centering
\caption{EEI scores versus empirical benchmark metrics. LRA Avg. values are from Table~\ref{tab:benchmarks} where available (``---'' indicates no published value; otherwise the value is reported directly in this table); speedup values are from Table~\ref{tab:speedups} where available.}
\label{tab:eei-validation}
\resizebox{0.60\textwidth}{!}{%
\sflt{\fontsize{11.5}{13.8}\selectfont}{\footnotesize}
\renewcommand{\arraystretch}{\sflt{2.93}{2.42}}
\begin{tabular}{p{2.5cm}c c c c c}
\toprule
Method & E & Ex & I & LRA Avg. & Speedup \\
\midrule
Transformer & 4 & 10 & 6 & 54.39 & 1.0$\times$ \\
FlashAttention-2/3 & 8 & 10 & 6 & --- & $\sim2.0\times$\textsuperscript{*} \\
GQA & 7 & 9 & 6 & --- & --- \\
MQA & 7 & 8 & 6 & --- & --- \\
PagedAttention & 8 & 10 & 6 & --- & --- \\
Ring Attention & 7 & 10 & 6 & --- & --- \\
Longformer & 8 & 7 & 7 & 53.46 & --- \\
BigBird & 7 & 8 & 7 & 55.01 & --- \\
Sparse Transformer & 7 & 7 & 6 & 51.24 & --- \\
Swin Transformer & 8 & 8 & 8 & --- & --- \\
Routing Transformer & 6 & 8 & 6 & --- & --- \\
Reformer & 6 & 7 & 5 & 50.67 & --- \\
Performer & 7 & 6 & 5 & 51.41 & --- \\
Linformer & 7 & 6 & 5 & 51.36 & 1.5$\times$ \\
xLSTM & 8 & 7 & 5 & --- & --- \\
Mamba & 10 & 7 & 5 & --- & up to 3$\times$ \\
Mamba-2 (SSD) & 10 & 8 & 6 & --- & 2--8$\times$ vs.\ Mamba-1 \\
Hyena & 9 & 7 & 5 & --- & 5$\times$ @ 8K \\
RWKV & 8 & 7 & 5 & 72.07$^\S$ & --- \\
Focus$^\dagger$ & 7 & 9 & 7 & --- & 2.0$\times$ \\
DashAttention$^\dagger$ & 8 & 9 & 7 & --- & up to 3.3$\times$ (kernel)\textsuperscript{$\dagger$} \\
\bottomrule
\multicolumn{6}{p{10cm}}{\scriptsize $^\dagger$Focus and DashAttention are arXiv preprints (mid-2026) without independent reproduction; benchmark values are from the original papers and have not been independently verified.}\\
\multicolumn{6}{p{10cm}}{\scriptsize $^*$Speedup shown for FlashAttention-2 versus FlashAttention-1; FlashAttention-3 is separately reported in Table~\ref{tab:speedups}. Speedups use different baselines and hardware per method (see Table~\ref{tab:speedups}) and are not directly comparable across rows.}\\
\multicolumn{6}{p{10cm}}{\scriptsize $^\S$RWKV reports its own LRA evaluation (72.07 excluding Path-X, Peng et al., 2023a, Table 4); all other LRA values follow the official protocol of Tay et al., 2021.}\\
\end{tabular}%
}%
\end{table*}

\paragraph{Formal EEI Score.}
To make the framework reproducible, we define an exploratory weighted EEI index for any attention method as a weighted linear combination:
\begin{equation}
\begin{aligned}
\mathrm{EEI}(w_E, w_{Ex}, w_I)
  &= w_E\,E + w_{Ex}\,\mathit{Ex} + w_I\,I,\\
w_E + w_{Ex} + w_I &= 1.
\end{aligned}
\end{equation}
where $E, \mathit{Ex}, I \in [1,10]$ are the axis scores assigned via the rubric in Table~\ref{tab:eei-rubric}. The numerical scales are ordinal within each axis; equal numerical values across axes do not imply commensurate quantities. Because the axes are ordinal, the weighted composite is an exploratory index rather than a statistically validated cardinal score; the weights reflect the relative importance of each dimension for a given deployment scenario.

We consider three canonical weight configurations to illustrate how the rubric-illustrative ranking of methods changes under different priorities. \textit{Equal weights} ($w_E = w_{Ex} = w_I = \frac{1}{3}$) assumes no preference. \textit{Retrieval-focused} weights ($w_E = 0.2, w_{Ex} = 0.6, w_I = 0.2$) prioritize expressiveness for long-context retrieval tasks such as Needle-in-a-Haystack. \textit{Deployment-focused} weights ($w_E = 0.6, w_{Ex} = 0.2, w_I = 0.2$) prioritize throughput and memory efficiency for production serving. Under equal weights, five methods tie at the top (FlashAttention-2/3, PagedAttention, DashAttention, Swin Transformer, and Mamba-2, EEI $= 8.0$), while Performer, Reformer, and Linformer tie at the bottom (EEI $= 6.0$). Under retrieval-focused weights, FlashAttention and PagedAttention lead (EEI $= 8.8$), followed by Ring Attention (EEI $= 8.6$), DashAttention (EEI $= 8.4$), and Focus (EEI $= 8.2$), reflecting the premium on exact attention for retrieval tasks. Under deployment-focused weights, Mamba-2 leads (EEI $= 8.8$), followed by Mamba (EEI $= 8.4$); FlashAttention-2/3, PagedAttention, DashAttention, and Swin Transformer tie at 8.0. This sensitivity analysis confirms that the EEI framework captures meaningful trade-offs: no method dominates across the three illustrative weight configurations, and the ranking shifts in interpretable ways that reflect each method's strengths.

The linear composite formula treats the three axes as partially substitutable: a method with $(E=10, \mathit{Ex}=5, I=5)$ scores the same under equal weights as one with $(E=6, \mathit{Ex}=8, I=6)$, though these represent very different architectural profiles. For deployment scenarios where a minimum threshold on each axis is required, for example any method used for retrieval must have $\mathit{Ex} \geq 8$ regardless of efficiency, a minimum-operator variant $\text{EEI}_{\min} = \min(E/\tau_E, \mathit{Ex}/\tau_{Ex}, I/\tau_I)$ for declared target levels $\tau_E, \tau_{Ex}, \tau_I$ would provide a complementary ranking that penalizes methods failing to meet any minimum acceptable threshold. A multiplicative formulation $\text{EEI}_{\times} = \prod_i \mathrm{score}_i^{w_i}$ with $\sum_i w_i = 1$ (a weighted geometric mean on the declared 1--10 scale) would similarly penalize imbalance. We retain the linear formulation as the default because it best reflects the typical practitioner scenario of trading off between axes, and provide the weight-sensitivity analysis in Table~\ref{tab:eei_sensitivity} as a robustness check. The choice of functional form is ultimately application-dependent, and we encourage EEI users to select the aggregation that matches their deployment constraints.

\paragraph{Score Consistency and Limitations.}
The EEI scores in Table~\ref{tab:eei-scores} are literature-derived assessments assigned using the rubric in Table~\ref{tab:eei-rubric}, drawing on published benchmark results, complexity analysis, and mechanistic interpretability findings from the cited literature. They are intended as structured evidence-based assessments rather than definitive measurements, reflecting the current absence of standardized quantitative metrics for every EEI dimension. These scores are single-rater estimates that carry an assumed $\pm1$ perturbation range: two raters could disagree on whether FlashAttention's efficiency merits an 8 or a 9 depending on how they weight asymptotic complexity versus practical speedup.

To quantify this ranking instability, we execute the following perturbation sensitivity protocol: independently perturb each axis score for each method by sampling uniformly from $\{\mathit{score}-1, \mathit{score}, \mathit{score}+1\}$ and clipping the perturbed values to the rubric domain $[1,10]$ (boundary scores such as $E=10$ or $\mathit{Ex}=10$ therefore draw from a reduced effective window), recompute composite EEI rankings under all three weight configurations using a single fixed tie convention (descending average ranks, in which tied methods receive the mean of the ranks they occupy; scipy.stats.rankdata with method='average'), and report the fraction of 200\,000 perturbation samples in which each method's rank changes by more than one position (Table~\ref{tab:eei_sensitivity}). Sampling uses NumPy's PCG64 generator with seed 42, pinned in the regenerable artifact (Scripts/eei\_sensitivity.py). This protocol assumes the per-axis perturbations are independent across methods and axes and that the uniform $\pm1$ band adequately captures single-rater assumed perturbation range; both assumptions are simplifying, and a multi-rater study with correlated disagreement would require a richer noise model.

The results reveal that rank instability is substantial across the score table. Under equal weights, the mean fraction of samples in which a method's rank changes by more than one position is 67\% (cross-method standard deviation of 13 percentage points), and in essentially every sample at least one method changes rank by more than one position. The tied five-way score of 8.00 under equal weights is broken in nearly every sample, but the tied-top methods show comparatively moderate instability (65--66\% rank-change fraction) because average-rank tie handling absorbs co-movement: perturbations that shift all five methods together preserve the tie group. The most stable methods are those at the extremes of the score distribution: under equal weights, Performer, Linformer, and Reformer (identical composite score 6.00, at the bottom of the table) show a 37\% rank-change fraction, while Transformer under deployment weights (26\%) is stable because its low E score (4) anchors its position regardless of $\pm1$ variation. Mid-tier methods (scores 7.0--7.7) show instability of roughly 65--80\%, reflecting tight clustering across all weight configurations.

The inverted-U stability profile (stable extreme ranks, unstable mid ranks) is reproduced by a rank-matched null model (Table~\ref{tab:eei_sensitivity}): when score triplets are drawn at random from the distinct values observed on each axis and subjected to the identical perturbation and ranking pipeline, the expected instability rises from 33--36\% at rank 1 to 70--78\% at mid ranks and falls to 28--29\% at the absolute bottom rank (rank 21), closely tracking the observed profile. Rank positions are bounded, so a top-ranked method cannot move upward and a bottom-ranked method cannot move downward, and most of the apparent stability at the extremes is a property of rank censoring rather than of the specific score configuration. The observed scores add two features beyond the null: dense ties among the observed composites push the top and middle tiers \emph{above} the null expectation (mean excess $+5$, $+3$, and $+10$~pp under equal, retrieval, and deployment weights, respectively), because tie groups flip ranks whenever any member is perturbed; and the bottom tier sits at or slightly below its null expectation. For the observed three-way tie occupying ranks 19--21, the expectation at the mean rank position, rank 20, is 38\% under equal weights, consistent with bottom-rank censoring. Neither feature supports claiming reliable differentiation at the extremes.

These results carry an important implication for EEI users: the perturbation analysis is a sensitivity check, not a confidence procedure, and it does not demonstrate reliable differentiation at the extremes; the apparent stability of the extreme ranks is reproduced by a rank-matched null model and therefore reflects rank censoring rather than score reliability, while dense ties push the top and middle tiers \emph{above} the null expectation. Concretely, any fine-grained claim that ``method A outperforms method B'' within the composite score range 7.0--7.7 (encompassing 8 of the 21 surveyed methods) cannot be supported by the current single-rater scores; only coarse tier-level groupings (top, mid, bottom), whose member scores differ by at least one composite point, survive the perturbation analysis. We recommend that future multi-rater studies focus on the mid-tier cluster, where inter-rater disagreements would have the largest impact on rankings, and that reported EEI composites for mid-tier methods report sensitivity frequencies under the assumed perturbation model. Each score is justified in detail in Appendix~\ref{app:eei-justification}, allowing readers to independently assess or contest individual assignments.

\textbf{Methodological disclosures.} The implementation ranks integer-weighted sums, $(1,1,1)$, $(1,3,1)$, and $(3,1,1)$, mathematically equivalent to the normalized weights above, to avoid floating-point artifacts that would split mathematical ties. Ties are handled with average ranks throughout; alternative conventions shift the aggregate results materially (mean rank-change fractions of 65.5/64.8/65.9\% under min ranks, 64.5/63.8/66.2\% under ordinal ranks, and 45.8/50.9/50.8\% under dense ranks, for equal/retrieval/deployment weighting), so the statistic should be read as an ordinal sensitivity indicator rather than a precise measurement. Perturbations are clipped to the rubric domain $[1,10]$; clipping pulls boundary scores inward (a score of 10 perturbed by $\{-1,0,1\}$ has mean 9.67 rather than 10), slightly deflating the apparent instability of boundary cells. The observed panel also occupies only part of the rubric: E takes values 4--10, Ex 6--10, and I 5--8, so the scores are coarse ordinal bands, not cardinal measures. Because the axes have different between-method variances (E: 1.68, Ex: 1.62, I: 0.71), the nominal weights do not equal the effective variance contributions: under equal weights, E, Ex, and I contribute approximately 42\%, 40\%, and 18\% of composite variance; under retrieval weights, Ex contributes 86\%; under deployment weights, E contributes 87\%.

\begin{table}[t]
\centering
\scriptsize
\renewcommand{\arraystretch}{1.08}
\caption{Monte Carlo score-perturbation sensitivity: fraction of samples in which a method's average-tie rank changes by more than one position, rounded to whole percentage points (Monte Carlo standard error per cell $\approx 0.1$~pp). \textit{Null expectation (rank-matched):} mean instability under the same perturbation process applied to 100 replicates of score triplets drawn from the distinct values observed on each axis (E: [4, 6, 7, 8, 9, 10]; Ex: [6, 7, 8, 9, 10]; I: [5, 6, 7, 8]), (100 reps, 50000 samples each, seed 43), averaged at each method's baseline rank. \textit{Excess:} observed mean minus null expectation; values near zero indicate that rank instability follows from perturbation magnitude and rank censoring rather than the specific score configuration.}
\label{tab:eei_sensitivity}
\begin{tabular}{lccc}
\toprule
Method & Equal & Retrieval & Deployment \\
\midrule
Transformer & 69\% & 66\% & 26\% \\
FlashAttention-2/3 & 65\% & 51\% & 77\% \\
GQA & 78\% & 80\% & 82\% \\
MQA & 77\% & 81\% & 79\% \\
PagedAttention & 65\% & 51\% & 77\% \\
Ring Attention & 74\% & 57\% & 82\% \\
Longformer & 78\% & 79\% & 82\% \\
BigBird & 78\% & 81\% & 82\% \\
Sparse Transformer & 73\% & 73\% & 76\% \\
Routing Transformer & 73\% & 79\% & 65\% \\
Reformer & 37\% & 59\% & 40\% \\
Performer & 37\% & 37\% & 65\% \\
Linformer & 37\% & 37\% & 65\% \\
xLSTM & 73\% & 73\% & 82\% \\
Mamba & 75\% & 79\% & 59\% \\
Mamba-2 (SSD) & 65\% & 80\% & 35\% \\
Hyena & 77\% & 77\% & 81\% \\
RWKV & 73\% & 73\% & 82\% \\
Focus & 77\% & 77\% & 83\% \\
DashAttention & 66\% & 76\% & 77\% \\
Swin Transformer & 66\% & 80\% & 77\% \\
\midrule
Mean & 67\% & 69\% & 70\% \\
Null expectation (rank-matched) & 62\% & 66\% & 60\% \\
Excess over null (pp) & +5 pp & +3 pp & +10 pp \\
Pop. Std. Dev. & 13\% & 14\% & 17\% \\
\bottomrule
\end{tabular}
\end{table}

\paragraph{Proposed EEI Reporting Protocol.}
A key contribution of this survey is to propose the EEI framework not merely as an analytical lens but as a provisional evaluation convention for future attention research. We define the \textbf{proposed EEI reporting protocol} (explicitly single-rater; it requires independent validation before it can serve as a standardized benchmark) as follows:

\noindent\textit{Efficiency axis.} Report wall-clock throughput (tokens/second) on at least two GPU generations (e.g., A100-80GB and H100-80GB) at sequence lengths $L \in \{2K, 8K, 32K, 128K\}$ with batch size 1. If the full-attention baseline cannot execute at a specified length within device memory, report OOM rather than substituting a different baseline; OOM itself is meaningful quality-efficiency evidence. Normalize by the throughput of a specified exact full-attention implementation under identical hardware, precision, model dimensions, batch size, causal mask, and software stack; the implementation and version must be reported. The conventional framework baseline must specify the exact operator and backend (e.g., \texttt{torch.nn.functional.scaled\-dot\-product\-attention} with the selected backend explicitly enabled or disabled) and the PyTorch/CUDA versions; the optimized exact-attention baseline must specify the kernel and its version (e.g., FlashAttention-2 or FlashAttention-3 with exact version and commit). Because FlashAttention and related kernels are themselves exact full-attention implementations, we require both a conventional framework baseline and an optimized exact-attention baseline so that approximations can be compared against both ends of the exact-attention implementation spectrum.

\noindent\textit{Expressiveness axis.} Report Needle-in-a-Haystack retrieval accuracy at $L = 32K$ and $L = 128K$, language modeling perplexity on WikiText-103 or an equivalent held-out corpus at $L = 8K$, and the Long Range Arena (LRA) average score (Tay et al., 2021). These metrics are complementary rather than interchangeable: NIAH probes targeted retrieval, language-modeling perplexity probes predictive sequence modeling, and LRA provides a heterogeneous long-range task suite; a method may perform very differently across them, so Ex should not be interpreted as a single latent quantity unless future validation establishes such a relationship.

\noindent\textit{Interpretability axis.} Report interpretability evidence as four separate fields that must not be collapsed into a single fraction unless a validation study establishes how they should be weighted: \textit{I1, component-level empirical interpretability}, the fraction of attention heads or SSM state dimensions, with separate reporting conventions for attention-head-level and state-level interpretability, for which a clear linguistic or functional role can be identified using established probing methods (Clark et al., 2019) or sparse autoencoder-based analysis (Bricken et al., 2023); \textit{I2, tooling availability}, the availability of open-source interpretability tools for the architecture; \textit{I3, formal theoretical characterization}, whether the mechanism has been analyzed under a formal theoretical framework (e.g., the SSD duality for attention-SSM hybrids); and \textit{I4, causal validation}, whether interventions on components have been causally linked to model behavior. A method with zero analyzed heads but excellent tooling should be reported as high on I2 and low on I1, not as a single intermediate fraction.

Adopting the proposed EEI reporting protocol would facilitate more direct comparability; it is a candidate convention, not yet a community standard across future attention research, and would provide a unified reporting convention for efficiency evaluation rather than a normative standard. This addresses the reproducibility gap identified in Section~7.7.

\paragraph{EEI Frontier Pattern.}
We emphasize at the outset that the following pattern is a descriptive summary of how the EEI rubric was defined and applied, not an empirical discovery about the attention design space. It should be read as an internal consistency check on the scoring procedure, not as a predictive law that constrains future architectures. The EEI scatter plot (Figure~\ref{fig:eei-plot}) reveals an observed pattern that we characterize as the \textit{EEI Frontier Pattern}: for exact attention methods, efficiency improvements beyond a threshold of $E \geq 6$ without approximation typically leave expressiveness unchanged (horizontal rightward movement); for approximate methods, efficiency gains beyond $E \geq 6$ are accompanied by expressiveness degradation under the rubric's scoring (a rubric-level expectation given the definition of the axes, not a cross-study measured finding) unless the approximation is input-adaptive.

We emphasize the epistemic status of this pattern at the outset: it is a descriptive characterization of how the EEI rubric was defined and applied, not an independently verified empirical law that constrains future architectures. The pattern follows in part from how Expressiveness is scored: approximation methods are penalized on the Ex axis by construction, and exact methods are not, so observing that no exact method has low Ex or that no approximation method has high Ex confirms the rubric's internal consistency rather than testing an external hypothesis. Whether new approximation methods can break the observed bound is an open empirical question that the framework itself cannot answer. With this caveat established, we examine the pattern across the surveyed methods.

The pattern is supported by three observations. \textit{Observation 1: Exact-computation IO-aware methods (FlashAttention, PagedAttention, Ring Attention) achieve $Ex = 9\text{--}10$; KV-sharing variants (GQA, MQA) trade a small, measured quality cost for memory savings ($Ex = 9$ and 8, respectively).} The full-attention IO-aware methods maintain exact computation (FlashAttention's tiling does not approximate the attention matrix) and therefore preserve full-attention expressiveness; GQA/MQA also compute exact softmax attention but alter the KV parameterization and therefore do not receive the same full-expressiveness designation. Within the EEI rubric, this appears as a Pareto improvement for exact full-attention implementations, in which efficiency rises while expressiveness remains constant relative to the exact baseline under the assigned ordinal scores; it is a score-level, not a measured empirical, Pareto claim. \textit{Observation 2: Fixed sparse patterns (Longformer, BigBird) achieve $E = 7\text{--}8$ but $Ex = 7\text{--}8$, a loss of 2--3 points.} The sparsity pattern is static and input-independent; tokens outside the pattern are simply ignored, which creates expressiveness blind spots. Under the rubric, the efficiency gain is scored against an expressiveness sacrifice in this static-pattern case. \textit{Observation 3: Adaptive sparsity methods (Focus, DashAttention) achieve $E = 7\text{--}8$ with $Ex = 9$, recovering nearly all expressiveness.} The key difference is that the sparsity is learned and input-dependent: Focus groups tokens semantically via centroids (Yao et al., 2026), DashAttention uses $\alpha$-entmax to adaptively allocate capacity (Huang et al., 2026). The expressiveness recovery is consistent with the mechanisms' design: the learned routing preserves the exact attention computation for the subset of tokens that matter for the current input; causal confirmation would require independent evaluation under a common benchmark.

The rubric's definitional structure produces two consistency checks that confirm the scoring logic is internally coherent rather than testing external hypotheses about the design space. \textit{Consistency Check 1: Within the current rubric, methods that reach $E \geq 7$ through score or kernel approximation receive at most $Ex = 6$, methods that reach $E \geq 7$ through fixed support selection receive at most $Ex = 8$, while adaptive support selection and structurally exact alternatives can receive higher $Ex$ scores.} Performer ($E=7, Ex=6$) and Linformer ($E=7, Ex=6$) illustrate the approximation bound; Reformer ($E=6, Ex=7$) falls below the $E \geq 7$ threshold of the bound. BigBird ($E=7, Ex=8$) is sparse selection rather than approximation, and Mamba ($E=10, Ex=7$) and Mamba-2 ($E=10, Ex=8$) are alternative sequence operators whose scores reflect their exact structured formulations rather than approximation bounds, so none of them is an exception to an approximation-based constraint. \textit{Consistency Check 2: Adaptive selection methods occupy the high-Ex region while retaining substantial efficiency.} DashAttention, for example, is assigned $(E=8, Ex=9)$ and reports up to $3.3\times$ kernel-level speedup over FlashAttention-3; whether an adaptive method can reach $(E=9, Ex=10)$ under the current rubric remains an open question, particularly at 128K+ sequence lengths where routing overhead grows.

Applying Consistency Check 1 to the full EEI score table (Table~\ref{tab:eei-scores}) confirms the rubric's internal consistency: score or kernel approximation at $E \geq 7$ yields $Ex = 6$ (Linformer: $E=7, Ex=6$; Performer: $E=7, Ex=6$), fixed support selection reaches at most $Ex = 8$ (Longformer: $E=8, Ex=7$ via sliding-window and global tokens; BigBird: $E=7, Ex=8$ via random, window, and global tokens, whose LRA average of 55.01 is only marginally above full attention's 54.39), and Mamba ($E=10, Ex=7$) and Mamba-2 ($E=10, Ex=8$) reach their scores through the SSD duality's exact structured formulation rather than approximation. Reformer ($E=6, Ex=7$) falls below the $E \geq 7$ threshold of the bound, combining LSH support selection with $Ex = 7$. No method achieves $Ex \geq 9$ through approximation or fixed support selection: among approximation- and fixed-pattern methods, Focus and DashAttention are the only ones reaching $E \geq 7$ with $Ex \geq 9$, and they do so via learned adaptive support selection rather than score or kernel approximation. Exact-computation methods reach the same region by a different route, namely GQA ($E=7, Ex=9$), FlashAttention-2/3 ($E=8, Ex=10$), PagedAttention ($E=8, Ex=10$), and Ring Attention ($E=7, Ex=10$), which is precisely what the rubric's structural assumption predicts: approximation is capped, exact computation and learned selection are not.

\paragraph{Limitations of EEI.}
The EEI framework has several inherent limitations that readers should consider when interpreting the scores and rankings.

\textit{Scores remain partly subjective by construction.} Although the rubric (Table~\ref{tab:eei-rubric}) anchors each score band to specific criteria, the assignment of a method to a particular integer score within a band relies on the author's judgment. Two raters could disagree on whether FlashAttention's efficiency merits an 8 or a 9 depending on how they weight asymptotic complexity versus practical speedup. We mitigate this by providing detailed per-method justifications (Appendix~\ref{app:eei-justification}), but the inherent ordinal subjectivity cannot be eliminated without a formal multi-rater validation study.

\textit{Interpretability lacks standardized benchmarks.} The Interpretability (I) axis is the least grounded dimension of EEI. Unlike efficiency (measurable via throughput) and expressiveness (measurable via retrieval accuracy), interpretability has no widely accepted quantitative metric. The I scores reflect the availability of mechanistic analysis tools and theoretical frameworks for each architecture, but this assessment remains qualitative and conflates two distinct concepts: whether a mechanism is analyzable in principle (architectural interpretability) versus whether tools happen to exist for its architecture class (tooling availability). A low I score for an SSM, for example, partly reflects that SSMs are newer and fewer interpretability tools target them, not that the mechanism is fundamentally opaque. Until the community develops standardized interpretability benchmarks (which we identify as an open problem in Section~7.1), the I axis will remain the most subjective component of EEI.

\textit{Weight selection changes rankings.} The composite EEI score under different weight configurations (equal, retrieval-focused, deployment-focused) produces different rankings, as shown in Section~2. A method ranked first under deployment weights (Mamba-2, EEI $= 8.8$) drops to a four-way tie at 8.0 under retrieval weights, while Mamba falls from second under deployment weights to a three-way tie near rank 12 of 21 (EEI $= 7.2$). This sensitivity is a feature of the framework, in that it captures genuine trade-offs, but it also means that no single EEI ranking is objective. Users of EEI must specify their weight configuration alongside any reported composite scores.

\textit{Future architectures may require additional axes.} The three-axis framework was designed to cover the current landscape, but emerging research suggests potential additional dimensions. Energy efficiency (FLOPs/watt) matters more for deployment at scale. Robustness to adversarial inputs is orthogonal to expressiveness but practically important. Training stability and convergence speed are relevant for method selection but not captured by the current axes. The EEI framework is extensible by design: new axes can be added by defining a scoring rubric analogous to Table~\ref{tab:eei-rubric} and extending the composite formula, but the current three-axis formulation may not fully capture all relevant trade-offs for future architectures.

\noindent The EEI framework should be interpreted as a conceptual vocabulary and organizing heuristic that structures the design space, not as a definitive ranking of architectures. The scores and rankings are intended to guide method selection and clarify trade-offs, not to replace task-specific empirical evaluation.

\section{Foundations of Attention Mechanisms}\label{foundations-of-attention-mechanisms}

\subsection{The Bahdanau--Luong Heritage}\label{the-bahdanauluong-heritage}

Attention mechanisms predate the Transformer by several years. The original formulation was introduced by Bahdanau et al.~(2015), who proposed an additive attention mechanism for neural machine translation. In their encoder-decoder architecture, a bidirectional RNN encoded the source sentence, and at each decoding step, a learned alignment model computed a context vector as a weighted sum of encoder hidden states. The alignment score between decoder hidden state $s_{i-1}$ and encoder hidden state $h_j$ was computed through a feedforward network: $e_{ij} = v^T \tanh(W s_{i-1} + U h_j)$. The weights $\alpha_{ij} = \text{softmax}(e_{ij})$ then defined a soft alignment between source and target positions.

Luong et al.~(2015) simplified this formulation by proposing multiplicative (dot-product) attention as an alternative to the additive form. Their global attention mechanism computed scores as $e_{ij} = s_i^T h_j$, dramatically reducing the parameter count and computational cost of the alignment model. They also introduced a local attention variant that first predicted a single aligned position and then computed attention over a window centered on that position, presaging later work on sparse and windowed attention patterns.

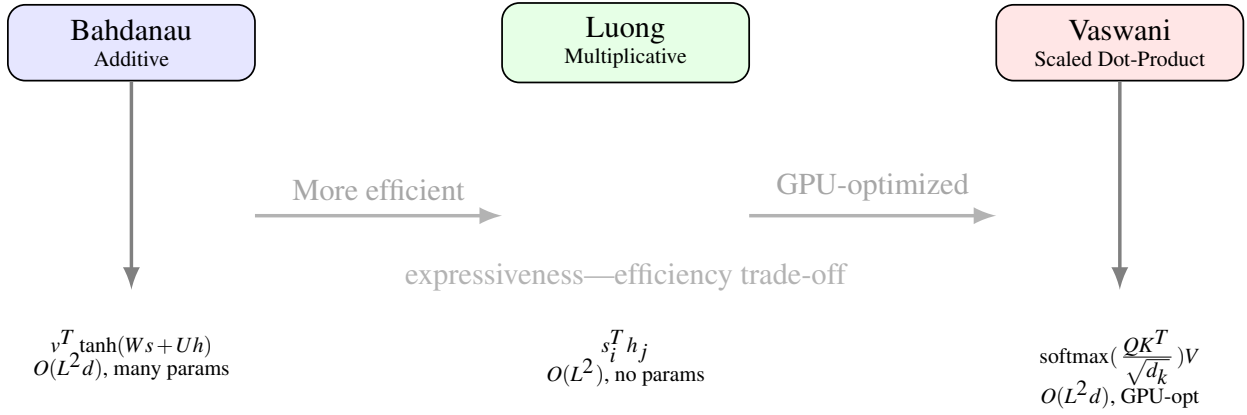
\begin{figure*}[!tbp]
\centering
\resizebox{\textwidth}{!}{\begin{tikzpicture}[
  box/.style={rectangle, draw, rounded corners, minimum width=2.0cm, minimum height=0.5cm, align=center, font=\scriptsize},
]
\node[box, fill=blue!10] (add) at (0,0) {Bahdanau\\[-2pt]\tiny Additive};
\node[box, fill=green!10] (mul) at (4,0) {Luong\\[-2pt]\tiny Multiplicative};
\node[box, fill=red!10] (sdp) at (8,0) {Vaswani\\[-2pt]\tiny Scaled Dot-Product};

\draw[->, >=latex, thick, gray] (add.south) -- (0,-2.0);
\draw[->, >=latex, thick, gray] (sdp.south) -- (8,-2.0);

\draw[->, >=latex, thick, gray!60] (1.0,-1.4) -- (3.0,-1.4);
\draw[->, >=latex, thick, gray!60] (5.0,-1.4) -- (7.0,-1.4);

\node[font=\scriptsize, text=gray!70, anchor=south] at (2.0,-1.4) {More efficient};
\node[font=\scriptsize, text=gray!70, anchor=south] at (6.0,-1.4) {GPU-optimized};

\node[font=\scriptsize, text=gray!60] at (4,-1.9) {expressiveness---efficiency trade-off};

\node[font=\tiny, align=center, anchor=north] at (0,-2.2) {$v^T\tanh(Ws+Uh)$\\[-2pt]$O(L^2 d)$, many params};
\node[font=\tiny, align=center, anchor=north] at (4,-2.2) {$s_i^T h_j$\\[-2pt]$O(L^2)$, no params};
\node[font=\tiny, align=center, anchor=north] at (8,-2.2) {softmax$(\frac{QK^T}{\sqrt{d_k}})V$\\[-2pt]$O(L^2 d)$, GPU-opt};
\end{tikzpicture}}
\caption{Three foundational attention variants: additive (Bahdanau), multiplicative (Luong), and scaled dot-product (Vaswani).}\label{fig:variants}
\end{figure*}

These two formulations defined the key trade-off that persists in attention research: additive attention uses a learned nonlinear scoring function but is computationally heavier, while multiplicative attention is simpler, faster, and amenable to hardware-optimized matrix multiply units. The Transformer chose the latter path, but the additive formulation persists in graph neural networks and multimodal architectures where the alignment function requires more expressiveness. Figure~\ref{fig:variants} shows the three foundational variants side by side, and Table~\ref{tab:variants} summarizes their complexity and interpretability.

\begin{table*}[!tbp]
\centering
\footnotesize
\caption{Comparison of foundational attention variants.}
\label{tab:variants}
\begin{tabular}{p{2.2cm}p{2.4cm}p{1.8cm}p{1.3cm}p{2.8cm}p{1.8cm}}
\toprule
Variant & Score Function & Complexity & Parameters & Expressiveness & Hardware \\
\midrule
Bahdanau (Additive) & $v^T \tanh(W s + U h)$ & $O(L_{\mathrm{dec}}L_{\mathrm{enc}}d)$ & High (W,U,v) & High: learned nonlinear & Low: no single batched $QK^\top$ \\
Luong (Multiplicative) & $s^T h$ & $O(L_{\mathrm{dec}}L_{\mathrm{enc}}d)$ & None & Moderate: linear & High: dot product \\
Vaswani (Scaled Dot-Product) & $QK^T / \sqrt{d_k}$ & $O(L^2 d)$ & None & Moderate + scaling & Very High: tensor cores \\
\bottomrule
\end{tabular}
\end{table*}

\subsection{Scaled Dot-Product and Multi-Head Attention}\label{scaled-dot-product-and-multi-head-attention}

Vaswani et al.~(2017) codified attention into the canonical form used by most modern architectures. Given queries Q, keys K, and values V, attention is computed as:

$\text{Attention}(Q,K,V) = \text{softmax}\left(\frac{QK^T}{\sqrt{d_k}}\right)V$

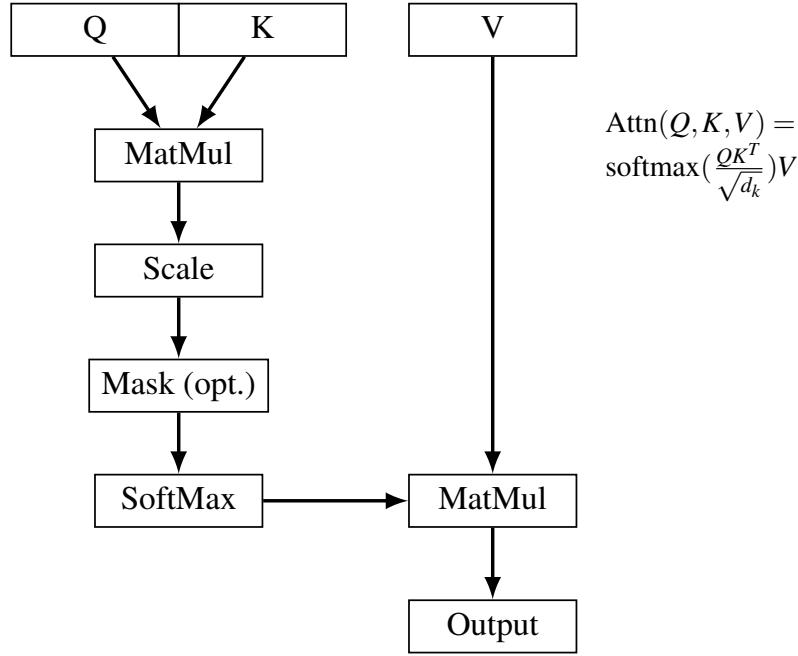
\begin{figure*}[!tbp]
\centering
\resizebox{0.65\textwidth}{!}{\begin{tikzpicture}[scale=0.85, transform shape,
  box/.style={rectangle, draw, minimum width=1.6cm, minimum height=0.5cm, align=center, font=\small, text height=0.6em, text depth=0.15em},
  arrow/.style={->, >=latex, thick}
]
\node[box] (Q) at (0,0) {Q};
\node[box] (K) at (1.6,0) {K};
\node[box] (V) at (3.8,0) {V};
\node[box] (matmul1) at (0.8,-1.2) {MatMul};
\node[box] (scale) at (0.8,-2.3) {Scale};
\node[box] (mask) at (0.8,-3.4) {Mask (opt.)};
\node[box] (softmax) at (0.8,-4.5) {SoftMax};
\node[box] (matmul2) at (3.8,-4.5) {MatMul};
\node[box] (output) at (3.8,-5.7) {Output};
\draw[arrow] (Q) -- (matmul1);
\draw[arrow] (K) -- (matmul1);
\draw[arrow] (matmul1) -- (scale);
\draw[arrow] (scale) -- (mask);
\draw[arrow] (mask) -- (softmax);
\draw[arrow] (softmax.east) -- (matmul2.west);
\draw[arrow] (V) -- (matmul2);
\draw[arrow] (matmul2) -- (output);
\node[align=left, font=\footnotesize] at (5.8,-1.2) {$\text{Attn}(Q,K,V)=$\\$\text{softmax}(\frac{QK^T}{\sqrt{d_k}})V$};
\end{tikzpicture}}
\caption{Scaled dot-product attention computation flow.}\label{fig:attention}
\end{figure*}

Figure~\ref{fig:attention} illustrates this computation.

Two design choices were critical. First, the scaling factor $1/\sqrt{d_k}$ prevents the dot products from growing large in magnitude as the dimension $d_k$ increases, which would push the softmax into regions of extremely small gradients. Second, the use of multiple attention heads allows the model to attend to different representation subspaces simultaneously. Each head computes independent attention over $d_k$-dimensional projections of Q, K, V, and the outputs are concatenated and projected back to the model dimension.

Multi-head attention was essential for expressiveness. A single softmax attention head distributes probability mass across a weighted combination of tokens, but a single distribution may be insufficient when a token needs to attend to multiple distinct positions for different reasons. With h heads, the model can allocate different heads to different functions: one head might attend to syntactic dependencies while another tracks positional information.

The keys and values of previous tokens are cached during incremental decoding, forming the key-value (KV) cache, a device now central to inference efficiency in autoregressive models: only the current query's attention computation needs to be performed, at the cost of $O(h \cdot L \cdot d_k)$ memory for $h$ heads that becomes significant at long contexts.

The KV cache memory footprint has driven two critical architectural innovations. Multi-Query Attention (MQA) (Shazeer, 2019) shares a single key and value head across all query heads, reducing the KV cache from $O(h \cdot L \cdot d_k)$ to $O(L \cdot d_k)$ at the cost of expressiveness. Grouped-Query Attention (GQA) (Ainslie et al., 2023) interpolates between MHA and MQA by partitioning query heads into $g$ groups, each with a dedicated KV head, yielding a KV cache of $O(g \cdot L \cdot d_k)$ where $g < h$. GQA has become a widely used attention variant in production-scale LLMs (Llama 2 (70B; Touvron et al., 2023), Llama 3 (Dubey et al., 2024), Mistral (Jiang et al., 2023)) because it preserves most of MHA's quality while reducing KV cache memory by a factor of h/g. The choice of g defines a quality-efficiency trade-off that interacts directly with the sparse and IO-aware methods discussed in Section 5.

\subsection{Positional Encoding}\label{positional-encoding}

Because attention is permutation-equivariant (shuffling the input shuffles the output in the same way), the Transformer must inject positional information through external encodings. Vaswani et al.~(2017) proposed sinusoidal positional encodings using fixed frequencies:

\begin{equation*}\text{PE}(pos, 2i) = \sin\left(pos / 10000^{2i/d_{\text{model}}}\right)\end{equation*}
\begin{equation*}\text{PE}(pos, 2i+1) = \cos\left(pos / 10000^{2i/d_{\text{model}}}\right)\end{equation*}

These encodings have the useful property that any offset can be represented as a linear function of the original position encoding, potentially allowing the model to generalize to unseen sequence lengths. In practice, learned positional embeddings were adopted by BERT (Devlin et al., 2019) and GPT models (Brown et al., 2020), trading the theoretical generalization of sinusoids for the flexibility of data-driven position representations.

A more recent innovation is Rotary Position Encoding (RoPE) (Su et al., 2021), which encodes position by rotating queries and keys in complex space. RoPE has become the dominant positional encoding in modern large language models because it makes attention scores explicitly dependent on relative positional offsets through phase rotations, integrates seamlessly with linear attention mechanisms, and exhibits better length generalization than learned absolute embeddings. RoPE's relative position bias emerges naturally from the rotation operation without requiring additional learnable parameters per position.

An alternative approach, Attention with Linear Biases (ALiBi) (Press et al., 2022), adds a static, distance-proportional bias to the attention scores before softmax: $score(i, j) = q_i \cdot k_j - m \cdot |i - j|$, where $m$ is a head-specific slope. In the original causal formulation this bias applies over the allowed past positions $j \leq i$; the symmetric absolute-distance form extends the penalty convention to bidirectional attention. ALiBi requires no learned positional embeddings and no modification to the attention computation beyond the bias term. Its key advantage is length generalization: models trained on short sequences (e.g., 1K tokens) can extrapolate to much longer sequences (e.g., 8K+) at inference time without additional training. This property made ALiBi the positional encoding of choice for several open-source model families, including MPT (MosaicML, 2023) and BLOOM (Le Scao et al., 2022). The trade-off is that ALiBi's fixed bias cannot adapt to task-specific positional relationships, and later work (YaRN, PI) showed that RoPE with interpolation techniques can achieve better length generalization than ALiBi when additional training or fine-tuning is available.

Building on RoPE's relative position bias, three interpolation techniques have extended its length generalization frontier. Position Interpolation (PI) (Chen et al., 2023) linearly scales position indices to fit within the pretrained context window during fine-tuning, enabling Llama-style models to extend from 2K to 32K tokens with continued fine-tuning and minimal perplexity degradation. YaRN (Yet another RoPE extensioN) (Peng et al., 2023b) combines two distinct operations: NTK-by-parts interpolation, which leaves short-wavelength dimensions unchanged, fully interpolates long-wavelength dimensions, and blends the intermediate range; and a separate attention-scaling term that adjusts the softmax temperature. The source reports faster convergence and longer-context performance than the compared interpolation baselines under its fine-tuning setups. NTK-aware scaling, a community-originated technique first publicly described in a community technical post (bloc97, 2023), draws on the Neural Tangent Kernel literature to scale RoPE frequencies non-uniformly, prioritizing high-frequency dimensions that encode fine-grained local position information while compressing low-frequency dimensions that encode global position. NTK-aware scaling influenced subsequent RoPE extension methods, including NTK-by-parts variants and YaRN (Peng et al., 2023b). These methods have been widely explored for extending RoPE-based context windows, but their adoption and effectiveness depend on the model family and training procedure; they do not themselves reduce the quadratic attention cost. For example, DeepSeek-V2 (DeepSeek-AI, 2024) uses YaRN-style scaling to support a 128K-token context window, whereas Llama-3-based models (Dubey et al., 2024) extend their contexts primarily through long-context fine-tuning rather than interpolation. These interpolation methods are largely complementary to the efficiency innovations in Section 5, making them especially impactful when combined with IO-aware attention or adaptive sparsity.

\subsection{Causal vs.~Bidirectional Architectures}\label{causal-vs.-bidirectional-architectures}

\begin{figure*}[!tbp]
\centering
\resizebox{0.65\textwidth}{!}{\begin{tikzpicture}[font=\footnotesize]
\foreach \x in {0,1,2} {
  \foreach \y in {0,1,2} {
    \fill[blue!40!white] (\x*0.55+0.15,-\y*0.55-0.15) rectangle ++(0.45,0.45);
  }
}
\draw (0.05,0.35) rectangle (1.80,-1.35);
\node[align=center, font=\fontsize{7}{7}\selectfont] at (0.95,0.75) {Bidirectional\\(Full)};
\node[below, font=\fontsize{7}{7}\selectfont] at (0.95,-1.45) {Token $N$};
\node[right, font=\fontsize{7}{7}\selectfont] at (-1.0,-0.50) {Token $1$};

\begin{scope}[xshift=3.5cm]
\foreach \x in {0,1,2} {
  \foreach \y in {0,1,2} {
    \ifnum\y<\x\relax
      \fill[gray!20!white] (\x*0.55+0.15,-\y*0.55-0.15) rectangle ++(0.45,0.45);
    \else
      \fill[blue!40!white] (\x*0.55+0.15,-\y*0.55-0.15) rectangle ++(0.45,0.45);
    \fi
  }
}
\draw (0.05,0.35) rectangle (1.80,-1.35);
\node[align=center, font=\fontsize{7}{7}\selectfont] at (0.95,0.75) {Causal\\(Masked)};
\node[below, align=center, font=\fontsize{7}{7}\selectfont] at (0.95,-1.45) {Token $N$\\Keys (columns)};
\node[right, align=center, font=\fontsize{7}{7}\selectfont] at (1.90,-0.50) {Token $1$\\Queries (rows)};
\end{scope}
\end{tikzpicture}}
\caption{Bidirectional (full) vs.\ causal (masked) attention patterns.}\label{fig:masks}
\end{figure*}
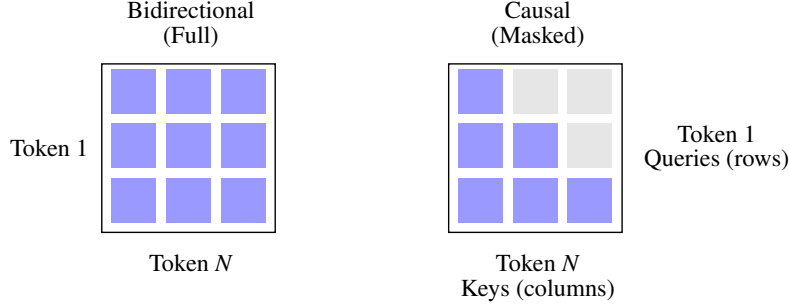

The attention pattern defines the architectural family of a Transformer model. Three canonical patterns have emerged. Encoder-only models such as BERT (Devlin et al., 2019) use bidirectional (full) attention, where every token can attend to every other token. This produces rich contextual representations ideal for classification and span-labeling tasks, but it is not suitable for language generation because future tokens are visible to the model during training. Figure~\ref{fig:masks} shows the bidirectional and causal masking patterns.

Decoder-only models employ causal (masked) attention, where each token can only attend to itself and preceding tokens. This autoregressive constraint enables left-to-right generation and has become the dominant paradigm for large language models. The GPT family (Brown et al., 2020) popularized this approach, and nearly all contemporary frontier models (GPT-4 (OpenAI, 2023), PaLM (Chowdhery et al., 2023), Llama (Touvron et al., 2023), Claude (Anthropic, 2024), Gemini (Gemini Team, 2024), DeepSeek (DeepSeek-AI, 2024)) are autoregressive next-token predictors. The causal mask can be implemented efficiently as a triangular matrix that sets future-position entries to $-\infty$ (or a sufficiently negative finite sentinel in numerical code) in the attention logits before the softmax operation. Setting forbidden logits to zero would generally assign them positive probability and is therefore incorrect unless followed by explicit probability masking and renormalization.

Encoder-decoder models, following the original Transformer formulation (Vaswani et al., 2017), combine a bidirectional encoder with a causal decoder connected through cross-attention. The encoder processes the input sequence bidirectionally, while cross-attention layers in the decoder attend to the encoder's output. This architecture remains standard in machine translation, text summarization, and speech recognition, though decoder-only models have increasingly absorbed these use cases through in-context learning.

The choice between these architectures carries implications for efficiency and interpretability. Bidirectional attention computes a full $O(L^2)$ matrix but can parallelize over all positions. Causal attention requires the same quadratic computation but can leverage the KV cache during decoding, shifting the bottleneck from compute to memory. Cross-attention introduces an additional asymmetry (the decoder queries attend to encoder keys) that enables distinct roles for the encoder and decoder representations. It adds an additional attention operation in the decoder with cost proportional to the product of encoder and decoder sequence lengths, $O(L_{\mathrm{dec}}L_{\mathrm{enc}}d)$, rather than a fixed multiple of a single self-attention computation.

\begin{center}\rule{0.5\linewidth}{0.5pt}\end{center}
\paragraph{Key Takeaways.} (i) Scaled dot-product attention with multi-head expansions remains the canonical formulation after a decade of alternatives. (ii) RoPE has emerged as the dominant positional encoding for decoder-only LLMs, while ALiBi offers length-generalization advantages for fixed-budget training. (iii) Causal masking and KV caching define the inference-time efficiency profile of autoregressive models, making them structurally different from bidirectional encoders. (iv) The choice of architecture (encoder-only, decoder-only, encoder-decoder) determines which attention variants are applicable and which efficiency optimizations are feasible.

\section{Attention in Computer Vision}\label{attention-in-computer-vision}

This section provides an illustrative rather than exhaustive treatment of attention in vision, covering four representative architectural patterns. Methods that fall outside this scope (including SAM (Kirillov et al., 2023), Perceiver IO (Jaegle et al., 2022), DeiT (Touvron et al., 2021), DINOv2 (Oquab et al., 2024), VideoMAE (Tong et al., 2022), and TimeSformer (Bertasius et al., 2021)) are discussed contextually in the Key Takeaways and referenced in the bibliography, but do not receive dedicated subsections. The EEI framework can be applied to these methods in future work.

\subsection{Vision Transformer}\label{vision-transformer}

The Vision Transformer (ViT) (Dosovitskiy et al., 2021) adapted the Transformer architecture to image classification by treating image patches as tokens. The key insight of ViT was that pure attention, without convolutional inductive biases, could match CNN performance when trained on sufficient data. An image of size $H \times W \times C$ is divided into $P \times P$ patches, each flattened and linearly projected into a patch embedding. A special \texttt{[CLS]} token is prepended to the sequence, and standard positional embeddings are added. The resulting sequence of $L = HW/P^2 + 1$ tokens is processed by a standard Transformer encoder with bidirectional self-attention.

On ImageNet-1k with standard 1.3M images, ViT underperformed ResNet-based models. But when pretrained on ImageNet-21k or JFT-300M (a proprietary Google dataset with 300M images), ViT matched or exceeded state-of-the-art CNNs. This data scaling requirement drove significant subsequent research into data-efficient vision transformers. DeiT (Touvron et al., 2021) introduced a teacher-student distillation strategy that enabled ViT to compete with CNNs using only ImageNet-1k, while MAE (He et al., 2022) showed that masked autoencoding (randomly masking 75\% of patches and reconstructing the missing pixels) works as a self-supervised pretraining objective for ViT. DINO (Caron et al., 2021) and DINOv2 (Oquab et al., 2024) extended self-supervised learning to vision transformers: DINO discovered that the \texttt{[CLS]} token's attention maps naturally segment objects, and DINOv2 showed that PCA of patch features separates objects and their parts, an emergent property that connects to the attention-interpretability themes explored in Section 6.

ViT's quadratic complexity in the number of patches limits its application to high-resolution images. A $224 \times 224$ image with $16 \times 16$ patches yields 196 tokens, which is manageable, but a $1024 \times 1024$ image produces 4096 tokens, requiring $\sim$440$\times$ more attention computation ($\sim$21$\times$ more tokens, quadratically compounded). This scaling challenge motivated hierarchical approaches that reduce token count through spatial merging.

\subsection{DETR: Object Queries as Learned Attention}\label{detr-object-queries-as-learned-attention}

The Detection Transformer (DETR) (Carion et al., 2020) reformulated object detection as a direct set prediction problem, eliminating the need for hand-crafted components like anchor boxes and non-maximum suppression. DETR uses a standard Transformer encoder-decoder architecture where the decoder consumes a fixed set of N learned object-query embeddings; these are learned queries, not positional embeddings, and each can specialize in predicting a specific object in the image.

The object queries function as learned slot attention. Each query competes with others through the decoder's self-attention to specialize on different objects, and the cross-attention layers between decoder queries and encoder image features determine where each query looks in the image. The bipartite matching loss (Hungarian algorithm) assigns each ground-truth object to a unique query, enabling end-to-end training without hand-coded post-processing.

DETR introduced a new paradigm of treating attention outputs as learned slots or object representations. This idea has been extended to video object detection (He et al., 2021), panoptic segmentation, and multimodal understanding (Kamath et al., 2021), where queries can represent not just spatial locations but also semantic concepts or temporal segments.

\subsection{Swin Transformer and Hierarchical Attention for Vision}\label{swin-transformer-and-hierarchical-attention-for-vision}

The Swin Transformer (Liu et al., 2021) addressed ViT's quadratic scaling and lack of hierarchical features by introducing windowed self-attention. The image is partitioned into non-overlapping local windows of M $\times$ M patches. Self-attention is computed only within each window, reducing complexity from $O(H_p^2W_p^2d)$ to $O(H_pW_pM^2d)$, where $H_p=H/P$ and $W_p=W/P$ are patch-grid dimensions. A shifted windowing mechanism alternates between regular and shifted partitions across consecutive layers, enabling cross-window connections while maintaining the efficiency of local attention.

The hierarchical windowing strategy in Swin bears a direct conceptual connection to the windowed attention used in Longformer (Section 5.1) and the sliding window attention deployed in Mistral and other decoder-only LLMs. In all three cases, local windows provide efficient fine-grained processing while alternative mechanisms (shifted windows in Swin, global tokens in Longformer) enable cross-window information flow. This recurring design pattern, local computation with sparse global mixing, appears across modalities and architectural families.

Video attention extends the windowing principle to the temporal dimension. TimeSformer (Bertasius et al., 2021) factorizes video attention into separate spatial and temporal components: spatial attention within each frame and temporal attention at each spatial location across frames. With $N$ spatial tokens per frame (the patch-grid count $(H/P)(W/P)$ in the notation of Section~4), this factorization reduces complexity from $O((T \cdot N)^2)$ to $O(T \cdot N^2 + N \cdot T^2)$, making video attention tractable while producing strong results on video classification benchmarks. VideoMAE (Tong et al., 2022) extended the masked autoencoding paradigm of MAE to video, using extremely high masking ratios (90--95\%) to learn spatiotemporal representations efficiently.

This design produces a hierarchical representation analogous to CNNs. Early layers have high-resolution features with small receptive fields (within windows), while deeper layers merge windows through spatial downsampling to produce lower-resolution features with larger effective receptive fields. The resulting feature pyramid is compatible with dense prediction tasks like semantic segmentation and object detection, where multi-scale representations are essential.

Swin Transformer achieved top results on ImageNet classification and established hierarchical windowed attention as a widely adopted and highly efficient vision Transformer design in 2021--2023. Its success demonstrated that carefully designed local attention patterns could match or exceed both pure Transformers and CNNs on vision tasks, without requiring the massive pretraining data that ViT needed.

A key vision-specific efficiency consideration is how the token count scales with input resolution. For a 4K$\times$4K satellite image at 16$\times$16 patch size, the token sequence is 65\,536 tokens (64K), well beyond the 32K range where IO-aware attention remains efficient. In such regimes, Swin's windowed attention ($O(H_pW_pM^2d)$ with $M=7$) maintains 49 tokens per window regardless of input size, making its complexity genuinely linear in pixel count. This scaling property, combined with the hierarchical feature pyramid that supports dense prediction tasks directly, is why Swin-based backbones remain competitive in high-resolution vision applications (medical imaging, remote sensing, video analysis) where patch-only ViT would be computationally prohibitive.

\subsection{Vision-Language Cross-Attention}\label{vision-language-cross-attention}

Multimodal models that align vision and language representations rely on cross-attention mechanisms to bridge modalities. CLIP (Contrastive Language-Image Pre-training; Radford et al., 2021) uses a dual-encoder architecture where an image encoder (typically ViT) and a text encoder (typically a Transformer) produce aligned embeddings through a contrastive loss. The alignment is learned at the representation level: the pooled representation of the image encoder and the final hidden state of the text encoder are projected into a shared embedding space where cosine similarity measures semantic alignment.

A distinct approach, exemplified by Flamingo (Alayrac et al., 2022), uses cross-attention to fuse visual features into language model decoders: visual features from a frozen vision encoder are treated as keys and values in cross-attention layers inserted between the self-attention and feedforward layers of an autoregressive language model, which generates text conditioned on visual information by attending to the visual token sequence. FLORENCE (Yuan et al., 2021) instead follows a CLIP-style two-tower architecture, training a hierarchical vision Transformer and a text Transformer with a contrastive (UniCL) objective, with a lightweight adapter for vision-and-language tasks.

The efficiency challenge is acute in vision-language models because visual token sequences are typically much longer than text sequences. A single high-resolution image can produce thousands of visual tokens. Perceiver-based architectures (Jaegle et al., 2021) address this by learning a smaller set of latent queries that cross-attend to the full visual feature map, effectively compressing the visual information before it enters the cross-attention layers. This query-based compression anticipates the hierarchical and learned sparsity approaches we examine in Section 5.

\begin{center}\rule{0.5\linewidth}{0.5pt}\end{center}
\paragraph{Key Takeaways.} (i) ViT demonstrated that pure attention without convolutional biases can match CNNs at scale, establishing patch-based tokenization as the standard for vision. (ii) DETR's object queries introduced a learned attention mechanism that replaces hand-crafted post-processing with end-to-end differentiable matching. (iii) Swin's hierarchical shifted windows remain a highly efficient and widely adopted vision-attention design, achieving near-CNN efficiency with Transformer-level expressiveness. (iv) Vision-language models incur a token-count asymmetry that makes cross-attention compression (Perceiver-style) an important practical strategy for high-resolution inputs. (v) This section covers a selective subset of vision attention architectures (ViT, DETR, Swin, CLIP-style dual-encoder contrastive alignment) as illustrative examples; notable methods not scored in our EEI framework include SAM (Kirillov et al., 2023), Perceiver IO (Jaegle et al., 2022), and DeiT (Touvron et al., 2021), which may exhibit different EEI profiles under our rubric.

\section{Efficiency Innovations}\label{efficiency-innovations}

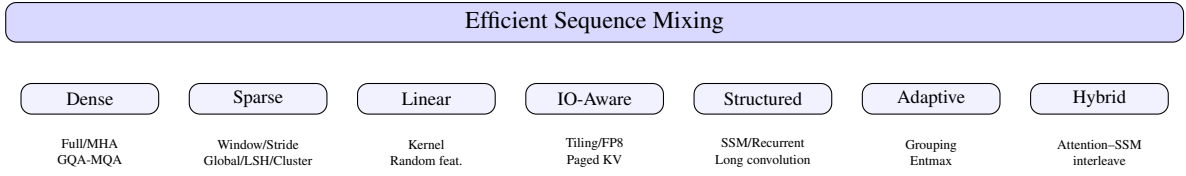
\begin{figure*}[!tbp]
\centering
\resizebox{0.95\textwidth}{!}{\begin{tikzpicture}[
  box/.style={rectangle, draw, rounded corners, minimum width=1.8cm, minimum height=0.4cm, align=center, font=\scriptsize, inner sep=2pt},
  leaf/.style={font=\tiny, align=center}
]
\node[rectangle, draw, rounded corners, minimum width=15.4cm, minimum height=0.5cm, align=center, font=\small, fill=blue!15] at (0,0) {Efficient Sequence Mixing};

\node[box, fill=blue!5] at (-6.6,-1.0) {Dense};
\node[box, fill=blue!5] at (-4.4,-1.0) {Sparse};
\node[box, fill=blue!5] at (-2.2,-1.0) {Linear};
\node[box, fill=blue!5] at (0,-1.0) {IO-Aware};
\node[box, fill=blue!5] at (2.2,-1.0) {Structured};
\node[box, fill=blue!5] at (4.4,-1.0) {Adaptive};
\node[box, fill=blue!5] at (6.6,-1.0) {Hybrid};

\node[leaf] at (-6.6,-1.7) {Full/MHA\\GQA-MQA};
\node[leaf] at (-4.4,-1.7) {Window/Stride\\Global/LSH/Cluster};
\node[leaf] at (-2.2,-1.7) {Kernel\\Random feat.};
\node[leaf] at (0,-1.7) {Tiling/FP8\\Paged KV};
\node[leaf] at (2.2,-1.7) {SSM/Recurrent\\Long convolution};
\node[leaf] at (4.4,-1.7) {Grouping\\Entmax};
\node[leaf] at (6.6,-1.7) {Attention--SSM\\interleave};
\end{tikzpicture}}
\caption{Taxonomy of efficient sequence-mixing methods (abridged; the seven families and leaf groupings follow Table~\ref{tab:deep-taxonomy}).}\label{fig:taxonomy}
\end{figure*}

The $O(L^2)$ complexity of standard attention is the central bottleneck in scaling Transformers to long sequences. The research community has pursued seven efficiency/architecture families to address this limitation, each with distinct trade-offs between efficiency, quality, and hardware compatibility. Table~\ref{tab:deep-taxonomy} details the seven-family taxonomy, and Figure~\ref{fig:taxonomy} summarizes it visually.

\subsection{Fixed Sparse Patterns}\label{fixed-sparse-patterns}

The earliest approaches to efficient attention replaced the dense attention matrix with predetermined sparse patterns. The Sparse Transformer (Child et al., 2019) introduced strided and fixed attention patterns that reduced complexity to $O(L\sqrt{L})$. The strided pattern allows each token to attend to nearby tokens and to every k-th token in the sequence, enabling information propagation across long distances at reduced cost. Fixed patterns require no learning, introduce no overhead beyond custom kernel implementations, and guarantee predictable memory usage. Their limitation is inflexibility: a fixed pattern cannot adapt to the content-dependent nature of attention, potentially missing important interactions that fall outside the predetermined connections.

\begin{table*}[!tbp]
\centering
\sflt{\fontsize{7}{8.4}\selectfont}{\scriptsize}
\renewcommand{\arraystretch}{\sflt{1.0}{1.05}}
\caption{Comprehensive taxonomy of sequence-efficiency methods. The seven top-level families are intentionally heterogeneous: they classify the primary source of efficiency or architectural change rather than enforcing a single ontological level (e.g., IO-aware is an implementation strategy, sparse is an attention pattern, structured sequence is an alternative operator family, and hybrid is an architectural composition). Here $d_{\mathrm{model}}=h d_k$, $g_{\mathrm{KV}}$ is the number of KV groups, $m$ is the kernel-feature dimension, $d_v$ is the value dimension, $K$ is the learned-group count, $B$ is block size, $s$ is the selected-block count, $r_{\mathrm{random}}$ and $n_{\mathrm{global}}$ are the random- and global-token counts, and $n_A,n_S,n_R$ count attention, SSM, and recurrent layers. Constant factors and linear-projection costs are suppressed.}
\label{tab:deep-taxonomy}
\begin{tabular}{p{2.3cm}p{2.2cm}p{2.2cm}p{2.8cm}p{2.0cm}}
\toprule
Category & Subcategory & Representative Works & Mechanism & Complexity \\
\midrule
\multirow{3}{2.3cm}{Dense} & Full Softmax & Vaswani et al., 2017 & $QK^{\top}$ pairwise & $O(L^2d_{\mathrm{model}})$ \\
 & Multi-Head & Vaswani et al., 2017 & $h$ projected heads & $O(L^2d_{\mathrm{model}})$ \\
 & GQA/MQA & Ainslie et al., 2023; Shazeer, 2019 & Shared KV heads & $O(L^2d_{\mathrm{model}})$ comp.; $O(g_{\mathrm{KV}}Ld_k)$ cache \\
\midrule
\multirow{5}{2.3cm}{Sparse} & Fixed Window & Beltagy et al., 2020; Liu et al., 2021 & $w$-token local window & $O(Lwd_{\mathrm{model}})$ \\
 & Fixed Strided & Child et al., 2019 & Every-$k$ pattern & $O(L\sqrt{L}\,d_{\mathrm{model}})$ \\
 & Hybrid Global & Beltagy et al., 2020; Zaheer et al., 2020 & Window + random + global & $O(L(w+r_{\mathrm{random}}+n_{\mathrm{global}})d_{\mathrm{model}})$ \\
 & LSH Hashing & Kitaev et al., 2020 & Hash, sort, bucket attention & $O(L\log L\,d_{\mathrm{model}})$ \\
 & Clustering & Roy et al., 2021 & $k$-means routing & $O(L^{1.5}d_{\mathrm{model}})$ (balanced clusters $k=\Theta(\sqrt{L})$) \\
\midrule
\multirow{3}{2.3cm}{Linear} & Kernel Trick & Katharopoulos et al., 2020 & Normalized feature attention & $O(Lmd_v)$ (+$O(Ldm)$ feature-map construction) \\
 & Random Features & Choromanski et al., 2021 & FAVOR+ approximation & $O(Lmd_v)$ (+$O(Ldm)$ feature-map construction) \\
 & BASED Hybrid & Arora et al., 2024 & Taylor linear attention + local softmax window & $O(Lmd_v+Lwd_v)$ for fixed $m,w$ \\
\midrule
\multirow{3}{2.3cm}{IO-Aware} & Tiled Exact & Dao et al., 2022; Dao, 2024 & SRAM tiling & $O(L^2d_k)$ comp.; $O(Ld_k)$ auxiliary memory \\
 & Async + FP8 & Shah et al., 2024 & Overlap + low precision & $O(L^2d_k)$ comp. \\
 & Paged KV & Kwon et al., 2023 & Paged KV-cache management & Exact-attention cost; reduced fragmentation \\
\midrule
\multirow{3}{2.3cm}{Structured Sequence} & State Space & Gu et al., 2022; Gu \& Dao, 2024 & Structured or selective recurrence & $O(L)$ scan; training form varies \\
 & Recurrent & Peng et al., 2023a; Beck et al., 2024 & Time mixing or gated matrix memory & $O(L)$ recurrence \\
 & Long Convolution & Poli et al., 2023 & Implicit long convolution & $O(L\log L)$ via FFT \\
\midrule
\multirow{2}{2.3cm}{Adaptive Sparse} & Learned Group & Yao et al., 2026 & Grouped assignment (top-$k$) + sort + local attention; inference & $O(Ld_{\mathrm{model}}K+L\log L+(\alpha L^2+Lw)d_v)$, $\alpha\in[1/K,1]$ the retained-pair fraction ($\alpha{=}1/K$ at balanced $k{=}1$; $\rightarrow 1$ as $k\rightarrow K$); factorized routing $O(Ld_{\mathrm{model}}d_g+LKd_g)$ \\
 & $\alpha$-entmax & Huang et al., 2026 & Block summaries + sparse block attention & $O(L^2/B+LsB)$, plus $O(LB)$ Stage-0 summarization; dimensions suppressed \\
\midrule
\multirow{2}{2.3cm}{Hybrid Architecture} & Attention--SSM & Lieber et al., 2024; Ren et al., 2024 & Interleaved attention and SSM layers & $O(n_A L^2d_{\mathrm{model}}+n_S Ld_{\mathrm{model}})$ \\
 & Attention--Recurrent & De et al., 2024 & Local attention + gated recurrence & $O(n_A Lwd_{\mathrm{model}}+n_R Ld_{\mathrm{model}})$ \\
\bottomrule
\end{tabular}
\end{table*}

Longformer (Beltagy et al., 2020) combined local windowed attention with task-specific global attention. Most tokens attend within a fixed window of size $w$, reducing complexity to $O(Lw)$. A small set of global tokens (e.g., \texttt{[CLS]} tokens or special task tokens) attend to the full sequence. This hybrid design preserves the efficiency of local attention while allowing critical tokens to maintain a global view. Longformer demonstrated that selective global tokens could recover most of the quality loss from pure local attention on document-level NLP tasks.

BigBird (Zaheer et al., 2020) formalized a theoretical framework for sparse attention by proving that a combination of three components, random attention, window attention, and global attention, could approximate a full Turing machine while maintaining $O(L)$ complexity. Each token attends to $r_{\mathrm{random}}$ random tokens, $w$ window tokens, and $n_{\mathrm{global}}$ global tokens. BigBird established theoretical expressiveness results showing that its sparse construction (random, window, and global attention combined) can simulate broad classes of sequence computations under the stated assumptions (approximating a full Turing machine with $O(L)$ complexity), establishing a foundation for later theoretical work on sparse attention expressiveness.

These fixed-pattern methods share a common limitation: the sparsity pattern is static and content-independent. A token attending to a random subset of tokens will inevitably miss important content-dependent interactions that fall outside the predetermined pattern.

In practice, fixed sparse patterns struggle on tasks requiring heterogeneous attention distributions, such as question answering over long documents where the query position varies unpredictably. The theoretical guarantees of BigBird's Turing-completeness provide limited guidance for practical pattern selection, and most deployments default to windowed attention with learned global tokens (Longformer's approach) due to its empirical robustness across diverse task types (Beltagy et al., 2020). The gap between theoretical expressiveness guarantees and practical performance shows that the choice of sparsity pattern matters more than asymptotic complexity alone.

\subsection{Content-Dependent Sparsity}\label{learnable-sparsity}

The next generation of methods made sparsity patterns content-dependent, through mechanisms that include hashing, clustering, and learned routing. Reformer (Kitaev et al., 2020) introduced locality-sensitive hashing (LSH) attention, which hashes queries and keys such that similar items fall into the same hash bucket with high probability. Attention is then computed only within each hash bucket, reducing complexity from $O(L^2)$ to $O(L \log L)$ under the assumption of uniform bucket distribution. The hash function is applied in a shared form across queries and keys, ensuring that the comparison is well-defined without additional parameters.

Reformer's LSH approach was an early demonstration that content-dependent sparsity could substantially reduce computation while retaining competitive quality. However, the hashing mechanism introduces a non-differentiable selection step, preventing direct gradient propagation through the routing operation and creating a mismatch between the routing objective and the task objective. Training must rely on multiple hashing rounds and averaging, increasing the constant factor of the computation.

The Routing Transformer (Roy et al., 2021) addressed this by using k-means clustering to group queries and keys into clusters, with attention computed only within clusters. Unlike Reformer's hashing, the routing decision in Routing Transformer is learned through online k-means with momentum updates, providing more stable clustering. The resulting attention is highly non-local, and different heads specialize in attending to very different parts of the input (Roy et al., 2021).

Both methods demonstrate that content-dependent sparsity can preserve model quality at significantly reduced complexity. However, the hard routing decisions (hash buckets or cluster assignments) create optimization challenges, and the overhead of the routing mechanism itself can diminish wall-clock speedups.

A practical limitation is that the computational overhead of hashing (Reformer) or clustering (Routing Transformer) can consume a substantial fraction of the theoretical savings, making these methods competitive with full attention only at longer sequence lengths. Furthermore, the non-differentiable routing decisions prevent direct gradient propagation through the routing operation, creating a mismatch between the routing objective and the task objective. This limitation motivated the differentiable sparsity approaches discussed in Section~5.6.

\subsection{Linear Attention}\label{linear-attention}

A radically different approach replaces the softmax normalization with a kernel trick that linearizes the attention computation. Linear Transformers (Katharopoulos et al., 2020) reformulate attention by replacing softmax with a general kernel function $\phi$ such that
\begin{equation*}
\text{Attention}(Q, K, V) = \frac{\phi(Q)\,(\phi(K)^\top V)}{\phi(Q)\,(\phi(K)^\top \mathbf{1})},
\end{equation*}
where $\mathbf{1}$ is a vector of ones. This noncausal form computes over the full key sequence and is suitable for bidirectional (encoder) attention. By associativity, the products $\phi(K)^\top V$ and $\phi(K)^\top \mathbf{1}$ can each be computed once and reused for all queries, reducing complexity to $O(L d^2)$ (linear in $L$ for fixed feature dimension $d$). For causal (autoregressive) attention, the sums must be computed incrementally:
\begin{equation*}
\begin{aligned}
S_i &= \sum_{j\le i}\phi(k_j)v_j^\top,\\
z_i &= \sum_{j\le i}\phi(k_j),\\
y_i &= \frac{\phi(q_i)^\top S_i}{\phi(q_i)^\top z_i+\varepsilon}.
\end{aligned}
\end{equation*}
Where $\varepsilon>0$ ensures denominator stability. The kernel feature map $\phi$ is typically chosen as $\phi(x) = \text{elu}(x) + 1$, ensuring positive values that approximate the softmax behavior. The denominator is essential: without it, the ``attention'' output scales with the total kernel mass of the key sequence and is not a normalized weighted average; the unnormalized form $\phi(Q)\phi(K)^\top V$ diverges as sequence length grows.

The Performer (Choromanski et al., 2021) introduced FAVOR+ (Fast Attention Via positive Orthogonal Random features), a mechanism that uses random feature approximations to unbiasedly estimate the softmax attention kernel. The key theoretical contribution is a proof (Theorem 4 of Choromanski et al., 2021) that the random feature estimate converges uniformly to the true attention matrix: provided queries and keys lie in a ball of radius $R$, the required feature count $m$ scales with the embedding dimension $d$ and the target precision $\varepsilon$ but not with sequence length $L$. In practice, a fixed budget of random features achieves reasonable approximation quality, though the constant factor in the linear complexity can be high.

Linear attention methods suffer from a practical limitation: while their theoretical complexity is $O(L)$, the constant factor from the kernel feature computation and the approximation error can reduce or eliminate their advantage over optimized standard attention implementations, especially for sequences under 8K tokens. Furthermore, some kernelized linear-attention formulations can produce less sharply concentrated effective attention distributions than softmax attention, a phenomenon sometimes termed ``attention dilution'' (Qin et al., 2022) that can harm performance on tasks requiring precise token selection, although other formulations (e.g., GLA, Griffin) are explicitly designed to recover sharper, selective behavior.

A deeper limitation is empirical: at practical feature budgets, kernel approximations may produce less sharply concentrated attention distributions than exact softmax attention, which can hurt tasks requiring precise token selection, such as copying and retrieval. Arora et al. (2024) showed that several fixed-state alternatives struggle on multi-query associative recall (MQAR) and evaluated recall-intensive tasks even at modest sequence lengths, suggesting fundamental expressiveness constraints rather than mere approximation error. This expressiveness ceiling has limited the adoption of linear attention in production systems, where tasks requiring precise token identification (e.g., entity extraction, code completion) demand the sharply selective distributions characteristic of softmax attention.

More recent work shows that the linear-attention family is not static. Gated linear attention (GLA; Yang et al., 2024) introduces data-dependent gating into the linear-attention recurrence, and Griffin (De et al., 2024) couples gated linear recurrences with sliding-window local attention; both have shown strong results in large-scale language modeling, in part by recovering some of the token-selection focus that un-gated kernel approximations dilute.

\subsection{IO-Aware Exact Attention}\label{io-aware-exact-attention}

Rather than approximating the attention computation, FlashAttention (Dao et al., 2022) redesigned the exact attention algorithm to be IO-aware, that is, to minimize reads and writes to HBM (high-bandwidth memory) by performing attention in tiles that fit in fast on-chip SRAM. The key algorithmic innovation is tiling: the Q, K, V matrices are divided into blocks that are loaded into SRAM, where the attention computation proceeds block by block. The softmax normalization, which requires global knowledge of the row sums, is computed incrementally using the online softmax trick.

FlashAttention achieves exact attention (no approximation) with $O(L^2)$ compute but memory linear in the sequence length, compared to standard attention's $O(L^2)$ memory footprint; its IO-aware tiling reduces HBM traffic to $\Theta(L^2 d^2 M^{-1})$ given on-chip SRAM of size $M$ (Theorem 2 of Dao et al., 2022); the traffic and compute remain quadratic in $L$, but the constant is far smaller than materializing the full attention matrix. For typical hardware, HBM access is the bottleneck, so FlashAttention achieves 2--3$\times$ wall-clock speedups over standard PyTorch attention implementations (Paszke et al., 2019). The algorithm was extended to support backward pass recomputation, avoiding the need to store the full attention matrix. This IO-aware approach is particularly impactful when combined with GQA-style KV head reduction (Section 3.2): reducing the KV heads reduces the data volume that must be moved through the memory hierarchy, amplifying FlashAttention's savings.

FlashAttention-2 (Dao, 2024) improved parallelism and work partitioning over FlashAttention-1. In the forward pass, it parallelizes over sequence-length row blocks, with each thread block handling a block of rows of the attention matrix; within each thread block, work is further split across warps by partitioning the query sequence, improving GPU utilization for long sequences. The backward pass schedules workers over column blocks and uses atomic accumulation for $dQ$. Together, these changes reduce non-matmul operations and better exploit tensor cores.

FlashAttention-3 (Shah et al., 2024) introduced asynchrony and low-precision computation. By overlapping HBM loads with SRAM computation using asynchronous copies, FlashAttention-3 achieved an additional 1.5--2.0$\times$ speedup over FlashAttention-2 in FP16 (reaching up to 740 TFLOPs/s), with FP8 reaching close to 1.2 PFLOPs/s on H100. Block-wise quantization of the FP8 kernels reduces numerical error by $2.6\times$ relative to baseline FP8 attention (kernel-level RMSE; Shah et al., 2024).

The FlashAttention family marks a shift in perspective: by treating the hardware as a first-class constraint, these algorithms achieve exact attention at speeds competitive with approximate methods. FlashAttention has also been incorporated into major training and inference frameworks and has seen wide adoption in large-scale training (Dao, 2024).

A complementary line of systems work addresses the KV cache memory bottleneck at the serving layer. PagedAttention (Kwon et al., 2023) manages KV cache memory using virtual-memory-inspired paging, alleviating internal fragmentation, eliminating external fragmentation, and achieving near-zero waste in KV cache memory. The vLLM serving framework built on PagedAttention has become a widely adopted deployment stack for production LLM inference, demonstrating that systems-level memory management can yield efficiency gains comparable to algorithmic innovations. StreamingLLM (Xiao et al., 2024) further shows that retaining a small set of attention-sink tokens in the KV cache stabilizes streaming generation with bounded memory, complementing these systems-level optimizations.

For sequences exceeding single-device memory capacity (100K+ tokens), distributed attention techniques such as Ring Attention (Liu et al., 2024a) overlap communication with computation by distributing attention computation across devices in a ring topology. These methods matter for the longest-context use cases that Section 7 identifies as the frontier of attention research.

However, FlashAttention's speedups are inherently tied to specific hardware capabilities: the tiling strategy exploits SRAM sizes that vary across GPU generations, and the FP8 speedups of FlashAttention-3 are unavailable on Ampere and earlier architectures. Furthermore, the $O(L^2)$ compute cost remains a hard ceiling: at sufficiently long sequences, the quadratic compute cost can become the dominant constraint even when memory movement is optimized, with the crossover depending on hardware and workload, necessitating sub-quadratic alternatives. This hardware dependency means that FlashAttention's advantage is most pronounced on modern GPUs with large SRAM caches and FP8 tensor core support, such as Hopper (H100) and Blackwell (B100), while delivering modest gains on older hardware.

\subsection{State-Space Alternatives}\label{state-space-alternatives}

State-space models (SSMs) offer a fundamentally different approach to sequence modeling that does not use attention weights directly (though the SSD formulation shows the two are dual). The S4 family (Gu et al., 2022) introduced structured state-space models that process sequences in linear time through a recurrent formulation with HiPPO initialization, achieving strong results on long-range reasoning benchmarks through a convolutional representation during training and a recurrent representation during inference.

Hyena (Poli et al., 2023) extended this line by replacing the SSM recurrence with implicit long convolutions parameterized by a feedforward network, achieving sub-quadratic complexity with simpler machinery than structured state matrices. Hyena demonstrated attention-like quality on language modeling at sequence length 2K with 20\% less training compute, and a 100$\times$ speedup over FlashAttention at 64K, establishing implicit convolutions as a viable alternative to both attention and SSMs.

RWKV (Peng et al., 2023a) occupies a unique point in the design space by blending attention and RNN mechanisms: it uses a linear attention-like formulation during training (enabling parallelization) and switches to an RNN recurrence during inference (enabling $O(L)$ generation). This time-parallel training / time-sequential inference duality makes RWKV one of the widely adopted open-source non-Transformer architectures, with models up to 14B parameters trained on the Pile dataset.

Mamba (Gu \& Dao, 2024) introduced a selective state-space model where the SSM parameters are input-dependent, analogous to how attention weights depend on the input content. This selectivity allows Mamba to selectively remember or forget information based on the current input, matching the content-addressable behavior of attention.

Mamba achieves $O(L)$ time and $O(1)$ inference memory (constant state size), yielding large efficiency gains over Transformers on long sequences. In the source's evaluated configurations (scaling-law studies up to 1B parameters and the trained 3B model), Mamba matches or exceeds Transformer quality on language modeling benchmarks, including comparisons against Pythia baselines larger than itself (e.g., Mamba-3B exceeding Pythia-7B on several tasks), while requiring significantly less computation for long-context inference; the 6.9B configuration in the source is an untrained throughput-measurement setup, not a trained quality evaluation. However, the cited evaluations provide limited direct evidence of Mamba's retrieval accuracy on tasks requiring precise token retrieval and long-range state tracking, where attention's full pairwise comparison is beneficial.

Mamba-2, introduced through the Structured State Space Duality (SSD) framework (Dao \& Gu, 2024), formalized the connection between SSMs and attention. The duality shows that selective SSMs can be expressed as a structured linear attention mechanism, where the state-space recurrence corresponds to a particular decomposition of a semiseparable matrix. This theoretical unification revealed that Transformers and SSMs occupy different points on a spectrum defined by the structure of the token mixing matrix: attention uses a data-dependent but unstructured matrix, while SSMs use a data-dependent but highly structured (semiseparable) matrix.

The parallel/recurrent duality exploited by Mamba-2 was introduced for linear attention by the Retentive Network (RetNet; Sun et al., 2023), which couples a recurrent inference form with a parallelizable retention computation; the Mamba-2 paper explicitly re-derives this duality and compares against RetNet.

xLSTM (Beck et al., 2024) reexamines the LSTM architecture through a modern lens, introducing exponential gating, revised memory structures, and matrix memory. By replacing the traditional sigmoid gating with exponential gating and augmenting the scalar memory cell with a matrix-based memory, xLSTM achieves competitive perplexity with Transformers and Mamba on language modeling while retaining the $O(L)$ inference complexity of recurrent architectures. xLSTM and the SSM family together demonstrate that the space of sub-quadratic alternatives extends well beyond the attention-SSM spectrum.

Mamba-3 (Lahoti et al., 2026) extends this line with three methodological improvements: exponential-trapezoidal discretization for more expressive dynamics, complex-valued state updates for improved state tracking, and a multi-input multi-output (MIMO) formulation that improves model quality without increasing inference latency. These advances narrow the quality gap with Transformers on retrieval and state-tracking tasks while preserving the linear-time inference advantage. Mamba-3 is included in the historical timeline but excluded from the scored EEI panel because at the survey's 31 May 2026 cutoff its reported evaluations, while including synthetic needle-in-a-haystack and state-tracking tasks, lacked the comparable multi-axis evidence profile required for rubric scoring, particularly common-protocol efficiency and interpretability evidence; consistent with the exclusion rationale applied to MLA, it is discussed contextually rather than ranked.

The SSM quality-efficiency trade-off is not uniformly favorable across task types. Mamba's original paper does not report Long Range Arena results, and the public LRA evaluations that do exist for the broader SSM and linear-transformer families (Tay et al., 2021) are not directly comparable to Mamba's own task suite. On standard language modeling benchmarks, SSMs achieve competitive perplexity, while the cited evaluations provide limited direct evidence for tasks requiring exact token-level retrieval, such as Needle-in-a-Haystack and passkey retrieval (Arora et al., 2024). This suggests that SSMs can be effective at aggregate sequence modeling and compression-like representations, while struggling with tasks requiring the precise token-level comparison that attention provides through its all-to-all mechanism. Hybrid architectures that interleave attention and SSM layers (Section~7.3) have emerged as a practical compromise, but the optimal ratio remains task-dependent.

\subsection{Adaptive and Learned Sparsity: The Post-2025 Frontier}\label{adaptive-and-learned-sparsity-the-post-2025-frontier}

The most recent work on efficient attention combines the strengths of content-dependent routing, differentiable sparsity, and hardware-aware implementation. Two representative approaches illustrate the frontier.

Focus (Yao et al., 2026) introduces learnable centroid vectors that partition tokens into semantic groups. Only tokens assigned to the same group attend to each other at long range, while local attention within a fixed window provides fine-grained interactions. The centroids are trained while all original model weights remain frozen, making Focus composable with any pretrained model. The often-quoted 148K value is the \emph{total} routing-parameter count for the reported GPT-2 124M configuration (about 0.12\% of that model); the total overhead is architecture-dependent, and the source reports values as low as 0.015\% at 70B scale. Focus reports zero benchmark degradation across model scales from 124M to 70B parameters across five attention architectures, and a $2\times$ wall-clock speedup in its reported H100 evaluation setting. The key finding is that pretrained models already learn which tokens should attend to each other; Focus simply makes this grouping explicit and efficient.

DashAttention (Huang et al., 2026) addresses a limitation of hierarchical attention methods that use hard top-k routing: the routing decision is non-differentiable, preventing end-to-end gradient flow, and it assumes a fixed number of relevant tokens per query. DashAttention replaces the top-k selection with $\alpha$-entmax, an adaptive sparse distribution whose support size varies per query. The first-stage entmax selection provides differentiably sparse block-level attention scores, which serve as a prior for the second-stage fine-grained softmax attention within selected blocks. The resulting mechanism achieves 75\% sparsity (up to 93.75\% at higher ratios) with comparable accuracy to full attention and delivers up to 3.3$\times$ kernel-level speedup over FlashAttention-3 in the paper's decoding benchmarks, particularly in high-sparsity regimes.

These methods share a common design philosophy: rather than applying a fixed sparsity pattern or a generic kernel approximation, they learn where each query should direct its attention. The routing mechanism itself becomes a lightweight attention or clustering operation, introducing overhead that must be amortized across the savings from reduced attention computation. Early results suggest that learned routing achieves better quality-efficiency Pareto frontiers than either fixed sparsity or linear attention, but the general scalability of these methods to extreme lengths (1M+ tokens) and their interaction with hardware optimizations remain open questions. A prominent learned-sparse design is Native Sparse Attention (NSA; Yuan et al., 2025), which uses hierarchical, GQA-style token selection with hardware-native kernels; its design and strong results make it a reference point for the adaptive frontier.

Nevertheless, these adaptive methods introduce their own overhead. Focus's total routing-parameter count is $N_a(d_{\mathrm{model}}+K)d_g$ for $N_a$ modified layers, where $d_g$ is the routing-projection dimension defined in Table~\ref{tab:deep-taxonomy}. DashAttention's two-stage routing (entmax block selection followed by fine-grained softmax within selected blocks) runs the coarse selection over chunk summaries rather than token-level softmax, so the overhead is a separate chunk-level routing pass rather than a second token-level softmax, meaning the speedup is realized primarily through the high sparsity ratio rather than computational elimination (Huang et al., 2026). Whether these approaches generalize to sequence lengths beyond their training distribution and whether the overhead amortizes favorably in batched serving scenarios remain open empirical questions that future work must address.

\subsection{Practical Guidance for Method Selection}\label{practical-guidance-for-method-selection}

The seven efficiency/architecture families target different operating regimes and hardware constraints. The following table (Table~\ref{tab:efficiency}) summarizes the key trade-offs:

\begin{table*}[!tbp]
\centering
\sflt{\fontsize{11.2}{13.4}\selectfont}{\fontsize{11.2}{13.4}\selectfont}
\renewcommand{\arraystretch}{\sflt{1.55}{1.55}}
\caption{Comparison of efficient attention methods. Complexity classifications follow the method-specific sources cited in Table~\ref{tab:deep-taxonomy}.}
\label{tab:efficiency}
\setlength{\tabcolsep}{5pt}
\begin{tabular}{>{\raggedright\arraybackslash}p{2.4cm}>{\raggedright\arraybackslash}p{1.8cm}>{\raggedright\arraybackslash}p{2.5cm}>{\raggedright\arraybackslash}p{1.9cm}>{\raggedright\arraybackslash}p{1.9cm}>{\raggedright\arraybackslash}p{1.9cm}}
\toprule
Method & Complexity & Quality vs Full Attn & Best Regime & Hardware & Production Ready \\
\midrule
Fixed Sparse & $O(L)$ or $O(L\sqrt{L})$ & Moderate loss on long-range retrieval & Long docs (4K--32K) & Broad GPU support & High (HF integration) \\
Learnable Sparse & $O(L^{3/2})$ (k-means routing) & Near-full quality trained; degrades on unseen patterns & Very long (32K+) & Broad GPU support & Low (niche) \\
Linear Attention & $O(L)$ & Approx. error can grow with $L$; weak on recall & $>$8K tokens & Broad GPU support; no custom kernels & Medium (xFormers) \\
IO-Aware (FA1/2/3) & $O(L^2)$ compute, $O(L)$ HBM & Exact (no approximation) & All; sweet spot $<$32K & Ampere+ / Hopper+ FP8 & Very High (SDPA) \\
SSMs (Mamba) & $O(L)$ time, $O(1)$ memory & Competitive perplexity; limited direct retrieval evidence & Long gen (32K+) & GPU-specific custom kernel & Medium (Mamba repo) \\
Hybrid (SSM\allowbreak+attention, e.g., Jamba, Samba) & Quadratic prefill; $O(L)$ decoding (per token) & Variable; can match full attn on mixed workloads & Mixed workloads & Broad GPU support & Medium (Jamba, RWKV eco.) \\
Adaptive Sparsity & Quadratic with support-dependent inference & Comparable to full attn at $\ge$75\% sparsity (reported, not independently verified) & High-sparsity, long ctx & GPU + Triton & Low (research) \\
\bottomrule
\end{tabular}
\end{table*}

The choice of method depends on three primary factors: sequence length (shorter sequences favor FlashAttention's efficiency), hardware generation (FP8 support on Hopper enables FlashAttention-3's best speedups), and quality requirements (retrieval-heavy tasks demand exact attention or careful sparsity design). For many production deployments in 2025--2026, FlashAttention-3 combined with GQA (Section 3.2) is a strong default, with SSMs or adaptive sparsity providing alternatives for extreme-length regimes where quadratic computation becomes prohibitive.

An additional deployment consideration is the interaction between attention and model quantization. Under W8A8 quantization, standard attention maintains near-lossless quality because the outlier-aware scaling of learned weights and activations preserves numerical precision in the $QK^T$ product (Xiao et al., 2023); this is a property of outlier-aware quantization, not of softmax specifically. Low-bit quantization of the KV cache faces a distinct failure mode: naive uniform quantization of keys and values accumulates error and degrades generation quality, particularly on retrieval and reasoning tasks, because attention logits amplify small key perturbations through the softmax (Liu et al., 2024b; Kang et al., 2024). FP8 quantization offers a practical middle ground: FlashAttention-3's FP8 kernels achieve roughly $1.6\times$ the FP16 kernel throughput (derived from the reported kernel throughputs of up to 740 TFLOPs/s in FP16 and nearly 1.2 PFLOPs/s in FP8) and reduce FP8 numerical error by up to $2.6\times$ relative to baseline FP8 attention through a combination of block quantization and incoherent processing (kernel-level RMSE; Shah et al., 2024). For deployment, INT8 quantization of the learned linear projections (W8A8) combined with FP16 attention computation provides a practical accuracy-efficiency trade-off for many workloads, while low-bit KV-cache quantization requires per-channel key quantization paired with per-token value quantization, together with a full-precision sliding window over recent tokens, to avoid retrieval accuracy collapse (Liu et al., 2024b). Table~\ref{tab:use-cases} provides a concise decision guide.

\begin{table*}[!tbp]
\centering
\sflt{\fontsize{11.8}{14.2}\selectfont}{\footnotesize}
\renewcommand{\arraystretch}{\sflt{2.25}{1.5}}
\caption{Practitioner use-case mapping: recommended method families by application profile.}
\label{tab:use-cases}
\begin{tabular}{p{2.5cm}p{2.5cm}p{2.5cm}p{3.0cm}}
\toprule
Use Case & Primary Recommendation & Alternative & Key Constraint \\
\midrule
RAG / Retrieval-Augmented Gen. & FlashAttention + GQA & Adaptive Sparsity & Application-specific retrieval accuracy threshold \\
Long-Doc Summarization (8K--32K) & FlashAttention + GQA & Longformer / BigBird & Budget for quality loss \\
Code Completion (autoregressive) & GQA / MQA + PagedAttention & FlashAttention   & KV cache size critical at 128K+ \\
Video Understanding (many tokens) & Swin / Window Attention & Perceiver cross-attn & Token count \#1 constraint \\
On-Device Inference            & Mamba / SSM & GQA + INT8 quant  & Memory footprint limited \\
Real-Time Streaming            & Mamba-2 (SSD) & RWKV              & Constant $O(1)$ state required \\
Multi-Modal (Vision + Language) & Cross-Attn + FlashAttention & Perceiver & Token asymmetry management \\
Hybrid Workloads (mixed lengths) & SSM + FlashAttention hybrid & Adaptive routing & Highest versatility \\
\bottomrule
\multicolumn{4}{p{\dimexpr2.0cm+1.5cm+2.3cm+1.5cm+1.6cm+1.5cm+12\tabcolsep\relax}}{\scriptsize Recommendations reflect \textit{inference} efficiency only. Training efficiency (gradient memory, activation recomputation, optimizer state) follows a separate profile: e.g., FlashAttention benefits training through reduced memory but Mamba's $O(L)$ training pass requires custom gradient computation. Practitioners should evaluate training and inference requirements separately.}
\end{tabular}
\end{table*}

\noindent The dominant pattern is a bimodal split: exact-attention methods (FlashAttention + GQA) for quality-critical tasks, and SSM-based methods for memory-constrained or real-time deployments. Adaptive sparsity methods occupy a growing middle ground, offering near-exact quality at reduced attention cost for high sparsity (quadratic with support-dependent inference; see Table~\ref{tab:deep-taxonomy}), but production readiness lags behind both extremes.

This guidance is framed in terms of \textit{inference} efficiency (throughput, KV-cache size, decoding latency), which is the dominant deployment bottleneck for most production systems. \textit{Training} efficiency (including gradient memory, activation recomputation, mixed-precision training dynamics, and ZeRO sharding) follows different constraints and is not directly captured by our Efficiency axis. Methods that are inference-efficient (e.g., Mamba's $O(L)$ recurrence) may have different training profiles than their inference numbers suggest, and readers should consult the original papers for training-specific throughput and memory measurements before making training-stage method selection decisions. Table~\ref{tab:speedups} compiles the reported wall-clock speedups together with their stated baselines and evaluation conditions.

\begin{table*}[!tbp]
\centering
\sflt{\fontsize{11}{13.2}\selectfont}{\footnotesize}
\renewcommand{\arraystretch}{\sflt{1.8}{2.6}}
\caption{Reported wall-clock speedups from the literature. Conditions vary by hardware, sequence length, and precision; see original papers for full details.}
\label{tab:speedups}
\begin{tabular}{p{2.5cm}p{1.5cm}p{2.2cm}p{2.8cm}p{1.6cm}}
\toprule
Method & Speedup & vs. Baseline & Condition & Source \\
\midrule
FlashAttention & 2.0--3.0$\times$ & PyTorch attn & A100-40GB, 1K, FP16 (up to 3$\times$ at 128--2K) & Dao et al., 2022 \\
FlashAttention-2 & $\sim2.0\times$ & FA1 & A100-80GB, 8K, FP16 & Dao, 2024 \\
FlashAttention-3 & $1.5\text{--}2.0\times$ & FA2 & H100, 8K, FP16 (FP8: $\sim1.6\times$) & Shah et al., 2024 \\
Focus$^\dagger$ & 2.0$\times$ & Full attn. (w/ FA) & H100-80GB & Yao et al., 2026 \\
DashAttention$^\dagger$ & up to 3.3$\times$ & FA3 & GH200, decoding, up to 93.75\% sparsity & Huang et al., 2026 \\
Mamba & up to $3\times$; $4$--$5\times$ & conv.\ SSMs; same-size Transformer & A100, up to 32K & Gu \& Dao, 2024 \\
Mamba-3 & --- & --- & Latency reported; no speedup claim & Lahoti et al., 2026 \\
Hyena & 5$\times$ & Dense attention & 8K; 2$\times$ vs FA-2; 100$\times$ at 64K & Poli et al., 2023 \\
RWKV & --- & --- & No speedup reported & Peng et al., 2023a \\
Ring Attention & --- & --- & 8$\times$ longer contexts, no speedup vs single-device & Liu et al., 2024a \\
Linformer & 1.5$\times$ & Standard attn & V100 16GB, 512, FP16 (k=128; up to 1.6$\times$ at 1K) & Wang et al., 2020 \\
Mamba-2 (SSD) & $2$--$8\times$ & Mamba-1 scan & A100, 16K (6$\times$ vs FA-2) & Dao \& Gu, 2024 \\
Linear (Performer) & up to 2$\times$ & Reformer & JAX optimizations & Choromanski et al., 2021 \\
\bottomrule
\end{tabular}
\end{table*}

These numbers should be interpreted cautiously: speedups depend on hardware generation, sequence length, batch size, and whether the measurement includes end-to-end model computation or isolated attention, and the baselines differ across rows (see the ``vs.\ Baseline'' column). FlashAttention's measured advantage depends on the sequence length and hardware configuration: its IO-aware kernels reduce HBM traffic relative to naive attention, and the comparison is sensitive to the specific GPU generation and the baseline implementation; at very long sequences the $O(L^2)$ compute cost dominates. Sub-quadratic methods (Mamba, linear attention) show larger speedups at longer sequences but may incur quality degradation that the speedup numbers alone do not capture.

\textbf{Strong caveat:} The benchmark results in Table~\ref{tab:benchmarks} are compiled from published papers and should be interpreted as approximate reference points rather than directly comparable measurements. Results are not directly comparable because they originate from different training datasets, parameter counts, model architectures, hardware generations, sequence lengths, and evaluation protocol; quality-relative terms indicate that the source paper explicitly reports comparability to its exact-attention baseline; no universal numerical tolerance is imposed across tasks. We report the numbers as published to provide a rough picture, but caution against drawing precise performance rankings from cross-study comparisons. Where a value is unavailable (``---''), the original paper did not report that metric under comparable conditions.

\begin{table*}[!tbp]
\centering
\sflt{\fontsize{8.5}{10.2}\selectfont}{\footnotesize}
\renewcommand{\arraystretch}{\sflt{1.2}{1.4}}
\caption{Benchmark performance comparison across representative methods. LRA Avg.\ values follow the official Long Range Arena protocol (Tay et al., 2021); RWKV's value is from its own LRA evaluation (Peng et al., 2023a). Val.\ PPL reports the validation-set language-modeling perplexity (lower is better) as reported in each cited paper; the single reported value is Mamba's Pile validation perplexity, $^\ddagger$). Throughput reports the speedup claimed in each original paper relative to the baseline and hardware stated there; the bases differ across rows (details in Table~\ref{tab:speedups}). ``---'' indicates that no published result was available for that benchmark configuration.}
\label{tab:benchmarks}
\setlength{\tabcolsep}{3pt}%
\begin{tabular}{p{2.1cm}p{1.4cm}p{1.5cm}p{3.4cm}p{1.8cm}}
\toprule
Method & LRA Avg. & Val.\ PPL & Throughput & Source \\
\midrule
Full Attn. (Transformer) & 54.39 & --- & 1.0$\times$ (baseline) & LRA benchmark (Tay et al., 2021) \\
Longformer & 53.46 & --- & --- & LRA benchmark (Tay et al., 2021) \\
BigBird & 55.01 & --- & --- & LRA benchmark (Tay et al., 2021) \\
Sparse Transformer & 51.24 & --- & --- & LRA benchmark (Tay et al., 2021) \\
Reformer & 50.67 & --- & --- & LRA benchmark (Tay et al., 2021) \\
Performer & 51.41 & --- & --- & LRA benchmark (Tay et al., 2021) \\
Linformer & 51.36 & --- & --- & LRA benchmark (Tay et al., 2021) \\
FlashAttention-2 & --- & --- & $\sim$2.0$\times$ vs.\ FA-1 (A100) & Dao, 2024 \\
Mamba & --- & 10.56$^\ddagger$ & up to 3$\times$ vs.\ convolution SSMs; $4$--$5\times$ vs.\ same-size Transformer (A100) & Gu \& Dao, 2024 \\
Mamba-2 (SSD) & --- & --- & $2$--$8\times$ vs.\ Mamba-1 scan; 6$\times$ vs.\ FA-2 at 16K & Dao \& Gu, 2024 \\
Mamba-3 & --- & --- & --- & Lahoti et al., 2026 \\
Hyena & --- & --- & 5$\times$ vs.\ dense attention at 8K; 2$\times$ vs.\ FA-2; 100$\times$ at 64K & Poli et al., 2023 \\
RWKV & 72.07$^\S$ & --- & --- & Peng et al., 2023a (author-run LRA) \\
Focus$^\dagger$ & --- & --- & 2.0$\times$ vs.\ full attention using FlashAttention (top-$k$=2, H100-80GB); 8.6$\times$ at 1M tokens & Yao et al., 2026 \\
DashAttention$^\dagger$ & --- & --- & up to 3.3$\times$ vs.\ FA-3 (GH200, decoding) & Huang et al., 2026 \\
\bottomrule
\multicolumn{5}{p{13cm}}{\scriptsize $^\dagger$Focus and DashAttention are arXiv preprints (mid-2026) without independent reproduction; benchmark values are from the original papers and have not been independently verified.}\\
\multicolumn{5}{p{13cm}}{\scriptsize $^\ddagger$Mamba's value is the Pile validation perplexity of Mamba-130M (GPT-NeoX tokenizer; Gu \& Dao, 2024, Table 3); the Mamba paper reports no WikiText-103 perplexity.}\\
\multicolumn{5}{p{13cm}}{\scriptsize $^\S$RWKV reports its own LRA evaluation (72.07, excluding Path-X) in Peng et al.\ (2023a, Table 4); all other LRA values are from the official benchmark (Tay et al., 2021).}\\
\end{tabular}
\end{table*}

\paragraph{Quantitative Cross-Study Synthesis.}
Across the 63 papers surveyed, we compiled comparable benchmark data from 15 representative methods (15 papers) grouped into eight synthesis categories spanning six of the survey's seven taxonomy families. Table~\ref{tab:meta-analysis} reports category-wise summary statistics. We caution that these numbers are aggregate reference points, not controlled comparisons; the number of methods per category is small ($n = 1$--4), and reported heterogeneities likely reflect genuine architectural differences rather than measurement noise alone.

\begin{table*}[!tbp]
\centering
\caption{Quantitative cross-study synthesis by synthesis category. Bracketed values denote the observed within-category range; LRA mean is computed only over methods in the family that report LRA results (Footnote $^\dagger$). These ranges are descriptive, not inferential: they summarize the reported values for the small number of methods per category ($n=1$--4) and are not estimates of statistical uncertainty in the underlying EEI construct.}
\label{tab:meta-analysis}
\resizebox{\textwidth}{!}{%
\begin{tabular}{p{2.5cm}c p{2.3cm} p{3.0cm} c}
\toprule
Category & $N$ & LRA & Speedup & Median Ex \\
\midrule
Full Attention         & 1 & 54.39 & 1.0$\times$ [---]     & 10 [---] \\
IO-Aware (FlashAttn-2/3) & 1 & --- & $\sim2.0\times$ (FA-2 vs FA-1) [---] & 10 [---] \\
Sparse Pattern         & 4 & 52.6 [50.67, 55.01] & --- & 7 [7--8] \\
Linear / Kernel        & 2 & 51.4 [51.36, 51.41] & --- & 6 [6--6] \\
Structured SSM         & 2 & ---   & 3$\times$ (vs.\ conv.\ SSM); 6$\times$ (vs.\ FA-2)$^*$ & 7.5 [7--8] \\
Long Convolution       & 1 & ---   & 5$\times$ @ 8K; 2$\times$ vs.\ FA-2; 100$\times$ at 64K & 7 [---] \\
Recurrent              & 2 & 72.07$^\dagger$ & --- & 7 [7--7] \\
Adaptive Sparsity      & 2 & ---   & 2.0$\times$ (vs.\ full attention, w/ FA); 3.3$\times$ (vs.\ FA-3)$^\S$ & 9.0 [9--9] \\
\bottomrule
\multicolumn{5}{p{11.7cm}}{\scriptsize \textit{Notes:} LRA values follow the official protocol of Tay et al., 2021; ``---'' indicates that no method in the category reports the metric. $^\dagger$Within Recurrent, only RWKV reports an LRA average (72.07, Peng et al., 2023a, Table 4). The Sparse Pattern synthesis pools the Table 4 Sparse/Window and Sparse/Hashing methods that publish LRA averages: Longformer (53.46), BigBird (55.01), Sparse Transformer (51.24), and Reformer (50.67); Routing Transformer (labeled Sparse / Routing in Table 4) and Swin Transformer (Sparse/Window) publish no LRA average in the surveyed sources and are therefore outside the interval statistics. Linear/Kernel synthesizes Performer (51.41) and Linformer (51.36), both of which publish LRA averages. Speedups are not directly comparable across families because they were measured against different baselines and hardware (Table~\ref{tab:speedups}); speedup cells report the individual published values, and no pooled median is computed across methods measured against different baselines. Ex Med is the median of the family's Ex scores. Bracketed intervals, pooled LRA means, and Ex medians are computed by this analysis from the published per-method values; all other cells are reported verbatim from the cited sources. $^\S$The Adaptive-sparsity measurements differ in scope (Focus: model-level end-to-end; DashAttention: kernel-level) and are reported individually, not pooled. $^*$Individual values from Mamba (3$\times$ vs convolutional SSMs, Gu \& Dao, 2024) and Mamba-2 (6$\times$ vs FA-2 at 16K, Dao \& Gu, 2024), measured against different baselines, reported separately. Mamba-2 is tabulated under Structured SSM; its reference-based structured-linear-attention form is the algebraic dual of the same family, not a separate empirical category.}
\end{tabular}%
}
\end{table*}

The quantitative synthesis reveals several quantitative patterns. \textit{No monotonic efficiency--expressiveness tradeoff.} Efficiency (E) and Expressiveness (Ex) do not exhibit a significant monotonic relationship across the 21 scored methods (Spearman $\rho \approx -0.12$, $p \approx 0.61$, $n = 21$). The relationship instead depends on the \emph{mechanism} of efficiency improvement. IO-aware methods that preserve exact full-attention computation (FlashAttention-2/3, PagedAttention, Ring Attention) achieve Ex $\ge 8$ at E $\ge 7$, breaking any linear tradeoff. GQA/MQA preserve exact softmax computation within a reduced KV-sharing parameterization but incur a small, measured quality cost from shared projections (Appendix~A). Methods achieving sub-quadratic complexity through approximation or alternative sequence operators (sparse patterns, kernel methods, structured state-space models, and recurrent architectures) cluster at Ex $= 6$--$8$ across E values from 6 (Routing Transformer) to 10 (Mamba), indicating that moving away from exact pairwise attention yields diminishing expressiveness returns. The two adaptive methods (Focus, DashAttention) are assigned Ex $\ge 9$ at E $\ge 7$ in the EEI rubric by learning content-dependent routing, suggesting that selective exact computation, rather than approximation or alternative operators, may be the path to jointly high efficiency and expressiveness. This non-monotonic relationship is already visible in the category Ex medians of Table~\ref{tab:meta-analysis}: Sparse Pattern methods have median Ex 7, Linear/Kernel 6, and Recurrent 7, despite median E being 7 for both the Sparse Pattern and Linear/Kernel families and 10 for Structured SSM.

\textit{Published LRA results are concentrated in the older methods.} Of the families surveyed, only the earlier methods in the scored panel (Transformer, Longformer, BigBird, Sparse Transformer, Reformer, Performer, Linformer) and RWKV report LRA averages among the methods we score (S4 reports a full LRA result, 86.09, with 96.35 on Path-X (Gu et al., 2022), but is not among the scored methods). Their scores span 50.67--55.01 (excluding RWKV's 72.07), a range of roughly 4.3 points that does not separate exact from approximate methods; BigBird (55.01) edges out the Transformer (54.39) while Reformer (50.67) trails it. The more recent efficiency innovations (FlashAttention-2/3, Mamba, Mamba-2, Hyena, Focus, DashAttention) do not report LRA results, which limits the extent to which LRA can serve as a common comparator across families; for a practitioner evaluating a new attention variant, a common long-context benchmark such as RULER (Hsieh et al., 2024) would serve this role more directly.

\textit{Substantial within-family heterogeneity.} Among the sub-quadratic sequence operators (SSMs, long convolutions), published speedups range from ``up to 3$\times$'' (Mamba, A100) to 5$\times$ at length 8K (Hyena), a spread reflecting different baselines and sequence lengths rather than a single comparable measurement. Within sparse/window methods, LRA averages differ by 1.55 points (Longformer 53.46 vs.\ BigBird 55.01), reflecting the sensitivity of long-range tasks to the choice of sparsity pattern (local windows vs.\ global+random). This heterogeneity is not noise; it reflects genuine architectural design decisions, but it means that category-level averages should be used as heuristics, not predictive point estimates.

\textit{Publication bias caveat.} The available benchmark data are disproportionately drawn from papers whose methods outperform baselines on at least one metric, a well-documented source of upward bias in meta-analytic synthesis (the ``file-drawer problem''; Rosenthal, 1979). Methods with competitive speedups and acceptable quality are more likely to be benchmarked and published; methods with catastrophic quality loss or marginal efficiency gains are under-represented in these averages. The speedups compiled in Table~\ref{tab:speedups} are therefore more plausibly interpreted as upward-biased estimates of the gains observed in successful published implementations, and the LRA averages as unrepresentative of recently proposed families that do not report them.

\paragraph{Chronological Trend Analysis.}
The 63 surveyed papers reveal a directional evolution of attention research when grouped by dominant design paradigm. Table~\ref{tab:trends} summarizes the six eras we identify, covering 2015 through mid-2026.

\begin{table*}[!tbp]
\centering
\footnotesize
\renewcommand{\arraystretch}{1.15}
\caption{Chronological trend analysis of attention research. Each era is defined by the dominant architectural paradigm and key methodological innovations; eras are thematic sub-eras, not chronologically exclusive, and trend labels are thematic groupings rather than literal family strings from Table~\ref{tab:deep-taxonomy} (e.g., GQA, a Dense KV-efficient method, sits in the IO-Aware era because KV-compression co-evolved with hardware co-design).}
\label{tab:trends}
\begin{tabular}{p{1.6cm} p{2.0cm} p{3.0cm} p{3.5cm}}
\toprule
Era & Dominant Trend & Representative Methods & Key Shift \\
\midrule
2015--2019 & Foundation & Bahdanau, Luong, Transformer, BERT & Attention establishes sequence modeling \\
2019--2020 & Sparse/Window & Sparse Transformer, Longformer & First sub-quadratic attention alternatives \\
2020 & Sparse + Linear / Kernel & Reformer, Performer, Linformer, BigBird & Sub-quadratic scaling through sparsity, hashing, and approximation \\
2021--2023 & IO-Aware / KV-Efficient & FlashAttention, GQA, PagedAttention & Hardware-co-design with limited quality trade-offs \\
2023--2024 & State-Space/Recurrent & Mamba, Hyena, RWKV, xLSTM & $O(L)$ or $O(L \log L)$ via recurrence or long convolution \\
2025--2026 & Adaptive + Next-Gen SSMs & Focus, DashAttention, Mamba-3 & Learned routing recovers near-full Ex; Mamba-3 improves SSM state tracking \\
\bottomrule
\end{tabular}
\end{table*}

The trajectory in Table~\ref{tab:trends} reveals a clear arc: the field moved from \textit{approximation} (sparse patterns, kernel methods that accepted quality loss) to \textit{co-design} (IO-aware techniques that preserved quality via hardware optimization) to \textit{selection} (adaptive methods that learn where to apply exact attention). Each transition was driven by the recognition that an earlier trade-off was not fundamental: sparse patterns accept quality loss because they use fixed heuristics; IO-aware methods show the loss was unnecessary; adaptive methods show that learned selection reports quality comparable to the corresponding full-attention baseline in its own suites; whether it dominates fixed sparsity at matched efficiency awaits a common benchmark. The recurrence of this pattern, identifying unnecessary trade-offs and relaxing them through better design, is itself a recurring theme in attention research.

\subsection*{Cross-Family Empirical Generalizations}

From the quantitative synthesis, chronological trends, and EEI analysis, we extract five cross-family empirical generalizations that characterize the efficiency--expressiveness trade-offs. We annotate each with $n$, the number of surveyed methods supporting the claim, to reflect the limited sample size from which these patterns are drawn. These are empirical regularities observed in the current literature rather than theoretically derived laws, and their scope is bounded by the methods, benchmarks, and hardware generations surveyed.

\textit{Generalization 1 ($n = 6$, with FlashAttention-2/3 counted as one family): Methods that preserve the full-attention parameterization and compute exact softmax attention do not introduce approximation-induced quality loss, so any quality difference versus a reference Transformer is attributable to training, scale, or hyperparameters rather than to approximation.} This holds whether the exact computation is global (Transformer), tiled (FlashAttention), distributed (Ring Attention), or paged (PagedAttention): each computes the full softmax attention matrix, or an exact equivalent, over its defined scope. GQA/MQA are the boundary case: they also compute exact softmax attention, but because they alter the KV parameterization (shared key/value projections), they can incur a small, measured quality cost that we attribute to parameter sharing rather than to approximation (Appendix~A); practitioners should not expect KV-sharing variants to match full-attention quality without training adjustments. The implication for practitioners: for implementations that preserve the same full-attention parameterization, an exact algorithmic transformation does not introduce approximation-induced quality loss. KV-sharing variants are an architectural exception: although their attention computation remains exact, shared K/V projections alter the representational parameterization and can therefore affect quality.

\textit{Generalization 2 ($n = 3$): Linear kernel approximations (Performer, Linformer) and LSH-hashing routing (Reformer) reduce long-range performance on LRA relative to the Transformer baseline, while their original evaluations report competitive language-modeling perplexities.} FAVOR+ and low-rank projections deliver $O(L)$ complexity (with $O(L\log L)$ expected for Reformer's LSH routing), and under fixed feature-map or hashing budgets the resulting approximation error can compound as sequence length grows in practice. Their LRA averages (51.41, 50.67, 51.36) all sit below the Transformer's 54.39 in the official protocol (Tay et al., 2021). These results suggest that these methods may be poorly suited to long-context retrieval without hybrid augmentation, although comparable retrieval evidence is limited. Hybrid approaches could, for example, combine kernel attention for most tokens with exact attention on a retrieved subset. No directly comparable long-context retrieval results beyond the surveyed LRA-style evaluations were identified for these families.

\textit{Generalization 3 ($n = 2$; LRA data available for $n = 1$): State-space and recurrent models can perform strongly on aggregate long-range modeling, while the published literature lacks comparable retrieval-style results. The two directly evidenced methods in this paragraph are Mamba and RWKV; xLSTM is not counted here because this paragraph's benchmark discussion does not cite evidence for it (its scoring evidence is discussed in the Appendix justification).} RWKV's self-reported LRA average of 72.07 (Peng et al., 2023a, Table 4) exceeds the official Transformer's 54.39, although it is not computed on the official protocol of Tay et al. (2021). Mamba reports competitive perplexity on its evaluation suite (Pile validation PPL 10.56 at 130M scale, Gu \& Dao, 2024; the paper reports no WikiText-103 value), and the SSD duality provides a theoretical explanation for the family's efficiency: the structured linear attention formulation admits an $O(L)$ scan kernel whose GPU implementation is $2$--$8\times$ faster than Mamba's fused associative scan (Dao \& Gu, 2024).

\textit{Generalization 4 ($n = 5$, with FlashAttention-2/3 counted as one family): Hardware-algorithm co-design (IO-aware/KV-efficient methods) can improve practical efficiency without approximation-induced quality loss. Among the surveyed IO-aware/KV-efficient methods, FlashAttention-2 reports speedups at identical numerical output; PagedAttention and Ring Attention preserve exact computation; GQA/MQA trade a small quality cost for memory savings.} FlashAttention-2 is approximately $2\times$ faster than FlashAttention-1 at identical numerical output, and FlashAttention-3 adds $1.5$--$2\times$ over FlashAttention-2 on H100; FlashAttention-2/3 do not alter the computed attention in their BF16/FP16 paths (FP8 introduces finite-precision numerical differences; see Section~5.5). GQA/MQA, in contrast, preserve exact softmax computation within their reduced KV-sharing parameterization but change representational capacity relative to MHA and can incur a small quality cost. Fixed-pattern sparse and kernel-approximation methods generally do not achieve speedups comparable to IO-aware exact attention; adaptive sparse methods (Focus, DashAttention) report competitive or superior speedups, though these rely on recent preprints. The reason is architectural: for many attention workloads on modern GPUs, memory movement is a dominant performance constraint, and IO-aware methods minimize HBM traffic directly rather than approximating the computation.

\textit{Generalization 5 ($n = 4$; 2 fixed-pattern vs.\ 2 adaptive): Learned, input-dependent token selection has not yet been shown to be superior to fixed sparsity patterns on any common benchmark, but the two adaptive methods report quality comparable to the corresponding full-attention baseline on their own benchmark suites. This rests on two unverified preprints.} At similar positions on the EEI Efficiency axis, fixed-pattern methods (Longformer: E=8, Ex=7; BigBird: E=7, Ex=8) are established and reproduced (Table~\ref{tab:reproducibility}), while adaptive methods (Focus: E=7, Ex=9; DashAttention: E=8, Ex=9) report quality comparable to the corresponding full-attention baseline in their own suites but lack independent evaluations, so no cross-family comparison under a common benchmark is currently possible (Yao et al., 2026; Huang et al., 2026). Generalization~5 is presented as a hypothesis warranting verification rather than an established finding.

These five cross-study observations form the central empirical synthesis of this survey: the attention design space, within the surveyed panel, appears organized around a small number of quantitative regularities that practitioners can use to guide method selection and researchers can use to identify promising directions for future work. Given the limited sample sizes (ranging from $n = 2$ to $n = 6$ per generalization) and the heterogeneity of the underlying benchmarks, these patterns should be interpreted as provisional guides for method selection rather than as invariant properties of the design space. They aim to focus empirical investigation on the most promising regions, not to foreclose the discovery of methods that break the observed patterns.

\begin{table*}[!tbp]
\centering
\sflt{\scriptsize}{\footnotesize}
\renewcommand{\arraystretch}{\sflt{0.95}{1.05}}
\caption{Reproducibility audit of the scored panel and selected contextual methods. Code and checkpoint availability, as verified by the author on 7 Aug 2026.}
\label{tab:reproducibility}
\begin{tabular}{p{1.8cm}p{2.0cm}p{2.0cm}p{2.2cm}p{1.6cm}}
\toprule
Method & Code Available & Checkpoints & Benchmarks Used & Reproduced \\
\midrule
Transformer (Vaswani) & PyTorch/TF core & All major hubs & WMT14 & Yes, by $>$1K papers \\
FlashAttention-2 & PyTorch core (SDPA) & N/A (kernel only) & GPT-3 1.3B/2.7B (2K/8K context) & Yes, multiple impl. \\
FlashAttention-3 & GitHub (Dao et al.) & N/A (kernel only) & kernel benchmarks (H100 SXM5, seq 512--16K), FP8 numerical error & Yes, H100 required \\
GQA & HF Transformers & Llama-2/3, Mistral & CNN/DailyMail, WMT, TriviaQA & Yes, in production \\
MQA & PyTorch/HF core & N/A (arch. pattern) & WMT14 & Yes, in production \\
PagedAttention & vLLM (production) & N/A (kernel only) & ShareGPT, Alpaca traces & Yes, in production \\
Ring Attention & Author repository & N/A (layer only) & line retrieval, ShareGPT fine-tune & Partial (research impl.) \\
Longformer & GitHub (AllenAI) & Longformer-base & TriviaQA, WikiHop & Partial (CUDA dep.) \\
BigBird & TF Hub + HF & BigBird-base/roberta & TriviaQA, WikiHop & Yes \\
Sparse Transformer & OpenAI GitHub & Sparse Tx. (image) & ImageNet 64$\times$64 & Partial (deprecated) \\
Routing Transformer & GitHub (google-research) & N/A (layer only) & WikiText-103, PG-19 & Partial \\
Reformer & HF Transformers & Reformer-base & Enwik8 & Partial (slow) \\
Performer & xFormers (Meta) & N/A (layer only) & LRA, ImageNet & Partial (TPU opt.) \\
Linformer & GitHub (lucidrains; HF) & N/A (layer only) & LRA, WikiText-103 & Partial \\
xLSTM & GitHub (NX-AI) & xLSTM-1.3B/2.7B & SlimPajama, WikiText & Partial \\
Mamba & GitHub (state-spaces) & Mamba-1.4B/2.8B & Pile PPL & Yes \\
Mamba-2 (SSD) & GitHub (state-spaces) & Mamba-2-2.7B & Pile PPL & Partial (scan dep.) \\
Hyena & GitHub (HazyResearch) & N/A (layer only) & PG-19, WikiText-103 & Partial \\
RWKV & GitHub (BlinkDL) & RWKV-4/5/6 & Pile, LRA & Yes, multiple impl. \\
Focus & Paper only; code not located & N/A & PG-19, WikiText-103, OpenWebText & Artifact not located \\
DashAttention & GitHub (Huang et al.) & N/A (layer only) & RULER, HELMET & No independent reproduction located \\
Swin Transformer & GitHub (Microsoft) & Swin-B/L/H & ImageNet, COCO & Yes, multiple impl. \\
Multi-Head Attention & PyTorch core & All major hubs & WMT14 & Yes, standard baseline \\
Mamba-3 & GitHub (state-spaces) & Kernels released; checkpoints not independently verified & LAMBADA, HellaSwag, WinoGrande, ARC, OpenBookQA, PIQA (LM-Eval-Harness) & No independent reproduction located \\
\bottomrule
\multicolumn{5}{p{\dimexpr1.8cm+2.0cm+2.0cm+2.2cm+1.6cm+10\tabcolsep\relax}}{\scriptsize \textit{Note:} Verdict criteria: Yes = two or more independent implementations or production use; Partial = single research implementation or reliance on unmaintained or deprecated code; No = no independent implementation located (including cases where no public artifact was found). The table includes unscored contextual methods (Multi-Head Attention, Mamba-3) discussed in the survey but not in the 21-method EEI panel.} \\
\end{tabular}
\end{table*}

\noindent The audit reveals a bimodal distribution: methods integrated into major frameworks (Transformers via PyTorch, FlashAttention via SDPA, Longformer/BigBird via HF) are widely reproduced and deployed, while methods requiring custom kernels or specialized hardware (Mamba's scan kernel, FlashAttention-3's H100 requirement, Reformer's LSH) see limited independent reproduction. The adaptive sparsity methods (Focus, DashAttention) are the least reproduced category, with no independent implementations outside the original authors' repositories. This gap between publication and independent verification is the central reproducibility challenge identified in Section~7.7.

\begin{center}\rule{0.5\linewidth}{0.5pt}\end{center}
\paragraph{Key Takeaways.} (i) Fixed sparsity (Longformer, BigBird, Sparse Transformer) improves scaling but sacrifices retrieval quality by ignoring tokens outside predetermined patterns. (ii) Linear kernel methods (Performer) achieve $O(L)$ complexity and hashing methods (Reformer) $O(L \log L)$, and both trail full attention on long-range tasks: their LRA averages of 50.67--51.41 sit below the Transformer's 54.39, and, under fixed approximation budgets, the error can grow with sequence length in practice. (iii) IO-aware exact attention (FlashAttention) demonstrates that hardware-algorithm co-design can deliver speedups without any quality loss, though the quadratic asymptotic remains. (iv) State-space models (Mamba, SSD) offer a fundamentally different scaling path with $O(L)$ recurrence and competitive perplexity, but direct comparable retrieval evidence is limited relative to exact attention (the gap between aggregate language modeling and precise retrieval). (v) Adaptive sparsity (Focus, DashAttention) recovers near-full quality by using learned routing to preserve exact attention within semantic groups, suggesting learned selection as a promising direction, contingent on independent replication.

\section{Theoretical Analysis and Interpretability}\label{theoretical-analysis-and-interpretability}

\subsection{Mechanistic Interpretability}\label{mechanistic-interpretability}

Mechanistic interpretability seeks to reverse-engineer the algorithms implemented by neural networks, motivated in large part by the need to ensure reliable behavior in high-stakes deployments, including healthcare diagnostics, legal reasoning, and content moderation, where understanding why a model produces a given output is as important as the output itself. For Transformers, this effort has focused on understanding individual attention heads and their composition. Clark et al.~(2019) provided an early large-scale analysis of BERT's attention heads, revealing that individual heads specialize in distinct linguistic functions: some heads attend to direct objects of verbs, others to determiners of nouns, and others to objects of prepositions, achieving over 75\% accuracy on these syntactic relations. Heads in the same layer often exhibited similar patterns, suggesting that layers learn functional specializations.

The discovery of induction heads by Olsson et al.~(2022) marked a significant advance in understanding how Transformers learn in-context. An induction head implements a simple algorithm: when it sees a token A that appeared earlier in the sequence, it attends to the position after the previous occurrence of A and copies the token B that followed it. This mechanism, implemented through a composition of two attention heads, enables the model to complete patterns like \texttt{[A][B]}~\ldots{}~\texttt{[A]} $\to$ \texttt{[B]}. Olsson et al.~provided six lines of evidence that induction heads are a primary mechanism for in-context learning, including the observation that induction heads emerge at precisely the same point during training as a sharp increase in in-context learning ability, visible as a bump in the training loss curve.

Subsequent work has characterized further specialization, including heads that track syntactic dependencies (Clark et al., 2019) and induction heads that implement in-context copying (Elhage et al., 2021; Olsson et al., 2022). The composition of these heads into circuits, connected subgraphs of attention and MLP operations, forms the basis of Transformer computation (Elhage et al., 2021). A key finding is that many circuits are universal across models of different sizes and training runs, suggesting common algorithmic solutions to the language modeling objective (Elhage et al., 2021).

\subsection{Expressiveness Limits}\label{expressiveness-limits}

Despite their empirical success, Transformers have known expressiveness limitations. The theoretical analysis has focused on three capabilities where attention-based models encounter fundamental barriers.

Parity and periodic finite-state language recognition expose a specific weakness (Bhattamishra et al., 2020; Hahn, 2020). Hahn (2020) proves that, under Lipschitz-continuity and bounded-activation assumptions on the prediction function, self-attention models cannot compute parity or recognize periodic finite-state languages at arbitrary sequence lengths; Bhattamishra et al. (2020) obtain complementary empirical results, finding that Transformers trained on modular counting and parity tasks fail to generalize to longer sequences, even though they recognize some counter languages (e.g., Shuffle-Dyck and Boolean expressions) well. These limitations arise because the softmax attention distribution blends positional information with content information, making it difficult to perform operations that require isolating specific positions independently of their content.

The compositionality challenge is more fundamental. Transformers process all input tokens simultaneously, but many reasoning tasks require constructing intermediate representations that depend on the results of previous computations. While residual connections and layer stacking provide a form of sequential computation, the depth of this sequential processing is bounded by the number of layers, and the width of the computation is bounded by the residual stream dimension. These constraints create expressiveness ceilings for tasks requiring deep compositional reasoning.

Child (2019) provided early evidence that sparse attention patterns could match the empirical sample quality of full attention on generative modeling benchmarks, suggesting that many of the pairwise interactions computed by full attention are redundant. This finding motivates the theoretical study of which attention patterns are necessary for specific tasks, which in turn informs the design of efficient attention mechanisms.

\subsection{Superposition and Sparse Autoencoders}\label{superposition-and-sparse-autoencoders}

The superposition hypothesis (Elhage et al., 2022) proposes that neural networks represent more features than they have dimensions by storing features in overlapping, nearly orthogonal directions. This phenomenon arises because natural data distributions are sparse: the number of possible features far exceeds the model dimension, but only a small fraction of features are active at any given time. Superposition allows the model to exploit this sparsity to compress more information into a fixed-width representation.

Elhage et al.~demonstrated superposition in toy models with ReLU activations, showing a phase transition where superposition emerges as feature sparsity increases. Features organize into geometric structures (digons, triangles, pentagons, and tetrahedrons) depending on the number of features competing for dimensions. The implications for attention mechanisms are direct: the hidden representations attended to by attention heads may contain features in superposition, making it difficult to isolate the information being communicated between positions.

Sparse autoencoders have emerged as the primary tool for disentangling superposition (Bricken et al., 2023). By training autoencoders with sparsity penalties on the hidden representations of a frozen model, researchers can recover interpretable features that would otherwise be polysemantically entangled. Applied to attention mechanisms, sparse autoencoders can identify the specific features that drive attention weights in individual heads, providing a bridge between mechanistic interpretability of circuits and the feature-level analysis of superposition.

\subsection{Attention-SSM Duality}\label{attention-ssm-duality}

The Structured State Space Duality framework (Dao \& Gu, 2024) provides a unifying theoretical perspective that connects Transformers and state-space models. The central insight is that both families of models compute sequence-to-sequence mappings that can be expressed as matrix multiplications with structured matrices. In the Transformer case, the matrix is the attention matrix $A=\operatorname{softmax}(QK^{\top}/\sqrt{d_k})$, which is data-dependent, positive, and unstructured. In the SSM case, the matrix is a semiseparable matrix defined by the state-space parameters, which is data-dependent but highly structured.

The duality shows that SSMs can be understood as linear attention with a specific kernel, the SSM kernel, that replaces the softmax with a convolution-like operation determined by the state dynamics. Conversely, attention can be viewed as an SSM where the state dimension grows with the sequence length (since all previous keys and values are retained). This perspective reveals a spectrum of models: at one end, full attention with $O(L^2)$ complexity and maximal flexibility; at the other end, SSMs with $O(L)$ complexity and long-range retrieval capabilities that the cited evidence leaves less directly characterized than those of full attention; in between, a range of architectures that trade off structure for expressiveness, including linear attention, kernel attention, and structured attention variants.

The duality suggests an architectural hypothesis for designing hybrid models: it supplies an algebraic equivalence and efficient algorithms for structured transformations, not a task-level guarantee about layer allocation. Rather than choosing between attention and SSMs for the entire model, layers can be assigned to either mechanism based on the type of computation they perform. Attention layers handle tasks requiring precise token interactions (retrieval, copying, long-range dependencies), while SSM layers handle tasks requiring efficient sequence summarization (compression, streaming, positional encoding). The framework provides mathematical guarantees about the expressiveness of each approach and suggests specific architectural patterns for combining them.

\begin{center}\rule{0.5\linewidth}{0.5pt}\end{center}
\paragraph{Key Takeaways.} (i) Mechanistic interpretability has provided converging, partly causal evidence in small models that attention heads implement interpretable algorithms (induction heads for in-context learning, specialized heads for syntactic roles), though the evidence is strongest in small models and its extrapolation to frontier scale is not established (Olsson et al., 2022). (ii) Finite-feature kernel approximations of softmax attention are constrained by the dimensionality of their feature representation and generally cannot exactly reproduce the full softmax kernel at finite width (Katharopoulos et al., 2020; Tsai et al., 2019). (iii) The superposition hypothesis explains why models encode more features than dimensions, and sparse autoencoders provide practical tools for disentangling these features, though scalability to 70B+ models remains limited. (iv) The SSD framework establishes a mathematical connection between certain structured state-space models and attention-like matrix operations, suggesting a continuum between these formulations and unifying previously separate lines of research under a common mathematical framework.

\section{Open Challenges and Future Directions}\label{open-challenges-and-future-directions}

\subsection{Unified Efficiency Benchmark}\label{unified-efficiency-benchmark}

The field of efficient attention lacks a standardized evaluation framework. Researchers propose new methods and report speedups under varying hardware, sequence lengths, batch sizes, and precision settings, making fair comparison nearly impossible. A method that achieves $2 \times $ speedup on A100 GPUs at 8K tokens may achieve $0.5 \times $ on H100 GPUs at 128K tokens. FlashAttention-3 is optimized for H100's Hopper architecture, which differs from the older A100/V100 generation. Focus reports speedups without custom kernels, while DashAttention uses Triton kernels.

A unified benchmark should specify hardware (GPU type, memory bandwidth), sequence length distribution (average and maximum), task suite (language modeling, retrieval, classification, generation), and evaluation metrics (wall-clock time, FLOPs, memory usage, perplexity, downstream accuracy). The benchmark should include both forward-pass-only measurements for training and incremental decoding measurements for inference. Quality-side suites such as LongBench (Bai et al., 2024), RULER (Hsieh et al., 2024), and HELMET (Yen et al., 2025) provide long-context evaluation infrastructure but do not standardize the efficiency axis; unifying quality and efficiency measurement in a single protocol is the missing piece. Establishing such a benchmark would accelerate progress by enabling clear winners to emerge for specific operating regimes.

\subsection{Learned Routing Policies for Hybrid Architectures}\label{learned-routing-policies-for-hybrid-architectures}

The SSD framework reveals a spectrum between full attention and pure SSMs, but the optimal routing between these mechanisms remains an open problem. Should routing be static (e.g., attention in early layers, SSM in later layers) or dynamic (e.g., attention for retrieval-heavy tokens, SSM for compression-heavy tokens)? Should the routing mechanism itself be differentiable (as in DashAttention) or discrete (as in Reformer)?

A parallel development in the Mixture-of-Experts (MoE) literature (Shazeer et al., 2017; Fedus et al., 2022) demonstrates a closely related routing problem at the feed-forward level: top-k gating routes each token to a subset of expert networks via a learned softmax over expert representations. The routing mechanisms in sparse attention (LSH hashing (Reformer), k-means clustering (Routing Transformer), and $\alpha$-entmax selection (DashAttention)) are structurally analogous to MoE gating, suggesting that insights from the MoE literature on load balancing, auxiliary losses, and expert capacity might transfer to attention routing. Conversely, attention's fully-differentiable routing (DashAttention) could inform softer MoE gating mechanisms, representing an underexplored cross-pollination opportunity.

Early evidence suggests that different layers learn different functions, and routing should respect this functional specialization. For example, early layers may benefit from attention for local feature interactions, middle layers from hybrid mechanisms for constructing representations, and late layers from SSMs for global summarization. Learned routing policies that adapt the attention-SSM allocation to the input and the layer could achieve better quality-efficiency trade-offs than either homogeneous architecture.

\subsection{Attention and Mixture-of-Experts Routing}\label{attention-and-mixture-of-experts-routing}

The intersection of attention and MoE routing is one of the most active architectural frontiers in 2024--2026, yet it is often treated separately from the efficiency-focused attention literature. MoE architectures (Shazeer et al., 2017; Fedus et al., 2022) increase model capacity without proportionally increasing compute by activating only a subset of feed-forward network parameters per token. The gating network, a learned softmax over expert representations, is structurally analogous to the content-based routing mechanisms used in sparse attention (Reformer's LSH, Routing Transformer's k-means, DashAttention's $\alpha$-entmax). Both face the same challenges: load balancing across routes, minimizing routing overhead, and maintaining end-to-end differentiability.

Several production architectures combine MoE with attention innovations. Mixtral 8$\times$7B (Jiang et al., 2024) uses a standard Transformer with GQA attention (Section 3.2) and top-2 MoE feed-forward layers, demonstrating that MoE can be combined with KV cache-efficient attention without interference. DeepSeek-V2 (DeepSeek-AI, 2024) and DeepSeek-V3 (DeepSeek-AI, 2025) extend this by introducing Multi-Head Latent Attention (MLA), which compresses the KV cache into a low-rank latent space, a form of learned KV compression that shares conceptual similarities with the grouped-query approach. MLA is a notable production-scale KV-compression technique classified here under Dense (KV-efficient) alongside GQA and MQA, because its mechanism is a low-rank parameterization of the standard attention computation rather than an implementation strategy; as with GQA in the trend analysis (Table~\ref{tab:trends}), it is nonetheless discussed together with IO-aware methods because both target the same serving-memory pressure, but it does not receive a formal EEI score in this survey because published per-benchmark long-range retrieval, throughput, and interpretability data under comparable conditions were not available at the time of scoring. To assign MLA a score under our rubric would require: (i) Long Range Arena averages or long-context retrieval accuracy on the same protocol used in Table~\ref{tab:benchmarks}, (ii) end-to-end throughput at equivalent batch sizes and sequence lengths on a comparable GPU generation, and (iii) an assessment of whether MLA's low-rank latent representation admits mechanistic interpretability tools comparable to those available for standard attention heads. Its inclusion as a qualitative reference in Section~7.3, without a score, reflects the survey's evidentiary threshold: methods that are cited architecturally but lack the multi-benchmark profile required for rubric-based scoring are discussed contextually rather than ranked. Hybrid designs that interleave SSM and attention layers were pioneered by H3 (Fu et al., 2023), which stacks diagonal state-space layers with attention blocks; Jamba extends this pattern by interleaving Mamba layers with attention. Jamba (Lieber et al., 2024) and Jamba-1.5 (Jamba Team, 2024) directly combine MoE with Mamba (SSM) layers, interleaving Transformer-attention and Mamba layers, with MoE replacing selected MLP sublayers, in a hybrid architecture that achieves efficient long-context processing.

The design space for attention-MoE hybrids spans four dimensions: the choice of token-mixing mechanism (attention, SSM, or hybrid), the routing strategy for expert selection (top-k, learned, or differentiable), the layer-level allocation of attention versus MoE compute, and the training regime (pretraining from scratch versus fine-tuning an existing dense model). This design space is largely unexplored beyond a few production configurations, and the EEI framework provides a natural lens for comparing hybrid architectures along the efficiency, expressiveness, and interpretability axes simultaneously.

\subsection{Length Generalization in Sub-Quadratic Models}\label{length-generalization-in-sub-quadratic-models}

Length generalization is a long-standing problem: Transformer-XL (Dai et al., 2019) introduced segment-level recurrence with relative positional encoding precisely to extend beyond fixed-length windows. Sub-quadratic attention methods (sparse attention, linear attention, SSMs) often fail to generalize to sequence lengths not seen during training. This is particularly problematic because the efficiency gains of these methods are most valuable at long sequences, but training at the target length is often prohibitively expensive. A model trained with 8K-token attention windows may degrade catastrophically when evaluated at 128K tokens, failing to retrieve information from beyond the training window.

The causes of length generalization failure differ across methods. For linear attention, approximation error can become more consequential at longer sequence lengths under fixed feature budgets and demanding retrieval tasks. For sparse attention, fixed patterns may not capture interactions at unseen distances. For SSMs, the recurrent state dynamics may not extrapolate to longer dependencies. Theoretical understanding of these failure modes is limited, and principled solutions (such as length-extrapolable positional encodings, length-adaptive routing, or curriculum learning over sequence lengths) remain open research directions.

\subsection{Scaling Mechanistic Interpretability}\label{scaling-mechanistic-interpretability}

Much of the detailed mechanistic-interpretability literature has focused on relatively small and mid-sized models and narrow tasks; scaling these methods reliably to frontier-scale models remains challenging. Scaling these methods to 70B+ models presents three challenges. First, as model depth, width, and the number of interacting components increase, the space of possible cross-layer composition patterns becomes substantially more difficult to analyze. Manual analysis of individual heads, as demonstrated for induction heads in small models, becomes infeasible at scale.

Second, superposition is hypothesized to be more severe in larger models, making it harder to isolate the features driving attention behavior. The number of features stored in superposition may grow with model capacity, which would increase the difficulty of sparse autoencoder training and feature interpretation.

Third, the relationship between attention behavior and model outputs may become more diffuse in larger models. A single output token depends on many attention heads across many layers through complex compositional circuits, making it difficult to attribute behavior to specific heads. Automated circuit discovery methods, such as activation patching (Wang et al., 2022) and attribution patching (Syed et al., 2023), have shown promise for scaling mechanistic interpretability, but their reliability and completeness at frontier scale remain open questions.

\subsection{Explicit Research Gap Analysis}\label{explicit-research-gap-analysis}

The preceding subsections identify challenges facing the field. Here we distill these into five specific, unsolved research gaps that define the frontier of attention research.

\textbf{Gap 1: Retrieval-Efficiency Pareto Frontier.} We identified no method in the surveyed literature that simultaneously achieves (i) exact retrieval accuracy across arbitrary context positions, (ii) sub-quadratic complexity, and (iii) hardware-efficient implementation. FlashAttention achieves (i) and (iii) but remains $O(L^2)$. Mamba achieves (ii) and (iii), though the cited evaluations provide only limited direct evidence of its retrieval quality on tasks requiring precise token retrieval (see Section~5). Adaptive sparsity methods approach (i) at high sparsity ratios but introduce routing overhead and are not yet production-ready. Closing this gap likely requires hybrid architectures that allocate exact attention to retrieval-critical tokens while applying sub-quadratic compression to the remainder.

\textbf{Gap 2: Scaling Sparse Attention to 1M+ Tokens.} Within our surveyed corpus, most sparse-attention evaluations used sequence lengths of up to 32K tokens. Extrapolating to 1M tokens introduces new failure modes: the routing mechanism's overhead grows with the number of groups or clusters, the sparsity pattern may need to adapt multiple times within a single sequence, and the memory footprint of auxiliary data structures (hash tables, cluster centroids, routing logits) becomes non-negligible. Within the surveyed literature, we identified no published work that demonstrates reliable attention at 1M+ tokens with both quality comparable to full attention and wall-clock speedup exceeding FlashAttention; the closest result, Focus reporting an $8.6\times$ speedup at 1M tokens, measures latency only, with quality not evaluated at that length (Yao et al., 2026).

\textbf{Gap 3: Theoretical Understanding of When Sparsity Succeeds.} BigBird's Turing-completeness proof guarantees that some sparse pattern can approximate any computation, but it provides no guidance for which pattern to choose for a given task. The field lacks a theoretical framework that predicts, for a given task and data distribution, which tokens can be safely pruned without quality loss. Such a framework would bridge the gap between the empirical success of learned routing (Focus, DashAttention) and the theoretical guarantees of structured sparse patterns (BigBird).

\textbf{Gap 4: Attention-SSM Routing without Task-Specific Tuning.} Hybrid architectures that interleave attention and SSM layers (e.g., Jamba (Lieber et al., 2024), Samba (Ren et al., 2024)) achieve strong results; however, the optimal layer arrangement remains a design choice, and task-adaptive dynamic architectures are open research. Learned routing policies that dynamically allocate computation at the token level, using attention for retrieval-critical tokens and SSM for compression, remain in early research stages. The connection to Mixture-of-Experts routing (Shazeer et al., 2017) is promising but underexplored.

\textbf{Gap 5: Interpretability-Guided Efficiency.} Current efficiency methods mostly reduce compute without routing on mechanistic head function; adaptive methods such as DashAttention already vary sparsity across tokens, heads, and layers, but none routes based on mechanistically identified head specialization. Mechanistic interpretability research shows that heads specialize (Clark et al., 2019; Olsson et al., 2022; Elhage et al., 2022): some heads perform precise retrieval (which may require exact attention), while others perform fuzzy pattern matching (which may tolerate approximation). A framework that uses interpretability to identify which heads can use approximate attention and which require exact computation could unlock a new efficiency dimension. This direction remains almost entirely unexplored.

\subsection{Reproducibility and Deployment Considerations}\label{reproducibility-and-deployment-considerations}

Reproducibility in efficient attention research faces structural challenges. Most papers report speedups on specific GPU configurations that are difficult to replicate: FlashAttention results depend on exact GPU model (A100-40GB vs A100-80GB), CUDA version, and PyTorch compilation settings. Mamba's results require a custom GPU scan kernel (CUDA with ROCm/HIP support for AMD GPUs) that is not available in standard deep learning frameworks. The absence of standardized benchmarking infrastructure means that claimed speedups often fail to reproduce across labs.

Open-source availability varies significantly across methods; as of mid-2026, the cited papers and their linked repositories document the following status. FlashAttention is integrated into PyTorch's scaled dot-product attention implementation (torch.nn.functional.scaled\_dot\_product\_attention; PyTorch Team, 2023) and widely deployed. PagedAttention is available through the vLLM serving framework (Kwon et al., 2023). Mamba's reference implementation is open-source but requires custom GPU kernels. Some linear-attention primitives are available through xFormers and related libraries. DashAttention has a released reference implementation; Focus code was not located as of August 7, 2026.

Deployment constraints also differ. FlashAttention-3 is optimized for Hopper-class H100 GPUs, with its FP8 path requiring Hopper-class FP8 tensor-core support. Its FP16/BF16 kernels do not require the FP8 path and therefore do not impose the same hardware restriction. PagedAttention requires careful tuning of block size and preemption thresholds. Mamba's constant-state inference is attractive for on-device deployment but requires the custom scan kernel to be compiled for target hardware. Many production LLM stacks pair exact-attention kernels (e.g., FlashAttention, integrated in PyTorch's scaled dot-product attention implementation (PyTorch Team, 2023) and vLLM (Kwon et al., 2023)) with KV-cache-compressing head-sharing variants such as GQA (Touvron et al., 2023; Dubey et al., 2024; Jiang et al., 2023), which are the most widely deployed combinations as of mid-2026; sub-quadratic methods are typically reserved for specialized use cases where the quality-efficiency trade-off is favorable. These deployment statements reflect publicly documented framework integrations as of mid-2026.

We recommend that future work adhere to the following reproducibility standards: (i) report speedups on at least two GPU generations (e.g., A100 and H100); (ii) measure end-to-end model throughput in addition to isolated attention kernel benchmarks; (iii) release reference implementations with permissive licenses; (iv) evaluate on standardized benchmarks (LRA, RULER (Hsieh et al., 2024), HELMET (Yen et al., 2025)) with published configurations; and (v) report quality metrics at multiple sequence lengths to characterize length-dependent degradation.

\subsection{Threats to Validity}\label{threats-to-validity}

This survey synthesizes a rapidly evolving field, and its conclusions are subject to several threats to validity that warrant explicit acknowledgment.

\textit{Publication bias.} The methods surveyed are predominantly those published in top-tier venues (NeurIPS, ICML, ICLR, ACL, EMNLP) or widely cited arXiv preprints. Negative results, methods that failed to achieve competitive quality or efficiency, are underrepresented in the literature and therefore in this survey. The EEI scores may thus overestimate the average quality of attention research, as unpublished failures are excluded by construction.

\textit{Preprint-heavy frontier.} A significant fraction of the newest efficiency innovations surveyed (Focus, DashAttention, DeepSeek-V2/V3) are available only as arXiv preprints at the time of writing. These works have not undergone peer review, and their reported benchmark numbers may change during the review process or fail to replicate in independent evaluations. We have noted this where relevant (e.g., the cross-study caveat before Table~\ref{tab:benchmarks}), but the reliance on non-peer-reviewed sources is an inherent limitation of surveying a field that moves faster than the publication cycle. Some earlier works cited as preprints (e.g., FlashAttention-3 at NeurIPS 2024, RULER at COLM 2024, HELMET at ICLR 2025, Ring Attention at ICLR 2024) have since been archived; the bibliography reflects their current status.

\textit{Cross-paper benchmark inconsistency.} As noted in Section~5.7, the benchmark numbers in Tables~\ref{tab:benchmarks} and~\ref{tab:speedups} are compiled from different papers using different hardware, different sequence lengths, different batch sizes, and different evaluation protocol; quality-relative terms indicate that the source paper explicitly reports comparability to its exact-attention baseline; no universal numerical tolerance is imposed across tasks. The Needle-in-a-Haystack test, while standardized in concept, is implemented differently across papers: some use single-needle retrieval, others multi-needle; some report accuracy at fixed depth, others aggregate across depths. The throughput numbers are particularly sensitive to measurement methodology (isolated attention kernel vs.\ end-to-end model inference). We have mitigated this threat by grouping methods into categories for quantitative cross-study synthesis (Table~\ref{tab:meta-analysis}) rather than claiming fine-grained rankings, but the fundamental incomparability of cross-study benchmarks limits the precision of any quantitative synthesis.

\textit{Subjectivity of EEI scoring.} Despite the rubric in Table~\ref{tab:eei-rubric} and the detailed score justifications in Appendix~\ref{app:eei-justification}, the EEI scores remain ordinal judgments that reflect the author's interpretation of published results. Systematic biases (such as over-weighting benchmark results from well-known papers) cannot be ruled out. A formal multi-rater validation study would strengthen confidence in the scores; we leave this to future work.

\textit{Rapidly changing frontier.} The attention efficiency field evolves on a timescale of months, not years. Since the initial drafting of this survey, Mamba-3 has been published at ICLR 2026 and at least three new adaptive sparsity methods have appeared on arXiv. FlashAttention-4 has extended the IO-aware line to Blackwell-class hardware (Zadouri et al., 2026); identified within the literature cutoff, it is excluded from the scored EEI panel and the quantitative synthesis because it does not yet provide the comparable multi-axis evidence profile required for rubric scoring, particularly common-protocol efficiency and interpretability evidence. More generally, several additional 2026 methods were identified but are not scored because they lack the same multi-axis evidence profile (long-range retrieval, throughput, and interpretability evidence under comparable conditions) at the 31 May 2026 cutoff; their exclusion is a rubric-consistency decision, not an assessment of their quality. The EEI scores and benchmark tables capture the state of the field at a specific point in time (mid-2026) and will require periodic updating to remain current. We have designed the EEI framework to facilitate such updates: adding a new method requires only assigning an $(E, \mathit{Ex}, I)$ score via the rubric and adding a row to the existing tables.

\textit{Limited hardware diversity in efficiency evaluation.} The throughput speedups in Table~\ref{tab:speedups} are dominated by NVIDIA GPU measurements (A100, H100). Methods that target alternative hardware (such as Apple Neural Engine, Google TPU, AMD MI300, or edge devices) may exhibit different relative performance characteristics. FlashAttention's measured speedups are dominated by NVIDIA GPU evaluations, particularly A100/H100 results; relative performance on AMD, TPU, Apple, and other accelerators may differ substantially, and the EEI Efficiency scores should not be generalized across hardware without independent validation. Similarly, Mamba's reference implementation relies on custom GPU kernels; the benchmark results surveyed here are predominantly NVIDIA/CUDA-based, although AMD/ROCm support is available. The EEI Efficiency scores are implicitly NVIDIA-centric, and we caution against generalizing the rankings to non-NVIDIA hardware without independent validation.

\subsection{Outlook Toward 2030}\label{outlook-toward-2030}

The trajectory traced by this survey suggests several plausible directions for attention research over the next five years.

\textit{Pure quadratic attention may become less attractive for 512K--1M token contexts, though this is a forecast rather than an established finding.} The $O(L^2)$ cost of standard attention, even with FlashAttention's hardware optimizations, becomes prohibitive beyond 128K tokens. Systems operating at 512K--1M token contexts will rely on hybrid architectures or sub-quadratic mechanisms for the majority of their computation, reserving exact attention for retrieval-critical segments. An open question is which hybrid configuration, if any, becomes standard.

\textit{Attention-SSM hybrid architectures will expand.} Jamba (Lieber et al., 2024), its successor Jamba-1.5 (Jamba Team, 2024), Samba (Ren et al., 2024), and related hybrids have demonstrated that interleaving attention and SSM layers combines the strengths of both paradigms. We expect this trend to accelerate, with learned routing policies (Section~7.2) replacing fixed layer assignments. The SSD framework's theoretical unification suggests that the boundary between attention and SSMs will continue to blur, potentially leading to a single parameterized mechanism that spans both regimes.

\textit{Adaptive routing will replace fixed sparsity.} The fixed sparse patterns that dominated the 2020--2022 era (Longformer, BigBird, Sparse Transformer) are increasingly being complemented by content-dependent learned routing (Focus, DashAttention). One plausible trajectory is increasing replacement of fixed sparsity by content-dependent routing for long-context attention by 2030, if the reported quality margins reproduce under independent evaluation: the quality advantage of learned routing at comparable efficiency levels is reported in the two preprints' own benchmark suites but remains unverified on common benchmarks. This prediction hinges on the unverified Focus and DashAttention benchmark results; if independent reproduction finds smaller quality margins, for instance if the adaptive methods degrade to near-fixed-sparsity quality under rigorous evaluation, then the trajectory may be toward hybrid fixed-adaptive schemes rather than pure learned routing.

\textit{Interpretability benchmarks will become standard.} Existing benchmarks such as MIB (Mueller et al., 2025) provide important standardized evaluations of mechanistic-interpretability methods across two tracks, four tasks, and five models, but no widely adopted benchmark directly compares interpretability across heterogeneous attention and sequence-mixing architectures; this limits systematic cross-method comparison as models approach 1T parameters. We predict that by 2028, at least one quantitative interpretability benchmark (analogous to LRA for efficiency or Needle-in-a-Haystack for expressiveness) will be widely adopted, enabling the kind of cross-method comparison that has driven progress in efficiency.

\textit{Hardware diversity will reshape efficiency rankings.} The current dominance of NVIDIA-specific optimizations (FlashAttention's tensor core usage, Mamba's custom GPU scan kernels) means that rankings are tied to specific hardware. The emergence of competing accelerator platforms and custom AI accelerators will create a more diverse hardware market where methods are evaluated across multiple targets. The methods best positioned for this future are those with portable implementations: IO-aware exact attention (portable with Triton), SSMs (portable with custom kernels targeting multiple backends), and adaptive sparsity (portable with standard PyTorch operations).

\textit{The EEI frontier will continue to expand.} The pattern observed in this survey, alternating phases of efficiency improvement, expressiveness recovery, and hybrid integration, suggests that the frontier will continue its outward expansion. A plausible research direction is toward methods that combine learned routing with hardware-aware kernels to approach higher efficiency and expressiveness at long contexts; this is a hypothesis, not a conclusion supported by the present cross-study evidence.

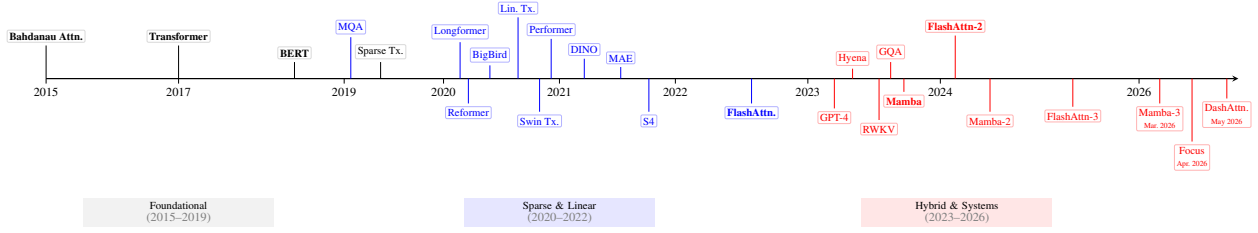
\begin{figure*}[!tbp]
\centering
\resizebox{\textwidth}{!}{\begin{tikzpicture}[x=0.38cm, y=0.38cm]
\scriptsize
\draw[thick, -stealth] (0,0) -- (72,0);
\foreach \x/\year in {0/2015, 8/2017, 18/2019, 24/2020, 31/2021, 38/2022, 46/2023, 54/2024, 66/2026}
  \draw (\x,0) -- (\x,-0.3) node[below,font=\footnotesize] {\year};
\draw[blue] (30.5,0) -- (30.5,2.4) node[above=2pt, fill=white, draw=blue!30, rounded corners=1pt, line width=0.2pt, inner sep=1.5pt, font=\scriptsize, text=blue, anchor=south] {Performer};
\draw[red] (54.9,0) -- (54.9,2.6) node[above=2pt, fill=white, draw=red!30, rounded corners=1pt, line width=0.2pt, inner sep=1.5pt, font=\scriptsize\bfseries, text=red, anchor=south] {FlashAttn-2};
\draw[blue] (18.4,0) -- (18.4,2.5) node[above=2pt, fill=white, draw=blue!30, rounded corners=1pt, line width=0.2pt, inner sep=1.5pt, font=\scriptsize, text=blue, anchor=south] {MQA};
\draw[blue] (28.5,0) -- (28.5,3.7) node[above=2pt, fill=white, draw=blue!30, rounded corners=1pt, line width=0.2pt, inner sep=1.5pt, font=\scriptsize, text=blue, anchor=south] {Lin.\ Tx.};
\draw (15,0) -- (15,1.0) node[above=2pt, fill=white, draw=gray!30, rounded corners=1pt, line width=0.2pt, inner sep=1.5pt, font=\scriptsize\bfseries, anchor=south] {BERT};
\draw (20.2,0) -- (20.2,1.0) node[above=2pt, fill=white, draw=gray!30, rounded corners=1pt, line width=0.2pt, inner sep=1.5pt, font=\scriptsize, anchor=south] {Sparse Tx.};
\draw[blue] (25.0,0) -- (25.0,2.2) node[above=2pt, fill=white, draw=blue!30, rounded corners=1pt, line width=0.2pt, inner sep=1.5pt, font=\scriptsize, text=blue, anchor=south] {Longformer};
\draw[blue] (26.8,0) -- (26.8,0.8) node[above=2pt, fill=white, draw=blue!30, rounded corners=1pt, line width=0.2pt, inner sep=1.5pt, font=\scriptsize, text=blue, anchor=south] {BigBird};
\draw[blue] (32.5,0) -- (32.5,1.2) node[above=2pt, fill=white, draw=blue!30, rounded corners=1pt, line width=0.2pt, inner sep=1.5pt, font=\scriptsize, text=blue, anchor=south] {DINO};
\draw[blue] (34.7,0) -- (34.7,0.7) node[above=2pt, fill=white, draw=blue!30, rounded corners=1pt, line width=0.2pt, inner sep=1.5pt, font=\scriptsize, text=blue, anchor=south] {MAE};
\draw[blue] (25.5,0) -- (25.5,-1.5) node[below=2pt, fill=white, draw=blue!30, rounded corners=1pt, line width=0.2pt, inner sep=1.5pt, font=\scriptsize, text=blue, anchor=north] {Reformer};
\draw[blue] (29.8,0) -- (29.8,-2.0) node[below=2pt, fill=white, draw=blue!30, rounded corners=1pt, line width=0.2pt, inner sep=1.5pt, font=\scriptsize, text=blue, anchor=north] {Swin Tx.};
\draw[blue] (42.6,0) -- (42.6,-1.5) node[below=2pt, fill=white, draw=blue!30, rounded corners=1pt, line width=0.2pt, inner sep=1.5pt, font=\scriptsize\bfseries, text=blue, anchor=north] {FlashAttn.};
\draw[blue] (36.4,0) -- (36.4,-2.0) node[below=2pt, fill=white, draw=blue!30, rounded corners=1pt, line width=0.2pt, inner sep=1.5pt, font=\scriptsize, text=blue, anchor=north] {S4};
\draw[red] (47.6,0) -- (47.6,-1.8) node[below=2pt, fill=white, draw=red!30, rounded corners=1pt, line width=0.2pt, inner sep=1.5pt, font=\scriptsize, text=red, anchor=north] {GPT-4};
\draw[red] (51.8,0) -- (51.8,-0.8) node[below=2pt, fill=white, draw=red!30, rounded corners=1pt, line width=0.2pt, inner sep=1.5pt, font=\scriptsize\bfseries, text=red, anchor=north] {Mamba};
\draw[red] (51.0,0) -- (51.0,1.0) node[above=2pt, fill=white, draw=red!30, rounded corners=1pt, line width=0.2pt, inner sep=1.5pt, font=\scriptsize, text=red, anchor=south] {GQA};
\draw[red] (48.7,0) -- (48.7,0.6) node[above=2pt, fill=white, draw=red!30, rounded corners=1pt, line width=0.2pt, inner sep=1.5pt, font=\scriptsize, text=red, anchor=south] {Hyena};
\draw[red] (50.3,0) -- (50.3,-2.5) node[below=2pt, fill=white, draw=red!30, rounded corners=1pt, line width=0.2pt, inner sep=1.5pt, font=\scriptsize, text=red, anchor=north] {RWKV};
\draw[red] (62.0,0) -- (62.0,-1.7) node[below=2pt, fill=white, draw=red!30, rounded corners=1pt, line width=0.2pt, inner sep=1.5pt, font=\scriptsize, text=red, anchor=north] {FlashAttn-3};
\draw[red] (57.0,0) -- (57.0,-2.0) node[below=2pt, fill=white, draw=red!30, rounded corners=1pt, line width=0.2pt, inner sep=1.5pt, font=\scriptsize, text=red, anchor=north] {Mamba-2};
\draw[red] (67.25,0) -- (67.25,-1.5) node[below=2pt, fill=white, draw=red!30, rounded corners=1pt, line width=0.2pt, inner sep=1.5pt, font=\scriptsize, text=red, anchor=north, align=center] {Mamba-3\\{\tiny Mar.\ 2026}};
\draw[red] (69.2,0) -- (69.2,-3.8) node[below=2pt, fill=white, draw=red!30, rounded corners=1pt, line width=0.2pt, inner sep=1.5pt, font=\scriptsize, text=red, anchor=north, align=center] {Focus\\{\tiny Apr.\ 2026}};
\draw[red] (71.3,0) -- (71.3,-1.2) node[below=2pt, fill=white, draw=red!30, rounded corners=1pt, line width=0.2pt, inner sep=1.5pt, font=\scriptsize, text=red, anchor=north, align=center] {DashAttn.\\{\tiny May 2026}};
\draw (0,0) -- (0,2.0) node[above=2pt, fill=white, draw=gray!30, rounded corners=1pt, line width=0.2pt, inner sep=1.5pt, font=\scriptsize\bfseries, anchor=south, align=center] {Bahdanau Attn.};
\draw (8,0) -- (8,2.0) node[above=2pt, fill=white, draw=gray!30, rounded corners=1pt, line width=0.2pt, inner sep=1.5pt, font=\scriptsize\bfseries, anchor=south, align=center] {Transformer};
\node[fill=gray!10, font=\scriptsize, minimum height=0.5cm, text width=4.2cm, align=center] at (8,-8.1) {Foundational\\{\footnotesize\color{gray}(2015--2019)}};
\node[fill=blue!10, font=\scriptsize, minimum height=0.5cm, text width=4.2cm, align=center] at (31,-8.1) {Sparse \& Linear\\{\footnotesize\color{gray}(2020--2022)}};
\node[fill=red!10,  font=\scriptsize, minimum height=0.5cm, text width=4.2cm, align=center] at (55,-8.1) {Hybrid \& Systems\\{\footnotesize\color{gray}(2023--2026)}};
\end{tikzpicture}}
\caption{Evolution of selected attention and sequence-modeling systems (2015--2026). Foundational period established the core architecture; the sparse/linear era introduced sub-quadratic alternatives; the current hybrid era blends systems co-design, SSMs, and learned sparsity. The timeline is a selected chronology of representative methods and system landmarks (including model-scale entries such as GPT-4), not an exhaustive listing of every surveyed method. Eras are thematic sub-eras, not chronologically exclusive. Node years denote the cited publication/release year; venue years may differ for methods whose initial public release preceded formal publication (e.g., Bahdanau et al. (2015) was posted to arXiv in 2014; BERT is cited to Devlin et al. (2019) but was first released in October 2018; Mamba and FlashAttention-2 are commonly cited by their 2024 venue years but were first released in December 2023 and July 2023, respectively). Horizontal offsets within a year are display-only and do not encode publication date. Year positions on the axis are schematic rather than proportional to elapsed time; years in which no surveyed method debuted are omitted.}
\label{fig:timeline}
\end{figure*}

Figure~\ref{fig:timeline} summarizes the chronology discussed above, locating the surveyed methods along the 2015--2026 timeline and grouping them into the three thematic eras described in Section~7.1.

The timeline reveals three distinct eras. The foundational era (2015--2019) established the core attention architecture and its dominant variants. The sparse and linear era (2020--2022) saw an explosion of methods targeting the $O(L^2)$ bottleneck through fixed patterns, content-dependent routing, and kernel approximations. The current hybrid and systems era (2023--2026) is characterized by three converging trends: hardware-aware co-design (FlashAttention), state-space alternatives with duality guarantees (Mamba, SSD), and learned differentiable sparsity (Focus, DashAttention). The figure makes clear that the rate of innovation has accelerated, with the majority of currently relevant methods emerging in the last three years.

\subsection*{Design Principles for Future Attention Architectures}

We distill the findings of this survey into five actionable design principles for researchers and practitioners developing next-generation attention mechanisms.

\textit{Principle 1: Preserve exact computation; select what to compute, do not approximate it.} The EEI evidence, rubric assignments rather than independent measurements (Section 2), is clear: methods that compute exact attention within a selected subset (Focus: Ex=9 at E=7; DashAttention: Ex=9 at E=8) report higher Ex than methods that approximate the attention kernel (Performer: Ex=6 at E=7; Reformer: Ex=7 at E=6; Reformer's approximation is in the LSH bucket assignment, with attention exact within buckets) at comparable or worse efficiency. The mechanism is architectural: the softmax distribution that gives attention its expressiveness, sharp input-dependent allocation of probability mass across tokens, can be altered by low-rank projections, random feature maps, or locality-sensitive hashing, potentially introducing errors in long-range or fine-grained retrieval. Routing tokens into groups or blocks and computing exact attention within them is a strong candidate design principle, pending replication of the adaptive results, rather than approximating the attention scores themselves.

\textit{Principle 2: Distinguish algorithmic complexity from hardware efficiency.} The $O(L^2)$ FlashAttention family achieves competitive practical throughput compared with $O(L)$ alternatives at many sequence lengths up to $\sim$128K (Table~\ref{tab:speedups}); the crossover is workload- and hardware-dependent, because for many attention workloads on modern GPUs, memory movement (HBM traffic) is a dominant performance constraint; the boundary is length- and workload-dependent (e.g., Mamba-2 reports a $6\times$ speedup over FlashAttention-2 at 16K under the specific configuration reported by Dao and Gu (2024) on its own benchmark). Algorithmic approximations that reduce FLOPs but do not address the memory wall will underperform IO-aware implementations that keep data in SRAM. For any attention method targeting GPU execution at sub-128K contexts, HBM traffic should be treated as a primary design consideration alongside asymptotic complexity. This reframes the conventional emphasis on asymptotic complexity that has guided efficient attention design since 2019: the memory hierarchy can dominate wall-clock speed at practical sequence lengths.

\textit{Principle 3: Favor adaptive selection over fixed sparsity.} Generalization~5 suggests that among methods occupying similar positions on the EEI Efficiency axis (E = 7--8), adaptive sparse methods may achieve substantially higher retrieval accuracy than fixed-pattern methods, though this rests on preprints awaiting independent verification. If replicated, the gap is causal: fixed patterns (windows, strides, random connections) cannot distinguish important from unimportant tokens because they do not consider token content, while adaptive methods (clustering, entmax routing, learned grouping) allocate the attention budget to tokens that matter for the current input. Future efficiency research should focus on improving the cost and scalability of learned routing rather than designing better fixed patterns.

\textit{Principle 4: Design interpretability in from the start.} The Interpretability (I) axis is the least developed EEI dimension (median I = 6 across the 21 surveyed methods, versus median E = 7 and median Ex = 8), and this gap is structural: none of the 21 methods in our scored panel was designed with interpretability as a primary objective. The highest-scoring methods (Swin: I=8; Longformer, BigBird: I=7) achieve their scores not through explicit design for interpretability but because their architectural properties (spatial locality for Swin, explicit decomposition of attention types for BigBird) create analyzable structure as a side effect. Future architectures should treat interpretability as a design constraint: include analyzable attention distributions, provide theoretical guarantees about head or state function, and support sparse autoencoder training without architectural modifications.

\textit{Principle 5: Hybridize rather than polarize.} No surveyed method combines the strongest observed levels across all three EEI axes (Table~\ref{tab:eei-scores}). Under equal weights, five methods share the top composite EEI score of 8.0: FlashAttention-2/3, PagedAttention, DashAttention, Swin Transformer, and Mamba-2; deployment weights single out Mamba-2 (8.8); retrieval weights lead exact IO-aware attention (FlashAttention-2/3 and PagedAttention at 8.8). Different architectural paradigms top the ranking under different weightings, and hybrid architectures that combine their complementary strengths (exact attention for retrieval, efficient state-space layers for context bandwidth, and hardware-aware kernels for throughput) are the natural extension of the EEI frontier's shape. The trend toward hybrid architectures (Jamba (Lieber et al., 2024), Samba (Ren et al., 2024), Mamba-3 (Lahoti et al., 2026)) is not incidental; it is the logical consequence of the EEI frontier's shape. Within the surveyed panel, single-paradigm methods occupy the current observed limits of the EEI map: pure attention tops out at Ex=10/E=4 or Ex=10/E=8 with IO-awareness; within the surveyed 21-method panel, the highest-scoring configuration among selective-scan state-space methods (Mamba, treated here as the pure SSM class, with Hyena classified as a long-convolution method and Mamba-2 as a Structured SSM/SSD method whose formulation has a structured-linear-attention dual) is E=10/Ex=7. The next generation of architectures will compose these primitives through learned routing policies that dynamically allocate tokens to the appropriate mechanism based on input content, sequence position, and hardware state.

These five principles, grounded in the quantitative analysis of Sections~5.4--5.5, provide actionable guidance for future research. They imply that the most promising direction for attention architecture design is a convergence of exact computation, learned routing, hardware-aware implementation, and interpretability-aware design into a single unified framework.

\section*{Conclusion}

The attention mechanism has evolved from a simple alignment model for machine translation to the dominant computational primitive in deep learning. This survey has followed its evolution across four threads that are often treated separately but are in fact tightly connected, organized through the Efficiency-Expressiveness-Interpretability (EEI) framework introduced in Section~1.

Viewed through the EEI lens, the history of attention research can be understood as a directed outward push along the EEI frontier, a continuous expansion of the Pareto-optimal boundary that no single method has yet fully dominated. Standard attention (2017) anchored the upper-left corner: maximal expressiveness, minimal efficiency. Sparse and linear methods (2020--2022) pushed the frontier rightward, gaining efficiency at expressiveness cost as scored by the rubric. IO-aware exact attention (2022--2024) sidestepped the trade-off by demonstrating that $O(L^2)$ computation could be hardware-efficient, achieving a significant efficiency gain without expressiveness loss. State-space models (2023 onward) introduced an entirely new Pareto layer with extreme efficiency but structural interpretability gaps. The most recent adaptive sparsity methods (2025--2026) are converging toward the upper-right quadrant within the surveyed panel, suggesting that, if the reported adaptive results reproduce, the trade-off may be partially avoidable for retrieval-focused regimes. Within the surveyed panel, this trajectory, from unconstrained expressiveness to budget-aware computation to learned adaptive routing, can be described as the defining narrative of a decade of attention research.

The foundational formulation, scaled dot-product attention with multiple heads, remains the most widely used despite a decade of proposed alternatives. Positional encodings have evolved from sinusoids to RoPE, and architectural choices (encoder-only, decoder-only, encoder-decoder) define the operational envelope of each attention variant. In computer vision, attention has been adapted through patch embeddings, object queries, hierarchical windows, and cross-modal alignment, each requiring specific modifications to the core mechanism.

Efficiency innovations have advanced along parallel tracks. Fixed sparse patterns and content-dependent routing reduce the number of computed interactions but introduce approximation or routing overhead. Linear attention provides $O(L)$ complexity through kernel tricks but suffers from approximation error and high constant factors. IO-aware exact attention, exemplified by the FlashAttention family, achieves exact computation at hardware-determined speeds by co-designing the algorithm with the memory hierarchy. State-space models offer a fundamentally different paradigm with linear-time recurrence, with the cited evidence leaving some long-range retrieval capabilities less directly characterized than those of full attention. The most recent work on adaptive sparsity (Focus (Yao et al., 2026), DashAttention (Huang et al., 2026)) converges these approaches, learning content-dependent routing with differentiable mechanisms that can be optimized end-to-end.

On the interpretability side, attention heads have been shown to implement interpretable algorithms ranging from simple syntactic dependencies to the induction heads that enable in-context learning. The superposition hypothesis provides a theoretical framework for understanding how models encode more features than dimensions, and sparse autoencoders offer a practical tool for disentangling these representations. The SSD framework deepens this theoretical understanding by establishing a mathematical connection between certain structured state-space models and attention-like matrix operations, suggesting a continuum defined by the structure of the token mixing matrix.

Despite these advances, fundamental challenges remain, as enumerated in our research gap analysis (Section~7.6): the retrieval-efficiency Pareto frontier is unresolved, sparse attention has not yet been independently reproduced at 1M+ tokens under a common quality-and-efficiency evaluation protocol, theoretical guidance for sparsity pattern selection is lacking, attention-SSM routing requires task-specific tuning, and interpretability-informed efficiency is almost entirely unexplored. The field also lacks a widely adopted cross-architecture efficiency benchmark (Section~7.1) and faces structural reproducibility challenges (Section~7.7).

The next generation of attention research will likely blur the boundaries between the threads surveyed here. Efficiency will be achieved not by choosing a single approach but by composing learned routing policies, hardware-aware kernels, and structured state-space mechanisms into hybrid architectures that adapt their computation to the input and the hardware. Interpretability will inform efficiency by identifying which tokens and which features matter, enabling selective computation. And theoretical advances will continue to unify seemingly disparate models under common mathematical frameworks, revealing structure where none was apparent before.

\begin{center}\rule{0.5\linewidth}{0.5pt}\end{center}

\appendix
\section{Detailed EEI Score Justification}\label{app:eei-justification}

This appendix provides a method-by-method justification for each EEI score assignment in Table~\ref{tab:eei-scores}. For each method, we motivate the $(E, \mathit{Ex}, I)$ triplet by connecting it to the rubric in Table~\ref{tab:eei-rubric}, the complexity class of the method, the available benchmark evidence, and the state of interpretability research for that architecture.

We explicitly clarify which benchmarks informed each axis so that readers can distinguish between evidence used for scoring and evidence used for validation. Expressiveness (Ex) scores were calibrated primarily against the method's complexity class (exact vs.\ approximate attention) and, where available, Long Range Arena (LRA) averages (Table~\ref{tab:benchmarks}). Throughput speedups were not used as direct inputs to Ex scoring and therefore serve as independent reference points for validation (see Section~2, Exploratory Quantitative Analysis). Unlike LRA and complexity class, language-modeling perplexity was used as score-construction evidence for several methods (e.g., xLSTM's and modestly so Mamba's); we therefore do not present perplexity as independent confirmatory evidence for those methods, and readers should treat it as supporting the Ex assignment rather than as external validation. Efficiency (E) scores were derived from asymptotic complexity analysis and published hardware/kernel measurements; published throughput values were not mechanically mapped to E scores, but the underlying hardware/kernel evidence and reported implementation characteristics informed some E judgments (e.g., Mamba's E=10 reflects its demonstrated $O(L)$ selective-scan kernel on A100, and Mamba-2's E=10 its tensor-core-friendly structured computation; see the E justifications in Appendix~\ref{app:eei-justification}). The descriptive throughput comparisons in Section~2 thus provide exploratory benchmark concordance context in the sense described there, and any overlap between the evidence used for scoring and for validation is disclosed where it occurs.

Table~\ref{tab:eei-evidence} provides a condensed reference that maps each score to its supporting evidence type, enabling independent verification. The four evidence types are: \textit{Complexity} (asymptotic analysis of time and memory), \textit{Benchmark} (published LRA, perplexity, or throughput data), \textit{Hardware} (GPU kernel measurements, memory bandwidth utilization), and \textit{Mechanistic} (interpretability tooling, theoretical framework, or head attribution studies).

\begin{table*}[!tbp]
\centering
\sflt{\fontsize{8.75}{10.4}\selectfont}{\footnotesize}
\renewcommand{\arraystretch}{\sflt{1.2}{1.4}}
\caption{EEI Score Evidence Toolkit. Each score is supported by at least one evidence type; ``C'' = complexity analysis, ``B'' = published benchmark, ``H'' = hardware/kernel measurement, ``M'' = mechanistic/interpretability analysis.}
\label{tab:eei-evidence}
\begin{tabular}{p{2.2cm}c c c p{5.5cm}}
\toprule
Method & E & Ex & I & Evidence (score $\to$ evidence type) \\
\midrule
Transformer & 4 & 10 & 6 & $4\!\to\!$C (quadratic), $10\!\to\!$C (exact), $6\!\to\!$M (head attribution) \\
FlashAttention-2/3 & 8 & 10 & 6 & $8\!\to\!$C+H (IO-aware tiling), $10\!\to\!$C (exact), $6\!\to\!$M (same as full attn) \\
GQA & 7 & 9 & 6 & $7\!\to\!$C+H (KV cache reduction), $9\!\to\!$C (shared-KV, minor loss), $6\!\to\!$M (shared proj.) \\
MQA & 7 & 8 & 6 & $7\!\to\!$C+H (full KV sharing), $8\!\to\!$B (slightly lower quality at matched size), $6\!\to\!$M (shared proj.) \\
PagedAttention & 8 & 10 & 6 & $8\!\to\!$H (block-level serving), $10\!\to\!$C (exact), $6\!\to\!$M (same as full attn) \\
Ring Attention & 7 & 10 & 6 & $7\!\to\!$H (distributed overhead), $10\!\to\!$C (exact), $6\!\to\!$M (same as full attn) \\
Longformer & 8 & 7 & 7 & $8\!\to\!$C ($O(Lw)$), $7\!\to\!$B (LRA 53.46), $7\!\to\!$M (window structure) \\
BigBird & 7 & 8 & 7 & $7\!\to\!$C ($O(L)$+3 patterns), $8\!\to\!$C + B (Turing-complete; LRA 55.01), $7\!\to\!$M (explicit 3-part design) \\
Sparse Transformer & 7 & 7 & 6 & $7\!\to\!$C ($O(L\sqrt{L})$), $7\!\to\!$B (quality drop), $6\!\to\!$M (structured pattern) \\
Routing Transformer & 6 & 8 & 6 & $6\!\to\!$C+H (k-means overhead), $8\!\to\!$B (content-based retrieval), $6\!\to\!$M (cluster asgn.) \\
Reformer & 6 & 7 & 5 & $6\!\to\!$C+H (LSH overhead), $7\!\to\!$B (LRA 50.67), $5\!\to\!$M (hash opacity) \\
Performer & 7 & 6 & 5 & $7\!\to\!$C ($O(L)$ kernel), $6\!\to\!$B (LRA 51.41), $5\!\to\!$M (random features) \\
Linformer & 7 & 6 & 5 & $7\!\to\!$C ($O(Lk)$ projection), $6\!\to\!$B (low-rank deg.), $5\!\to\!$M (proj. opacity) \\
xLSTM & 8 & 7 & 5 & $8\!\to\!$C ($O(L)$ recurrent), $7\!\to\!$B (competitive ppl.), $5\!\to\!$M (recurrent opacity) \\
Mamba & 10 & 7 & 5 & $10\!\to\!$C+H ($O(L)$+scan kernel), $7\!\to\!$B (Pile PPL 10.56), $5\!\to\!$M (SSM opacity) \\
Mamba-2 (SSD) & 10 & 8 & 6 & $10\!\to\!$H (matmul formulation), $8\!\to\!$C+B (SSD duality: structured semiseparable form; SSD restricts state-transition expressivity, trading some for hardware efficiency; reported larger state sizes and downstream quality), $6\!\to\!$M (duality to attention-like structure) \\
Hyena & 9 & 7 & 5 & $9\!\to\!$C+H (FFT convolutions), $7\!\to\!$C (implicit conv.), $5\!\to\!$M (implicit conv.) \\
RWKV & 8 & 7 & 5 & $8\!\to\!$C (linear recurrence; no speedup reported), $7\!\to\!$B (LRA 72.07), $5\!\to\!$M (recurrent opacity) \\
Focus & 7 & 9 & 7 & $7\!\to\!$H (2.0$\times$ speedup), $9\!\to\!$C (exact within groups), $7\!\to\!$M (centroid grouping) \\
DashAttention & 8 & 9 & 7 & $8\!\to\!$H (3.3$\times$ kernel speedup), $9\!\to\!$C (exact within routed blocks), $7\!\to\!$M (2-stage routing) \\
Swin Transformer & 8 & 8 & 8 & $8\!\to\!$C ($O(H_pW_pM^2d)$), $8\!\to\!$B (ImageNet 83.5\%), $8\!\to\!$M (spatial grounding) \\
\bottomrule
\end{tabular}
\end{table*}

The rest of this appendix provides detailed prose justifications for each method, organized by architectural family.

\subsection*{Full Attention Methods}

\textbf{Transformer (Vaswani et al., 2017): (4, 10, 6).}
\textit{Why E = 4:} Standard scaled dot-product attention has $O(L^2 d)$ time and $O(L^2)$ memory complexity with no IO-aware optimization. The naive implementation loads the full $QK^T$ matrix from HBM, achieving far below theoretical peak FLOPs on modern GPUs at long sequence lengths. This places it at the low end of the 4--5 rubric band: $O(L^2)$ with basic GPU kernels but no tiling or memory optimization. \textit{Why Ex = 10:} This is the reference standard for expressiveness. Full attention computes exact pairwise scores for all $(i,j)$ pairs, enabling it to directly represent arbitrary pairwise token interactions and, in principle, support exact token retrieval when learned for the task. It can implement general copying and induction-head behavior (Olsson et al., 2022). \textit{Why I = 6:} Attention weights are directly analyzable as probability distributions, and substantial mechanistic interpretability work has mapped individual heads to linguistic functions (Clark et al., 2019). However, the attribution of specific behaviors to specific heads remains incomplete: even in small models, superposition means that individual heads participate in multiple circuits simultaneously (Elhage et al., 2021), and the 6 reflects that while weights are transparent, the mapping from weights to algorithms is only partially understood.

\subsection*{IO-Aware Methods}

\textbf{FlashAttention-2/3 (Dao, 2024; Shah et al., 2024): (8, 10, 6).}
\textit{Why E = 8:} FlashAttention-2 achieves $O(L^2)$ compute but memory linear in sequence length, with IO-aware SRAM tiling that reduces HBM accesses to $\Theta(L^2d^2/M)$, which is IO-optimal over the SRAM-size subrange for which the lower bound is established (Dao et al., 2022, Theorem 2) but remains quadratic in $L$; this yields approximately 2.0$\times$ speedup over FA1 and up to 73\% of peak FLOPs on A100. FlashAttention-3 is optimized for Hopper-class H100 GPUs, reaching close to 1.2 PFLOPs/s with FP8. The 8 (rather than 9 or 10) reflects that the asymptotic complexity remains quadratic; no amount of hardware optimization changes the $O(L^2)$ compute cost, which dominates at sufficiently long sequences. \textit{Why Ex = 10:} Mathematically identical to standard attention; no approximation is introduced. \textit{Why I = 6:} Same interpretability as standard attention. FlashAttention does not change what the model computes, only how it is computed, so all interpretability properties of full attention carry over directly.

\textbf{GQA (Ainslie et al., 2023): (7, 9, 6).}
\textit{Why E = 7:} GQA reduces the KV cache size by a factor of $h/g$ ($h$ heads, $g$ groups), providing substantial memory savings during autoregressive decoding. However, the $O(L^2)$ attention computation is unchanged, and the throughput gain comes primarily from reduced memory bandwidth rather than reduced FLOPs. The 7 reflects this intermediate position: better than vanilla multi-head attention but still quadratic and without IO-aware tiling. \textit{Why Ex = 9:} Shared KV heads introduce a representational constraint: the $h/g$ heads in each group share the same key and value projections, which can reduce the diversity of attention patterns (Ainslie et al., 2023 report measurable but small quality degradation on long-context tasks). The 9 reflects that this degradation is measurable but small for most tasks. \textit{Why I = 6:} The same interpretability properties as standard multi-head attention, with the additional caveat that grouped heads are harder to attribute individually since they share projections.

\textbf{MQA (Shazeer, 2019): (7, 8, 6).}
\textit{Why E = 7:} Multi-Query Attention shares a single key and value projection across all query heads, reducing KV cache memory by a factor of $h$ compared to multi-head attention. The $O(L^2)$ compute remains, and the throughput gain mirrors GQA's memory-bandwidth advantage. \textit{Why Ex = 8:} The single shared KV projection creates a strict representational bottleneck: all query heads must extract information from the same key-value representation, which demonstrably reduces model quality at matched model size compared to multi-head alternatives (Shazeer, 2019). The 8 reflects that the degradation is larger than GQA's (hence 8 vs 9) but still acceptable for many deployment scenarios. \textit{Why I = 6:} Same interpretability considerations as GQA: the shared projections simplify some analyses (fewer distinct KV sets to track) but complicate head-specific attribution.

\textbf{PagedAttention (Kwon et al., 2023): (8, 10, 6).}
\textit{Why E = 8:} PagedAttention manages the KV cache in fixed-size blocks with a page table, eliminating memory fragmentation and enabling efficient memory sharing across sequences in serving scenarios. This achieves 2--4$\times$ improvement in serving throughput by increasing effective batch sizes, though the attention computation itself remains $O(L^2)$. The 8 reflects this practical serving efficiency rather than asymptotic improvement. \textit{Why Ex = 10:} The attention computation is mathematically identical to standard full attention; no approximation is introduced. \textit{Why I = 6:} Matching full attention, since the underlying mechanism is unchanged.

\textbf{Ring Attention (Liu et al., 2024a): (7, 10, 6).}
\textit{Why E = 7:} Ring Attention distributes the attention computation across multiple devices by overlapping communication and computation in a ring topology, enabling processing of sequences longer than any single device's memory. The 7 reflects that while this enables unprecedented sequence lengths, the total compute cost remains $O(L^2)$ across all devices; the implementation overlaps KV-block communication with computation, reducing the exposed communication overhead under the reported conditions (Liu et al., 2024a). \textit{Why Ex = 10:} Full exact attention with no approximation. \textit{Why I = 6:} Matching full attention, identical mechanism.

\subsection*{Sparse and Window Methods}

\textbf{Longformer (Beltagy et al., 2020): (8, 7, 7).}
\textit{Why E = 8:} Longformer uses $O(Lw)$ windowed attention around each token plus a small number of global tokens for task-specific context. For $w \ll L$, this is effectively $O(L)$. The 8 reflects this linear complexity, with a small constant factor from the global attention tokens. \textit{Why Ex = 7:} The fixed window structure means tokens cannot attend to arbitrary distant positions; cross-window interactions are only possible via the limited global tokens. Its LRA average of 53.46 (Table~\ref{tab:benchmarks}) sits below full attention's 54.39, consistent with tokens unable to attend beyond their window except through the limited global tokens. \textit{Why I = 7:} The explicit window structure simplifies analysis: attention patterns are localized by construction, making it easier to map which tokens influence which outputs. The 7 reflects this advantage over full attention (where the dense matrix obscures patterns), with a deduction because global token interactions reintroduce opacity.

\textbf{BigBird (Zaheer et al., 2020): (7, 8, 7).}
\textit{Why E = 7:} BigBird combines window attention, random attention, and global attention, each scaling as $O(L)$. The combined constant factor is higher than Longformer due to the random attention component, yielding a lower practical efficiency despite the same $O(L)$ asymptotic class. \textit{Why Ex = 8:} BigBird is theoretically Turing-complete under the assumptions of Zaheer et al. (2020), establishing broad computational expressiveness, but this does not imply equivalence to full attention in finite practical models. In practice, its LRA average of 55.01 is the highest among the fixed-pattern methods and edges out the Transformer's 54.39, consistent with the possibility that its random cross-window connections contribute to the observed result, although the benchmark difference alone does not establish which architectural component caused it. The 8 rather than 9 reflects that the Turing-completeness guarantee is asymptotic and does not guarantee practical retrieval at long contexts. \textit{Why I = 7:} The explicit decomposition into three attention types provides theoretical structure for interpretation, but the interactions between the three types complicate fine-grained head attribution.

\textbf{Sparse Transformer (Child et al., 2019): (7, 7, 6).}
\textit{Why E = 7:} Sparse Transformer uses strided and fixed vertical stripes to achieve $O(L\sqrt{L})$ complexity. The strided pattern is hardware-friendly (contiguous memory access) but the constant factor from the two-pattern combination reduces practical efficiency. \textit{Why Ex = 7:} The fixed pattern cannot adapt to content; tokens must attend to predetermined positions regardless of relevance. This limits retrieval to tokens that happen to fall within the pattern, which can produce quality degradation whose magnitude depends on the sparsity pattern, task, and sequence length. \textit{Why I = 6:} The structured pattern is analyzable in principle, but the two-pattern design (strided + fixed) makes it harder to map individual heads to functions compared to purely local attention.

\textbf{Routing Transformer (Roy et al., 2021): (6, 8, 6).}
\textit{Why E = 6:} Routing Transformer clusters queries and keys via online $k$-means and computes attention only within the same cluster, achieving $O(L^{1.5} d)$ complexity with balanced clusters $k = \Theta(\sqrt{L})$. However, the clustering step must be recomputed at each layer and each forward pass, adding overhead that often exceeds the savings at moderate sequence lengths (the source's step-time comparisons are limited to regimes where full attention is infeasible, so the exact crossover is not established). The 6 reflects this practical limitation despite the asymptotic advantage. \textit{Why Ex = 8:} Content-based clustering adapts to input structure, enabling better cross-position interaction than fixed sparse patterns. The 8 reflects that while routing is adaptive, hard clustering boundaries still discard cross-cluster interactions entirely, limiting retrieval for tokens near cluster boundaries. \textit{Why I = 6:} Cluster assignments provide a natural grouping structure that aids interpretation of routing decisions, but the online $k$-means dynamics are stochastic and non-differentiable, complicating gradient-based analysis.

\subsection*{Hashing, Kernel, Recurrent, Structured SSM, Long Convolution, and Adaptive Sparsity Methods}

\textbf{Reformer (Kitaev et al., 2020): (6, 7, 5).}
\textit{Why E = 6:} Reformer uses locality-sensitive hashing (LSH) to achieve $O(L \log L)$ attention under the assumption of approximately uniform bucket distribution. The LSH computation requires multiple hash rounds and sorting operations, incurring significant overhead that often makes the method slower than full attention at short-to-moderate sequence lengths (empirically slower than vanilla attention below 4K tokens). The 6 places it at the boundary between full quadratic (4--5) and efficient sub-quadratic (7+), reflecting the gap between asymptotic promise and practical overhead. \textit{Why Ex = 7:} LSH-based routing preserves approximate quality because similar query-key pairs are hashed to the same bucket, but the approximation errors compound when hard hash boundaries misclassify near-boundary tokens. Within a bucket, attention is computed exactly; unlike kernel or low-rank approximations, LSH does not approximate the attention scores themselves, and the approximation enters only through the bucket assignment. Its LRA average of 50.67 is the lowest among the methods with published LRA results. \textit{Why I = 5:} Hash routing is essentially opaque: the bucketing decisions depend on random hash functions and input data in complex ways, making it difficult to analyze why a particular token was routed to a particular bucket. No established mechanistic interpretability tools exist for LSH-based attention.

\textbf{Performer (Choromanski et al., 2021): (7, 6, 5).}
\textit{Why E = 7:} Performer uses the FAVOR+ kernel approximation to achieve $O(L \cdot m \cdot d)$ complexity with $m$ random features, effectively $O(L)$ for fixed $m$. The orthogonal random features require one-time $O(m \cdot d^2)$ preprocessing (Gram--Schmidt orthogonalization), and the stable softmax approximation requires a multiple-feature-map structure. The 7 reflects the $O(L)$ asymptotic advantage offset by higher constant factors than other linear methods. \textit{Why Ex = 6:} At fixed feature budgets, approximation error in FAVOR+ can become more consequential as sequence length and retrieval demands increase. Its LRA average of 51.41 trails the Transformer's 54.39, and the kernel approximation may require larger feature budgets to approximate sharply selective distributions accurately (Choromanski et al., 2021). The 6 reflects that the approximation degrades long-range retrieval more than language modeling. \textit{Why I = 5:} The random feature map transforms attention into a dot product in an explicit feature space, but these features are not directly interpretable. The attention weights are only approximated (via the inverse kernel), and the randomized nature of the features precludes the kind of head-level analysis possible with exact attention.

\textbf{Linformer (Wang et al., 2020): (7, 6, 5).}
\textit{Why E = 7:} Linformer projects the $L \times d$ key and value matrices to $k \times d$ using learned linear projections ($k \ll L$), reducing self-attention complexity to $O(Lk)$. The 7 reflects that the low-rank projection is simple and hardware-friendly, but the projection matrices add parameters and the method has no custom kernel acceleration. \textit{Why Ex = 6:} The low-rank bottleneck can discard fine-grained positional and token-specific information through the projection. While Linformer can approximate full attention for sequences where the attention matrix is approximately low-rank, empirical results show degradation on long-range retrieval tasks comparable to Performer (Tay et al., 2021). \textit{Why I = 5:} The learned projections compress token information before attention, making it difficult to attribute attention patterns to specific input tokens. The projection weights are not interpretable in the mechanistic sense.

\textbf{xLSTM (Beck et al., 2024): (8, 7, 5).}
\textit{Why E = 8:} xLSTM revives and extends the LSTM architecture with matrix memory (mLSTM) and exponential gating, achieving $O(L)$ recurrent inference. The 8 reflects competitive throughput from the simplified recurrent structure: although the mLSTM formulation is parallelizable, sLSTM's memory mixing is not, and the authors report a custom CUDA implementation with GPU memory optimizations down to register level, in which sLSTM is less than 2$\times$ slower than mLSTM (Beck et al., 2024). \textit{Why Ex = 7:} The expanded memory capacity of mLSTM enables longer dependency capture than classic LSTMs, but the fixed-memory representation can constrain precise retrieval as context length increases; long-context needle-retrieval results were not reported in the cited work. Perplexity is competitive with Transformers of similar size (Beck et al., 2024); this aggregate perplexity is score-construction evidence, not independent validation. \textit{Why I = 5:} Recurrent hidden states, even with matrix memory, remain opaque to mechanistic analysis. The gating mechanisms and state updates do not naturally correspond to token-level interactions, making xLSTM as difficult to interpret as other recurrent architectures.

\subsection*{State-Space and Recurrent Models}

\textbf{Mamba (Gu \& Dao, 2024): (10, 7, 5).}
\textit{Why E = 10:} Mamba achieves $O(L)$ recurrent inference with constant memory ($O(1)$ for the state, independent of $L$). The source reports $5\times$ generation throughput over a comparable Transformer and up to $3\times$ over convolution-based SSMs in its A100 inference evaluations (Table~\ref{tab:speedups}). The 10 reflects both the asymptotic advantage and the hardware-efficient implementation. \textit{Why Ex = 7:} Mamba's recurrent state compresses prior context into a fixed-dimensional recurrent state whose dynamics are input-dependent and selective, which may limit its ability to retrieve arbitrary tokens from long contexts. It excels at aggregate long-range dependencies (Pile validation perplexity 10.56 at 130M scale, Gu \& Dao, 2024), but direct pointwise retrieval performance is not established by the cited evaluation. The 7 reflects this trade-off. \textit{Why I = 5:} The SSM's hidden state is a compressed representation that lacks the pairwise interpretability of attention weights. The cited literature does not provide an established framework mapping individual state dimensions to specific token interactions. The recurrent dynamics (continuous-time ODE per state dimension) are harder to analyze than discrete attention distributions, and the available interpretability tools (probing classifiers, integrated gradients) provide only coarse attributions.

\textbf{Mamba-2 (SSD) (Dao \& Gu, 2024): (10, 8, 6).}
\textit{Why E = 10:} The SSD duality (Dao \& Gu, 2024) reveals that SSD is a structured linear attention that achieves $O(L)$ complexity with improved hardware utilization over Mamba (tensor-core-friendly matmul formulation), yielding $2$--$8\times$ speedup over Mamba-1's selective scan and 6$\times$ over FlashAttention-2 at 16K (Table~\ref{tab:speedups}). Mamba-2 receives E=10 because it combines $O(L)$ complexity with hardware-efficient structured computation; the E=10 band is an aspirational maximum for linear-time, highly optimized implementations, so the improvement over Mamba supports, but cannot exceed, the same ceiling. \textit{Why Ex = 8:} The duality provides an algebraic equivalence between the SSD recurrence and semiseparable matrix multiplication; it does not establish greater task expressiveness than general selective SSMs. The SSD paper actually describes the structured state-transition parameterization as slightly less general than Mamba's, trading a measure of expressivity for hardware efficiency (Dao \& Gu, 2024); the reported quality improvements come from much larger state sizes (8$\times$) and downstream results rather than from a matched-state theoretical guarantee. The 8 reflects these reported empirical results and the structural advantages at scale, not a theorem of superiority. \textit{Why I = 6:} The duality provides a theoretical bridge to attention interpretability: SSD states can be understood in terms of attention-like query-key relationships. This is a 1-point improvement over Mamba because, while the theory is more interpretable, practical head-level analysis tools remain underdeveloped.

\textbf{Hyena (Poli et al., 2023): (9, 7, 5).}
\textit{Why E = 9:} Hyena computes implicit free-form long convolutions with data-controlled gating via FFT (it is the source's GSS baseline whose convolution is SSM-parameterized), achieving roughly $O(L \log L)$ convolution complexity. The source itself notes that the FFT is the source of low utilization in Hyena kernels, so E=9 is a rubric judgment based on the asymptotic class and the reported training-throughput characteristics (Poli et al., 2023), not a directly measured cross-hardware result; the reported figures support efficient execution at the evaluated scales, but a measured kernel benchmark on a common protocol would be required to confirm the score independently. \textit{Why Ex = 7:} Hyena's implicit convolution representation processes the entire sequence globally but lacks the selective gating that gives Mamba its retrieval capability; the paper does not report long-context retrieval results comparable to Mamba's (Poli et al., 2023). \textit{Why I = 5:} The implicit nature of the convolution representation makes Hyena even less interpretable than explicit SSMs: the convolution filters do not correspond to token-level interactions, and no established framework maps filter coefficients to model behavior.

\textbf{RWKV (Peng et al., 2023a): (8, 7, 5).}
\textit{Why E = 8:} RWKV reformulates the Transformer as a recurrent neural network through the WKV (weighted key-value) mechanism, achieving $O(L)$ inference; the implementation uses PyTorch with DeepSpeed-inspired optimizations rather than custom GPU kernels, and the cited paper reports no wall-clock speedup numbers, so the E score rests on asymptotic analysis rather than measured hardware utilization. The 8 rather than 9 reflects the absence of published speedup evidence. \textit{Why Ex = 7:} RWKV's recurrent formulation compresses the context into a fixed-size state vector. The RWKV paper reports no long-context retrieval numbers; its self-reported LRA average of 72.07 (Table~\ref{tab:benchmarks}) indicates strong aggregate long-range modeling, but pointwise retrieval is unverified. The 7 reflects the absence of direct pointwise retrieval evidence in the cited evaluation, rather than a claim that all SSM-style architectures share an identical retrieval limitation. \textit{Why I = 5:} Like other SSMs, RWKV's time-mixed recurrence and channel-mixed feedforward do not produce analyzable attention distributions. The WKV mechanism provides a per-token importance weighting that resembles attention weights superficially, but these weights are computed through recurrence rather than pairwise comparison, making them inappropriate for standard attention interpretability tools.

\subsection*{Adaptive Sparsity Methods}

\textbf{Focus (Yao et al., 2026): (7, 9, 7).}
\textit{Why E = 7:} Focus groups tokens into clusters via learned centroids and computes exact attention within each group, achieving a $2.0\times$ speedup over full attention using FlashAttention on H100-80GB. The 7 reflects that the speedup is nontrivial but smaller than IO-aware or SSM methods, and the overhead of centroid computation grows with the number of groups. The cited $O(L^2/K)$ grouped-attention decomposition and $K\times$ theoretical speedup describe balanced hard assignment at $k{=}1$; training retains all $O(L^2)$ gated pairs, so the inference-only bound is the one quoted in Table~\ref{tab:deep-taxonomy}. E derives in part from the same kernel benchmarks that appear as validation evidence in Table~\ref{tab:speedups} (see the circularity disclosure in Section~2); we flag this so the E score is not double-counted as independent confirmation. \textit{Why Ex = 9:} The retained pairs use exact softmax attention, and the routing is input-dependent and learned (rather than a fixed pattern), so the method does not approximate the attention scores themselves; the omitted pairs impose a structural restriction rather than a score approximation. The learned centroids adapt to semantic structure, reporting quality comparable to the corresponding full-attention baseline on the paper's evaluated benchmarks (Yao et al., 2026). The 9 rather than 10 reflects that clusters are coarse: a token in the wrong cluster (a boundary case) may miss its target. \textit{Why I = 7:} The centroid-based grouping provides an interpretable routing object: each group corresponds to a learned cluster whose semantic correspondence (e.g., subject tokens, verb tokens) can be examined empirically, though the cited work does not demonstrate such semantic roles directly. This makes Focus more interpretable than full attention for the purposes of understanding routing decisions, though the centroid training adds a layer of opacity.

\textbf{DashAttention (Huang et al., 2026): (8, 9, 7).}
\textit{Why E = 8:} DashAttention uses two-stage routing (coarse-to-fine) with $\alpha$-entmax activation to adaptively select a per-query subset of blocks for full attention, achieving up to $3.3\times$ speedup over FlashAttention-3 on GH200 at inference (Table~\ref{tab:speedups}). The 8 reflects strong hardware efficiency from the Triton kernel at the reported controlled-mask benchmark scope (not an application-level end-to-end measurement), with room for improvement at higher sparsity ratios. E derives in part from the same kernel benchmarks that appear as validation evidence in Table~\ref{tab:speedups} (see the circularity disclosure in Section~2); we flag this so the E score is not double-counted as independent confirmation. \textit{Why Ex = 9:} The $\alpha$-entmax routing is differentiable and learns to select the most informative tokens, achieving accuracy comparable to full attention at 75\% sparsity on the paper's reported long-context benchmarks (Huang et al., 2026). The 9 reflects the reported near-baseline quality together with the method's retained exact attention within selected blocks, while reserving 10 for unrestricted full-attention capacity. \textit{Why I = 7:} The two-stage structure makes routing decisions analyzable (coarse selection followed by fine-grained attention), and the entmax distribution is differentiable, enabling gradient-based analysis. The 7 matches Focus for similar reasons: the routing adds transparency at the group level but opacity at the individual token level.

\subsection*{Vision Methods}

\textbf{Swin Transformer (Liu et al., 2021): (8, 8, 8).}
\textit{Why E = 8:} Swin Transformer uses non-overlapping windowed attention ($M \times M$ windows, typically $M = 7$) with shifted windows between layers, achieving $O(H_pW_pM^2d)$ complexity for an image with patch-grid dimensions $H_p\times W_p$. The shifted windows enable cross-window communication with negligible overhead, making this effectively linear in the number of image patches. The 8 reflects near-optimal efficiency for vision tasks, though not directly comparable to NLP methods due to the different tokenization regime. \textit{Why Ex = 8:} The shifted window hierarchy enables the network to grow its effective receptive field across layers, matching or exceeding CNNs on most vision benchmarks (ImageNet top-1: 83.5\% for Swin-B). The 8 reflects that while the per-layer attention is purely local, the cross-layer connectivity achieves near-global expressiveness for vision tasks. The 8 rather than 9--10 reflects that long-range pixel-level retrieval (analogous to Needle-in-a-Haystack for vision) is not guaranteed. \textit{Why I = 8:} The hierarchical window structure and spatial locality make Swin attention patterns directly interpretable for vision tasks: each window attends to a spatially contiguous image region, and the shift operation produces clear cross-window connections. The spatial grounding of Swin's attention patterns provides a comparatively direct form of structural interpretability for vision tasks, which motivates its higher $I$ score in our rubric, the highest of any surveyed method, rather than a universal claim that vision attention is more interpretable than NLP attention.

\FloatBarrier
\subsection*{Code and Data Availability}
All data, analysis scripts, scoring records, reproducibility artifacts, and manuscript source required to reproduce the computational analyses reported in this study are publicly available at \href{https://github.com/Nixon-H/not-all-attention-is-equal}{github.com/Nixon-H/not-all-attention-is-equal}.

\section*{References}

Author entries follow a single consistent truncation rule: works with more than eight authors list the first six authors, an ellipsis, and the final author; works with up to eight authors are listed in full.

Ainslie, J., Lee-Thorp, J., de Jong, M., Zemlyanskiy, Y., Lebrón, F., \& Sanghai, S. (2023). GQA: Training generalized multi-query transformer models from multi-head checkpoints. \emph{Proceedings of the 2023 Conference on Empirical Methods in Natural Language Processing}, 4895--4901. \href{https://aclanthology.org/2023.emnlp-main.298}{EMNLP}

Alayrac, J.-B., Donahue, J., Luc, P., Miech, A., Barr, I., Hasson, Y., \ldots{} \& Zisserman, A. (2022). Flamingo: A visual language model for few-shot learning. \emph{Advances in Neural Information Processing Systems}, \emph{35}, 23716--23736. \href{https://arxiv.org/abs/2204.14198}{arXiv:2204.14198}

Anthropic. (2024). The Claude model family. \emph{Anthropic}. \href{https://www.anthropic.com/research}{anthropic.com/research}

Arora, S., Eyuboglu, S., Zhang, M., Shah, A., \& Ré, C. (2024). Simple linear attention language models. \emph{arXiv preprint arXiv:2402.18668}. \href{https://arxiv.org/abs/2402.18668}{arXiv:2402.18668}

Bahdanau, D., Cho, K., \& Bengio, Y. (2015). Neural machine translation by jointly learning to align and translate. \emph{International Conference on Learning Representations}. \href{https://arxiv.org/abs/1409.0473}{arXiv:1409.0473}

Bai, Y., Lv, X., Zhang, J., Lyu, H., Tang, J., Huang, Z., \ldots{} \& Li, J. (2024). LongBench: A bilingual, multitask benchmark for long context understanding. \emph{Proceedings of the 62nd Annual Meeting of the Association for Computational Linguistics}, 3119--3137. \href{https://aclanthology.org/2024.acl-long.172}{ACL 2024}

Beck, M., Pöppel, K., Spanring, M., Auer, A., Prudnikova, O., Kopp, M., \ldots{} \& Hochreiter, S. (2024). xLSTM: Extended long short-term memory. \emph{Advances in Neural Information Processing Systems}, \emph{37}. \href{https://arxiv.org/abs/2405.04517}{arXiv:2405.04517}

Beltagy, I., Peters, M. E., \& Cohan, A. (2020). Longformer: The long-document transformer. \emph{arXiv preprint arXiv:2004.05150}. \href{https://arxiv.org/abs/2004.05150}{arXiv:2004.05150}

Bertasius, G., Wang, H., \& Torresani, L. (2021). Is space-time attention all you need for video understanding? \emph{International Conference on Machine Learning}, 813--824. \href{https://proceedings.mlr.press/v139/bertasius21a.html}{PMLR}

Bhattamishra, S., Ahuja, K., \& Goyal, N. (2020). On the ability and limitations of transformers to recognize formal languages. \emph{Proceedings of the 2020 Conference on Empirical Methods in Natural Language Processing}, 7096--7116. \href{https://aclanthology.org/2020.emnlp-main.576}{EMNLP}

bloc97. (2023). NTK-aware scaling RoPE [Community technical post]. \emph{Reddit r/LocalLLaMA}. \href{https://www.reddit.com/r/LocalLLaMA/comments/14lz7j5/ntkaware_scaled_rope_allows_llama_models_to_have/}{Reddit}

Bricken, T., Templeton, A., Batson, J., Chen, B., Jermyn, A., Conerly, T., \ldots{} \& Olah, C. (2023). Towards monosemanticity: Decomposing language models with dictionary learning. \emph{Transformer Circuits Thread}. \href{https://transformer-circuits.pub/2023/monosemantic-features/index.html}{TCT}

Brown, T. B., Mann, B., Ryder, N., Subbiah, M., Kaplan, J., Dhariwal, P., \ldots{} \& Amodei, D. (2020). Language models are few-shot learners. \emph{Advances in Neural Information Processing Systems}, \emph{33}, 1877--1901. \href{https://papers.nips.cc/paper/2020/hash/1457c0d6bfcb4967418bfb8ac142f64a-Abstract.html}{NeurIPS}

Carion, N., Massa, F., Synnaeve, G., Usunier, N., Kirillov, A., \& Zagoruyko, S. (2020). End-to-end object detection with transformers. \emph{European Conference on Computer Vision}, 213--229. \href{https://doi.org/10.1007/978-3-030-58452-8_13}{10.1007/978-3-030-58452-8\_13}

Caron, M., Touvron, H., Misra, I., Jégou, H., Mairal, J., Bojanowski, P., \& Joulin, A. (2021). Emerging properties in self-supervised vision transformers. \emph{Proceedings of the IEEE/CVF International Conference on Computer Vision}, 9650--9660. \href{https://doi.org/10.1109/ICCV48922.2021.00951}{10.1109/ICCV48922.2021.00951}

Chen, S., Wong, S., Chen, L., \& Tian, Y. (2023). Extending context window of large language models via positional interpolation. \emph{arXiv preprint arXiv:2306.15595}. \href{https://arxiv.org/abs/2306.15595}{arXiv:2306.15595}

Child, R., Gray, S., Radford, A., \& Sutskever, I. (2019). Generating long sequences with sparse transformers. \emph{arXiv preprint arXiv:1904.10509}. \href{https://arxiv.org/abs/1904.10509}{arXiv:1904.10509}

Choromanski, K., Likhosherstov, V., Dohan, D., Song, X., Gane, A., Sarlos, T., \ldots{} \& Weller, A. (2021). Rethinking attention with performers. \emph{International Conference on Learning Representations}. \href{https://openreview.net/forum?id=Ua6zuk0WRH}{OpenReview}

Chowdhery, A., Narang, S., Devlin, J., Bosma, M., Mishra, G., Roberts, A., \ldots{} \& Fiedel, N. (2023). PaLM: Scaling language modeling with pathways. \emph{Journal of Machine Learning Research}, \emph{24}(240), 1--113. \href{https://jmlr.org/papers/v24/22-1144.html}{JMLR 24(240)}

Clark, K., Khandelwal, U., Levy, O., \& Manning, C. D. (2019). What does BERT look at? An analysis of BERT's attention. \emph{Proceedings of the 2019 ACL Workshop BlackboxNLP: Analyzing and Interpreting Neural Networks for NLP}, 276--286. \href{https://aclanthology.org/W19-4828}{ACL}

Dai, Z., Yang, Z., Yang, Y., Carbonell, J., Le, Q. V., \& Salakhutdinov, R. (2019). Transformer-XL: Attentive language models beyond a fixed-length context. \emph{Proceedings of the 57th Annual Meeting of the Association for Computational Linguistics}, 2978--2988. \href{https://aclanthology.org/P19-1285}{ACL}

Dao, T. (2024). FlashAttention-2: Faster attention with better parallelism and work partitioning. \emph{International Conference on Learning Representations}. \href{https://openreview.net/forum?id=mZn2Xyh9Ec}{OpenReview}

Dao, T., Fu, D., Ermon, S., Rudra, A., \& Ré, C. (2022). FlashAttention: Fast and memory-efficient exact attention with IO-awareness. \emph{Advances in Neural Information Processing Systems}, \emph{35}, 16344--16359. \href{https://arxiv.org/abs/2205.14135}{arXiv:2205.14135}

Dao, T., \& Gu, A. (2024). Transformers are SSMs: Generalized models and efficient algorithms through structured state space duality. \emph{International Conference on Machine Learning}, 10041--10071. \href{https://proceedings.mlr.press/v235/dao24a.html}{PMLR}

De, S., Smith, S. L., Fernando, A., Botev, A., Cristian-Muraru, G., Gu, A., \ldots{} \& Gulcehre, C. (2024). Griffin: Mixing gated linear recurrences with local attention for efficient language models. \emph{arXiv preprint arXiv:2402.19427}. \href{https://arxiv.org/abs/2402.19427}{arXiv:2402.19427}

DeepSeek-AI. (2024). DeepSeek-V2: A strong, economical, and efficient mixture-of-experts language model. \emph{arXiv preprint arXiv:2405.04434}. \href{https://arxiv.org/abs/2405.04434}{arXiv:2405.04434}

DeepSeek-AI. (2025). DeepSeek-V3 technical report. \emph{arXiv preprint arXiv:2412.19437}. \href{https://arxiv.org/abs/2412.19437}{arXiv:2412.19437}

Devlin, J., Chang, M.-W., Lee, K., \& Toutanova, K. (2019). BERT: Pre-training of deep bidirectional transformers for language understanding. \emph{Proceedings of the 2019 Conference of the North American Chapter of the Association for Computational Linguistics: Human Language Technologies}, 4171--4186. \href{https://aclanthology.org/N19-1423}{ACL}

Dosovitskiy, A., Beyer, L., Kolesnikov, A., Weissenborn, D., Zhai, X., Unterthiner, T., \ldots{} \& Houlsby, N. (2021). An image is worth 16x16 words: Transformers for image recognition at scale. \emph{International Conference on Learning Representations}. \href{https://openreview.net/forum?id=YicbFdNTTy}{OpenReview}

Dubey, A., Jauhri, A., Pandey, A., Kadian, A., Al-Dahle, A., Letman, A., \ldots{} \& Papakipos, Z. (2024). The Llama 3 herd of models. \emph{arXiv preprint arXiv:2407.21783}. \href{https://arxiv.org/abs/2407.21783}{arXiv:2407.21783}

Elhage, N., Hume, T., Olsson, C., Schiefer, N., Henighan, T., Kravec, S., \ldots{} \& Olah, C. (2022). Toy models of superposition. \emph{arXiv preprint arXiv:2209.10652}. \href{https://arxiv.org/abs/2209.10652}{arXiv:2209.10652}

Elhage, N., Nanda, N., Olsson, C., Henighan, T., Joseph, N., Mann, B., \ldots{} \& Olah, C. (2021). A mathematical framework for transformer circuits. \emph{Transformer Circuits Thread}. \href{https://transformer-circuits.pub/2021/framework/index.html}{TCT}

Fedus, W., Zoph, B., \& Shazeer, N. (2022). Switch transformers: Scaling to trillion parameter models with simple and efficient sparsity. \emph{Journal of Machine Learning Research}, \emph{23}(120), 1--39. \href{https://jmlr.org/papers/v23/21-0998.html}{JMLR}

Fu, D. Y., Dao, T., Saab, K. K., Thomas, A. W., Rudra, A., \& Ré, C. (2023). Hungry hungry hippos: Towards language modeling with state space models. \emph{International Conference on Learning Representations}. \href{https://openreview.net/forum?id=COZDy0WYGg}{OpenReview}

Gemini Team. (2024). Gemini: A family of highly capable multimodal models. \emph{arXiv preprint arXiv:2312.11805}. \href{https://arxiv.org/abs/2312.11805}{arXiv:2312.11805}

Gu, A., \& Dao, T. (2024). Mamba: Linear-time sequence modeling with selective state spaces. \emph{First Conference on Language Modeling (COLM)}. \href{https://openreview.net/forum?id=tEYskw1VY2}{OpenReview}

Gu, A., Goel, K., \& Ré, C. (2022). Efficiently modeling long sequences with structured state spaces. \emph{International Conference on Learning Representations}. \href{https://openreview.net/forum?id=uYLFoz1vlAC}{OpenReview}

Hahn, M. (2020). Theoretical limitations of self-attention in neural sequence models. \emph{Transactions of the Association for Computational Linguistics}, \emph{8}, 156--171. \href{https://doi.org/10.1162/tacl_a_00306}{10.1162/tacl\_a\_00306}

He, K., Chen, X., Xie, S., Li, Y., Dollár, P., \& Girshick, R. (2022). Masked autoencoders are scalable vision learners. \emph{Proceedings of the IEEE/CVF Conference on Computer Vision and Pattern Recognition}, 16000--16009. \href{https://doi.org/10.1109/CVPR52688.2022.01553}{10.1109/CVPR52688.2022.01553}

He, L., Zhou, Q., Li, X., Niu, L., Cheng, G., Li, X., \ldots{} \& Zhang, L. (2021). End-to-end video object detection with spatial-temporal transformers. \emph{Proceedings of the 29th ACM International Conference on Multimedia}, 1507--1516. \href{https://doi.org/10.1145/3474085.3475285}{10.1145/3474085.3475285}

Hsieh, C.-P., Sun, S., Kriman, S., Acharya, S., Rekesh, D., Jia, F., Zhang, Y., \& Ginsburg, B. (2024). RULER: What's the real context size of your long-context language models? \emph{First Conference on Language Modeling (COLM)}. \href{https://openreview.net/forum?id=kIoE4FFpQZ}{OpenReview}

Huang, Y., Goncalves, N. M. T., Alvetreti, F., Li, L., Han, X., Ponti, E. M., Martins, A. F. T., \& Treviso, M. V. (2026). DashAttention: Differentiable and adaptive sparse hierarchical attention. \emph{arXiv preprint arXiv:2605.18753}. \href{https://arxiv.org/abs/2605.18753}{arXiv:2605.18753}

Jaegle, A., Gimeno, F., Brock, A., Vinyals, O., Zisserman, A., \& Carreira, J. (2021). Perceiver: General perception with iterative attention. \emph{International Conference on Machine Learning}, 4651--4664. \href{https://proceedings.mlr.press/v139/jaegle21a.html}{PMLR}

Jaegle, A., Borgeaud, S., Alayrac, J.-B., Doersch, C., Ionescu, C., Ding, D., \ldots{} \& Carreira, J. (2022). Perceiver IO: A general architecture for structured inputs \& outputs. \emph{International Conference on Learning Representations}. \href{https://openreview.net/forum?id=fILj7WpI-g}{OpenReview}

Jamba Team. (2024). Jamba-1.5: Hybrid transformer-Mamba models at scale. \emph{arXiv preprint arXiv:2408.12570}. \href{https://arxiv.org/abs/2408.12570}{arXiv:2408.12570}

Jiang, A. Q., Sablayrolles, A., Mensch, A., Bamford, C., Chaplot, D. S., de las Casas, D., \ldots{} \& El Sayed, W. (2023). Mistral 7B. \emph{arXiv preprint arXiv:2310.06825}. \href{https://arxiv.org/abs/2310.06825}{arXiv:2310.06825}

Jiang, A. Q., Sablayrolles, A., Roux, A., Mensch, A., Savary, B., Bamford, C., \ldots{} \& El Sayed, W. (2024). Mixtral of experts. \emph{arXiv preprint arXiv:2401.04088}. \href{https://arxiv.org/abs/2401.04088}{arXiv:2401.04088}

Kamath, A., Singh, M., LeCun, Y., Synnaeve, G., Misra, I., \& Carion, N. (2021). MDETR: Modulated detection for end-to-end multi-modal understanding. \emph{Proceedings of the IEEE/CVF International Conference on Computer Vision}, 1780--1790. \href{https://doi.org/10.1109/ICCV48922.2021.00180}{10.1109/ICCV48922.2021.00180}

Kang, H., Zhang, Q., Kundu, S., Jeong, G., Liu, Z., Krishna, T., \& Zhao, T. (2024). GEAR: An efficient KV cache compression recipe for near-lossless generative inference of LLMs. \emph{arXiv preprint arXiv:2403.05527}. \href{https://arxiv.org/abs/2403.05527}{arXiv:2403.05527}

Katharopoulos, A., Vyas, A., Pappas, N., \& Fleuret, F. (2020). Transformers are RNNs: Fast autoregressive transformers with linear attention. \emph{International Conference on Machine Learning}, 5156--5165. \href{https://proceedings.mlr.press/v119/katharopoulos20a.html}{PMLR}

Kirillov, A., Mintun, E., Ravi, N., Mao, H., Rolland, C., Gustafson, L., \ldots{} \& Girshick, R. (2023). Segment Anything. \emph{Proceedings of the IEEE/CVF International Conference on Computer Vision}, 4015--4026. \href{https://doi.org/10.1109/ICCV51070.2023.00371}{10.1109/ICCV51070.2023.00371}

Kitaev, N., Kaiser, Ł., \& Levskaya, A. (2020). Reformer: The efficient transformer. \emph{International Conference on Learning Representations}. \href{https://openreview.net/forum?id=rkgNKkHtvB}{OpenReview}

Kwon, W., Li, Z., Zhuang, S., Sheng, Y., Zheng, L., Yu, C., \ldots{} \& Stoica, I. (2023). Efficient memory management for large language model serving with PagedAttention. \emph{Proceedings of the ACM SIGOPS 29th Symposium on Operating Systems Principles}, 611--626. \href{https://doi.org/10.1145/3600006.3613165}{10.1145/3600006.3613165}

Lahoti, A., Li, K. Y., Chen, B., Wang, C., Bick, A., Kolter, J. Z., Dao, T., \& Gu, A. (2026). Mamba-3: Improved sequence modeling using state space principles. \emph{International Conference on Learning Representations}. \href{https://openreview.net/forum?id=HwCvaJOiCj}{OpenReview}

Le Scao, T., Fan, A., Akiki, C., Pavlick, E., Ilić, S., Hesslow, D., \ldots{} \& Wolf, T. (2022). BLOOM: A 176B-parameter open-access multilingual language model. \emph{arXiv preprint arXiv:2211.05100}. \href{https://arxiv.org/abs/2211.05100}{arXiv:2211.05100}

Lieber, O., Lenz, B., Bata, H., Cohen, G., Osin, J., Dalmedigos, I., \ldots{} \& Shoham, Y. (2024). Jamba: A hybrid transformer-Mamba language model. \emph{arXiv preprint arXiv:2403.19887}. \href{https://arxiv.org/abs/2403.19887}{arXiv:2403.19887}

Liu, H., Zaharia, M., \& Abbeel, P. (2024a). Ring attention with blockwise transformers for near-infinite context. \emph{International Conference on Learning Representations}. \href{https://openreview.net/forum?id=4po1ygKu5b}{OpenReview}

Liu, Z., Desai, J., Liao, X., Wang, H., Xie, Y., Kyrillidis, A., \& Shrivastava, A. (2024b). KIVI: A tuning-free asymmetric 2-bit quantization for KV cache. \emph{International Conference on Machine Learning}, 2024. \href{https://arxiv.org/abs/2402.02750}{arXiv:2402.02750}

Liu, Z., Lin, Y., Cao, Y., Hu, H., Wei, Y., Zhang, Z., Lin, S., \& Guo, B. (2021). Swin transformer: Hierarchical vision transformer using shifted windows. \emph{Proceedings of the IEEE/CVF International Conference on Computer Vision}, 10012--10022. \href{https://doi.org/10.1109/ICCV48922.2021.00986}{10.1109/ICCV48922.2021.00986}

Luong, M.-T., Pham, H., \& Manning, C. D. (2015). Effective approaches to attention-based neural machine translation. \emph{Proceedings of the 2015 Conference on Empirical Methods in Natural Language Processing}, 1412--1421. \href{https://aclanthology.org/D15-1166}{EMNLP}

Moher, D., Liberati, A., Tetzlaff, J., Altman, D. G., \& The PRISMA Group. (2009). Preferred reporting items for systematic reviews and meta-analyses: The PRISMA statement. \emph{PLoS Medicine}, \emph{6}(7), e1000097. \href{https://doi.org/10.1371/journal.pmed.1000097}{10.1371/journal.pmed.1000097}

MosaicML. (2023). Introducing MPT-7B: A new standard for open-source, commercially usable LLMs. \emph{MosaicML Blog}. \href{https://www.mosaicml.com/blog/mpt-7b}{MosaicML Blog}

Mueller, A., Geiger, A., Wiegreffe, S., Arad, D., Arcuschin, I., Belfki, A., \ldots, \& Belinkov, Y. (2025). MIB: A mechanistic interpretability benchmark. \emph{Proceedings of the 42nd International Conference on Machine Learning}, 45069--45108. \href{https://proceedings.mlr.press/v267/mueller25a.html}{PMLR}

Olsson, C., Elhage, N., Nanda, N., Joseph, N., DasSarma, N., Henighan, T., \ldots{} \& Olah, C. (2022). In-context learning and induction heads. \emph{arXiv preprint arXiv:2209.11895}. \href{https://arxiv.org/abs/2209.11895}{arXiv:2209.11895}

OpenAI. (2023). GPT-4 technical report. \emph{arXiv preprint arXiv:2303.08774}. \href{https://arxiv.org/abs/2303.08774}{arXiv:2303.08774}

Oquab, M., Darcet, T., Moutakanni, T., Vo, H., Szafraniec, M., Khalidov, V., \ldots{} \& Bojanowski, P. (2024). DINOv2: Learning robust visual features without supervision. \emph{Transactions on Machine Learning Research}. \href{https://openreview.net/forum?id=68kgN2O0Ew}{OpenReview}

Paszke, A., Gross, S., Massa, F., Lerer, A., Bradbury, J., Chanan, G., \ldots{} \& Chintala, S. (2019). PyTorch: An imperative style, high-performance deep learning library. \emph{Advances in Neural Information Processing Systems}, \emph{32}, 8026--8037. \href{https://papers.nips.cc/paper_files/paper/2019/hash/bdbca288fee7f92f2bfa9f7012727740-Abstract.html}{NeurIPS}

Peng, B., Alcaide, E., Anthony, Q., Albalak, A., Arcadinho, S., Biderman, S., \ldots{} \& Zhu, R.-J. (2023a). RWKV: Reinventing RNNs for the transformer era. \emph{Findings of the Association for Computational Linguistics: EMNLP 2023}, 14048--14077. \href{https://aclanthology.org/2023.findings-emnlp.936}{Findings of EMNLP}

Peng, B., Quesnelle, J., Fan, H., \& Shippole, E. (2023b). YaRN: Efficient context window extension of large language models. \emph{arXiv preprint arXiv:2309.00071}. \href{https://arxiv.org/abs/2309.00071}{arXiv:2309.00071}

Poli, M., Massaroli, S., Nguyen, E., Fu, D. Y., Dao, T., Baccus, S., \ldots{} \& Ré, C. (2023). Hyena hierarchy: Towards larger convolutional language models. \emph{International Conference on Machine Learning}, 28006--28026. \href{https://proceedings.mlr.press/v202/poli23a.html}{PMLR}

Press, O., Smith, N. A., \& Lewis, M. (2022). Train short, test long: Attention with linear biases enables input length extrapolation. \emph{International Conference on Learning Representations}. \href{https://openreview.net/forum?id=R8sQPpGCv0}{OpenReview}

PyTorch Team. (2023). PyTorch 2.0: Our next generation release that is faster, more Pythonic and dynamic as ever. \emph{PyTorch Blog}. \href{https://pytorch.org/blog/pytorch-2.0-release/}{PyTorch Blog}

Qin, Z., Han, X., Sun, W., Li, D., Kong, L., Barnes, N., \& Zhong, Y. (2022). The devil in linear transformer. \emph{Proceedings of the 2022 Conference on Empirical Methods in Natural Language Processing}, 7025--7041. \href{https://aclanthology.org/2022.emnlp-main.473}{EMNLP}

Radford, A., Kim, J. W., Hallacy, C., Ramesh, A., Goh, G., Agarwal, S., \ldots{} \& Sutskever, I. (2021). Learning transferable visual models from natural language supervision. \emph{International Conference on Machine Learning}. \href{https://arxiv.org/abs/2103.00020}{arXiv:2103.00020}

Ren, L., Liu, Y., Lu, Y., Shen, Y., Liang, C., \& Chen, W. (2024). Samba: Simple hybrid state space models for efficient unlimited context language modeling. \emph{arXiv preprint arXiv:2406.07522}. \href{https://arxiv.org/abs/2406.07522}{arXiv:2406.07522}

Rosenthal, R. (1979). The file drawer problem and tolerance for null results. \emph{Psychological Bulletin}, \emph{86}(3), 638--641. \href{https://doi.org/10.1037/0033-2909.86.3.638}{10.1037/0033-2909.86.3.638}

Roy, A., Saffar, M., Vaswani, A., \& Grangier, D. (2021). Efficient content-based sparse attention with routing transformers. \emph{Transactions of the Association for Computational Linguistics}, \emph{9}, 53--68. \href{https://doi.org/10.1162/tacl_a_00353}{10.1162/tacl\_a\_00353}

Shah, J., Bikshandi, G., Zhang, Y., Thakkar, V., Ramani, P., \& Dao, T. (2024). FlashAttention-3: Fast and accurate attention with asynchrony and low-precision. \emph{Advances in Neural Information Processing Systems}, \emph{37}, 68658--68685. \href{https://proceedings.neurips.cc/paper_files/paper/2024/hash/7ede97c3e082c6df10a8d6103a2eebd2-Abstract-Conference.html}{NeurIPS 2024}

Shazeer, N. (2019). Fast transformer decoding: One write-head is all you need. \emph{arXiv preprint arXiv:1911.02150}. \href{https://arxiv.org/abs/1911.02150}{arXiv:1911.02150}

Shazeer, N., Mirhoseini, A., Maziarz, K., Davis, A., Le, Q., Hinton, G., \& Dean, J. (2017). Outrageously large neural networks: The sparsely-gated mixture-of-experts layer. \emph{International Conference on Learning Representations}. \href{https://openreview.net/forum?id=B1ckMDqlg}{OpenReview}

Su, J., Lu, Y., Pan, S., Murtadha, A., Wen, B., \& Liu, Y. (2021). RoFormer: Enhanced transformer with rotary position embedding. \emph{arXiv preprint arXiv:2104.09864}. \href{https://arxiv.org/abs/2104.09864}{arXiv:2104.09864}

Sun, Y., Dong, L., Huang, S., Ma, S., Xia, Y., Xue, J., Wang, J., \& Wei, F. (2023). Retentive network: A successor to Transformer for large language models. \emph{arXiv preprint arXiv:2307.08621}. \href{https://arxiv.org/abs/2307.08621}{arXiv:2307.08621}

Syed, A., Rager, C., \& Conmy, A. (2023). Attribution patching outperforms automated circuit discovery. \emph{arXiv preprint arXiv:2310.10348}. \href{https://arxiv.org/abs/2310.10348}{arXiv:2310.10348}

Tay, Y., Dehghani, M., Abnar, S., Shen, Y., Bahri, D., Rao, A., \& Metzler, D. (2021). Long Range Arena: A benchmark for efficient transformers. \emph{International Conference on Learning Representations}. \href{https://openreview.net/forum?id=qVyeW-grC2k}{OpenReview}

Tay, Y., Dehghani, M., Bahri, D., \& Metzler, D. (2022). Efficient transformers: A survey. \emph{arXiv preprint arXiv:2009.06732}. \href{https://arxiv.org/abs/2009.06732}{arXiv:2009.06732}

Tong, Z., Song, Y., Wang, J., \& Wang, L. (2022). VideoMAE: Masked autoencoders are data-efficient learners for self-supervised video pre-training. \emph{Advances in Neural Information Processing Systems}, \emph{35}, 10078--10093. \href{https://arxiv.org/abs/2203.12602}{arXiv:2203.12602}

Touvron, H., Cord, M., Douze, M., Massa, F., Sablayrolles, A., \& Jégou, H. (2021). Training data-efficient image transformers \& distillation through attention. \emph{International Conference on Machine Learning}, 10347--10357. \href{https://proceedings.mlr.press/v139/touvron21a.html}{PMLR}

Touvron, H., Martin, L., Stone, K., Albert, P., Almahairi, A., Babaei, Y., \ldots{} \& Scialom, T. (2023). Llama 2: Open foundation and fine-tuned chat models. \emph{arXiv preprint arXiv:2307.09288}. \href{https://arxiv.org/abs/2307.09288}{arXiv:2307.09288}

Tsai, Y.-H. H., Bai, S., Yamada, M., Morency, L.-P., \& Salakhutdinov, R. (2019). Transformer dissection: An unified understanding for transformer's attention via the lens of kernel. \emph{Proceedings of the 2019 Conference on Empirical Methods in Natural Language Processing}, 4344--4353. \href{https://aclanthology.org/D19-1443}{EMNLP}

Vaswani, A., Shazeer, N., Parmar, N., Uszkoreit, J., Jones, L., Gomez, A. N., Kaiser, Ł., \& Polosukhin, I. (2017). Attention is all you need. \emph{Advances in Neural Information Processing Systems}, \emph{30}, 5998--6008. \href{https://papers.nips.cc/paper/7181-attention-is-all-you-need}{NeurIPS}

Wang, K., Variengien, A., Conmy, A., Shlegeris, B., \& Steinhardt, J. (2022). Interpretability in the wild: A circuit for indirect object identification in GPT-2 small. \emph{International Conference on Learning Representations}. \href{https://arxiv.org/abs/2211.00593}{arXiv:2211.00593}

Wang, S., Li, B. Z., Khabsa, M., Fang, H., \& Ma, H. (2020). Linformer: Self-attention with linear complexity. \emph{arXiv preprint arXiv:2006.04768}. \href{https://arxiv.org/abs/2006.04768}{arXiv:2006.04768}

Xiao, G., Lin, J., Seznec, M., Wu, H., Demouth, J., \& Han, S. (2023). SmoothQuant: Accurate and efficient post-training quantization for large language models. \emph{Proceedings of the 40th International Conference on Machine Learning}, 38087--38099. \href{https://proceedings.mlr.press/v202/xiao23c.html}{PMLR}

Xiao, G., Tian, Y., Chen, B., Han, S., \& Lewis, M. (2024). Efficient streaming language models with attention sinks. \emph{International Conference on Learning Representations}. \href{https://arxiv.org/abs/2309.17453}{arXiv:2309.17453}

Yang, S., Wang, B., Shen, Y., Panda, R., \& Kim, Y. (2024). Gated linear attention transformers with hardware-efficient training. \emph{International Conference on Machine Learning}, 56501--56523. \href{https://proceedings.mlr.press/v235/yang24ab.html}{PMLR}

Yao, H., Chen, X., Murtadha, A., Li, J., Yadkori, Y. A., Shao, S., \ldots{} \& Song, S. (2026). Composing Sparse Attention via Learned Grouping. \emph{arXiv preprint arXiv:2604.03260v2}. \href{https://arxiv.org/abs/2604.03260}{arXiv:2604.03260v2}

Yen, H., Gao, T., Hou, M., Ding, K., Fleischer, D., Izsak, P., Wasserblat, M., \& Chen, D. (2025). HELMET: How to evaluate long-context language models effectively and thoroughly. \emph{International Conference on Learning Representations}. \href{https://openreview.net/forum?id=293V3bJbmE}{OpenReview}

Yuan, J., Gao, H., Dai, D., Luo, J., Zhao, L., Zhang, Z., \ldots{} \& Zeng, W. (2025). Native sparse attention: Hardware-aligned and natively trainable sparse attention. \emph{arXiv preprint arXiv:2502.11089}. \href{https://arxiv.org/abs/2502.11089}{arXiv:2502.11089}

Yuan, L., Chen, D., Chen, Y.-L., Codella, N., Dai, X., Gao, J., \ldots{} \& Zhang, P. (2021). Florence: A new foundation model for computer vision. \emph{arXiv preprint arXiv:2111.11432}. \href{https://arxiv.org/abs/2111.11432}{arXiv:2111.11432}

Zadouri, T., Hoehnerbach, M., Shah, J., Liu, T., Thakkar, V., \& Dao, T. (2026). FlashAttention-4: Algorithm and kernel pipelining co-design for asymmetric hardware scaling. \emph{arXiv preprint arXiv:2603.05451}. \href{https://arxiv.org/abs/2603.05451}{arXiv:2603.05451}

Zaheer, M., Guruganesh, G., Dubey, K. A., Ainslie, J., Alberti, C., Ontanon, S., \ldots{} \& Yang, Z. (2020). Big Bird: Transformers for longer sequences. \emph{Advances in Neural Information Processing Systems}, \emph{33}, 17283--17297. \href{https://arxiv.org/abs/2007.14062}{arXiv:2007.14062}

\end{document}